\documentclass[10pt,twocolumn,letterpaper]{article}

\usepackage{cvpr}
\usepackage{times}
\usepackage{epsfig}
\usepackage{graphicx}
\usepackage{amsmath,amssymb,amsthm}
\usepackage{bm}

\usepackage{array}
\newcolumntype{P}[1]{>{\centering\arraybackslash}p{#1}}

\usepackage{algorithm,algpseudocode}

\usepackage[]{multirow,rotating,subcaption}
\usepackage[justification=justified]{caption}

\graphicspath{{./}{figures/}}

\newtheorem{theorem}{Theorem}[section]
\newtheorem{lemma}[theorem]{Lemma}

\usepackage[pagebackref=true,breaklinks=false,colorlinks,bookmarks=false]{hyperref}

\cvprfinalcopy % *** Uncomment this line for the final submission

\def\cvprPaperID{4265} % *** Enter the CVPR Paper ID here

\ifcvprfinal\fi
\begin{document}

%%%%%%%%% TITLE
\title{Compassionately Conservative Balanced Cuts for Image Segmentation}

\author{Nathan D. Cahill $\quad$ Tyler L. Hayes $\quad$ Renee T. Meinhold $\quad$ John F. Hamilton \\
Rochester Institute of Technology\\
{\tt\small \{ndcsma, tlh6792, rtm9271, jfhsms\}@rit.edu}}

\maketitle

%%%%%%%%%
\begin{abstract}
The Normalized Cut (NCut) objective function, widely used in data clustering and image segmentation, quantifies the cost of graph partitioning in a way that biases clusters or segments that are balanced towards having lower values than unbalanced partitionings. However, this bias is so strong that it avoids \emph{any} singleton partitions, even when vertices are very weakly connected to the rest of the graph. Motivated by the B\"uhler-Hein family of balanced cut costs, we propose the family of \emph{Compassionately Conservative Balanced (CCB) Cut} costs, which are indexed by a parameter that can be used to strike a compromise between the desire to avoid too many singleton partitions and the notion that \emph{all} partitions should be balanced. We show that CCB-Cut minimization can be relaxed into an orthogonally constrained $\ell_{\tau}$-minimization problem that coincides with the problem of computing Piecewise Flat Embeddings (PFE) for one particular index value, and we present an algorithm for solving the relaxed problem by iteratively minimizing a sequence of reweighted Rayleigh quotients (IRRQ). Using images from the BSDS500 database, we show that image segmentation based on CCB-Cut minimization provides better accuracy with respect to ground truth and greater variability in region size than NCut-based image segmentation.  

\end{abstract}

%%%%%%%%%
\section{Introduction}

The Normalized Cut (NCut) graph partitioning cost function was first introduced over two decades ago to tackle the perceptual grouping problem \cite{shi1997nci,shi2000nci}, and its approximate minimization via continuous relaxation into a generalized eigenvector problem has emerged as one of the fundamental techniques of data clustering \cite{jain2010dcl}.  

\begin{figure}[t]
	\centering
  \begin{subfigure}[t]{0.15\textwidth}
		\centering
    \includegraphics[width=\textwidth]{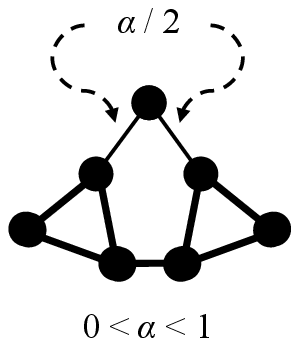}
		\caption{Toy Graph}
    \label{fig:toyGraph:orig}
  \end{subfigure}
  \,
  \begin{subfigure}[t]{0.15\textwidth}
		\centering
    \includegraphics[width=\textwidth]{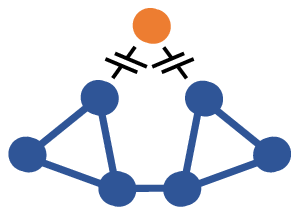}
    \caption{$\operatorname{arg\,min}$ Cut}
    \label{fig:toyGraph:minCut}
  \end{subfigure}
  \,
  \begin{subfigure}[t]{0.15\textwidth}
		\centering
    \includegraphics[width=\textwidth]{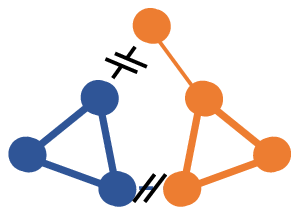}
    \caption{$\operatorname{arg\,min}$ NCut}
    \label{fig:toyGraph:minNCut}
  \end{subfigure} 
	\caption{(a) A graph having unit weight edges except for the two edges with weight $\alpha/2$. (b) Minimizing the Cut cost removes the weakly-connected vertex, whereas (c) minimizing the Normalized Cut (NCut) cost yields a more "balanced" partitioning.}
  \label{fig:toyGraph}
\end{figure}

The key observation that motivated the normalization of graph cut cost functions is that when graph partitioning is performed by minimizing the sum of the weights of edges that are cut (the Cut cost), the resulting partitions will be \emph{unbalanced}; vertices that are weakly connected to the rest of the graph are likely to be separated to form singleton partitions, while the rest of the graph vertices are likely to remain in large partitions. Normalizing the cut cost by the degrees of each partition (NCut \cite{shi1997nci,shi2000nci}), the size of each partition (Average Cut \cite{sarkar2000sll}), or the minimum of the size of each partition and its complement (Ratio or Cheeger Cut \cite{cheeger1969lbs}) yield balanced partitions that have similar total degree or size when minimized.

Is it possible that these types of normalization go too far? Consider the example of Figure \ref{fig:toyGraph}, in which we wish to partition the toy graph into two subgraphs. This graph (\ref{fig:toyGraph:orig}) contains seven vertices and nine edges; all but two of the edges have unit weight, and the two indicated edges have weight $\alpha/2$ for $\alpha\in\left(0,1\right)$. Since $\alpha<1$, the degree of the vertex sharing these two edges is guaranteed to be smaller than the degree of any other vertex in the graph. Hence, as shown in (\ref{fig:toyGraph:minCut}), minimizing the Cut cost will separate that weakly-connected vertex from the rest of the graph. Minimizing the NCut cost creates a more balanced partitioning as shown in (\ref{fig:toyGraph:minNCut}); however, this partitioning will \emph{always} have a lower-cost NCut than removing the weakly-connected vertex, \emph{even when $\alpha$ is infinitesimally small}.   Perhaps the Ratio Cut (RCut) is a better choice: for this graph, $\alpha=2/3$ is a critical value above which (\ref{fig:toyGraph:minNCut}) minimizes the RCut cost, and below which (\ref{fig:toyGraph:minCut}) yields the minimum. But is there really anything special about $\alpha = 2/3$?

If this graph represented an image to be segmented, one would want to use ground-truth segmentation maps from training images to suggest an optimal critical value for $\alpha$, above which the more ``balanced'' partitioning is desired, and below which the weakly-connected vertex should be separated. The ground truth might dictate that even the Ratio Cut cost provides too much normalization, or instead, that it does not provide enough. 

\begin{figure}[t]
  \centering
	\begin{subfigure}[t]{0.23\textwidth}
		\centering
    \includegraphics[width=\textwidth]{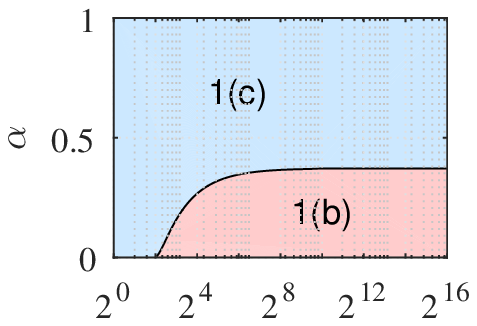}
		\caption{$\operatorname{arg\,min}\textrm{BHN}_{p}$}
    \label{fig:toyGraph:BHNRegions}
  \end{subfigure}
  \,
  \begin{subfigure}[t]{0.23\textwidth}
		\centering
    \includegraphics[width=\textwidth]{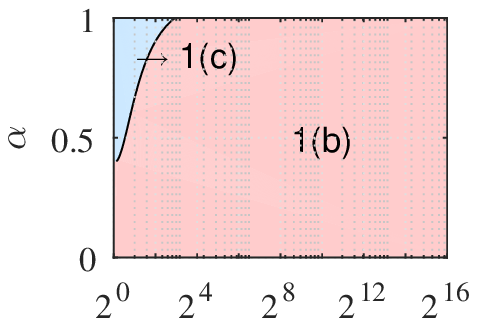}
    \caption{$\operatorname{arg\,min}\textrm{BHR}_{p}$}
    \label{fig:toyGraph:BHRRegions}
	\end{subfigure} 
	\caption{Using the B\"uhler-Hein $p$-Normalized (a) and $p$-Ratio (b) Cut costs to partition the graph in Fig. (\ref{fig:toyGraph:orig}). Over all possible $\left(p,\alpha\right)$, either the partitionings in Fig.'s (\ref{fig:toyGraph:minCut}) or (\ref{fig:toyGraph:minNCut}) yield the minimum value. For both families of costs, there are substantial ranges of $\alpha$ for which only one partitioning emerges for all $p$.}
  \label{fig:toyGraph:BH}
\end{figure}

One potential idea is to use the infinite families of balanced cost functions defined by B\"uhler and Hein \cite{buhler2009scb} and indexed by a parameter $p\in\left(1,\infty\right)$. We refer to these families as the B\"uhler-Hein $p$-Normalized ($\textrm{BHN}_{p}$) and the B\"uhler-Hein $p$-Ratio ($\textrm{BHR}_{p}$) Cuts. $\textrm{BHN}_{p}$ is equivalent to NCut for $p=2$ and the Normalized Cheeger (NCh) Cut as $p\rightarrow 1^{+}$, and $\textrm{BHR}_{p}$ is equivalent to RCut for $p=2$ and the Ratio Cheeger (RCh) Cut as $p\rightarrow 1^{+}$, so both families of cut costs can be thought of as interpolating between different normalization factors. However, as shown in Fig. (\ref{fig:toyGraph:BH}), when we minimize $\textrm{BHN}_{p}$ Cut and $\textrm{BHR}_{p}$ Cut for the graph in Fig. (\ref{fig:toyGraph}), there are substantial ranges of $\alpha$ for which no value of $p$ will allow a bifurcation between solutions.   

\begin{figure}[t]
  \centering
  \begin{subfigure}[t]{0.23\textwidth}
		\centering
    \includegraphics[width=\textwidth]{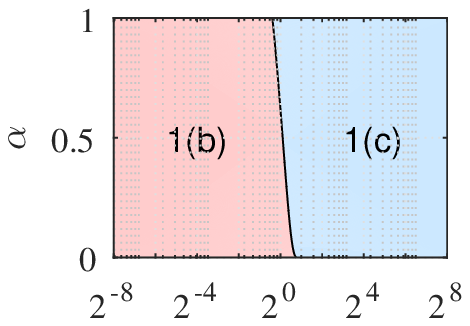}
    \caption{$\operatorname{arg\,min}\textrm{CCN}_{\tau}$}
    \label{fig:toyGraph:CCNRegions}
  \end{subfigure}
  \,
  \begin{subfigure}[t]{0.23\textwidth}
		\centering
    \includegraphics[width=\textwidth]{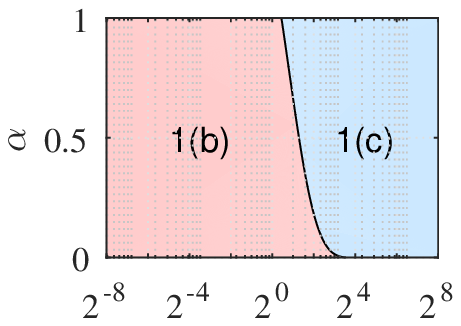}
    \caption{$\operatorname{arg\,min}\textrm{CCR}_{\tau}$}
    \label{fig:toyGraph:CCRRegions}
  \end{subfigure}	
	%\vspace{6pt}
	\caption{Using the $\textrm{CCN}_{\tau}$ (a) and $\textrm{CCR}_{\tau}$ (b) Cut costs to partition the graph in Fig. (\ref{fig:toyGraph:orig}). Over all possible $\left(p,\alpha\right)$, either the partitionings in Fig.'s (\ref{fig:toyGraph:minCut}) or (\ref{fig:toyGraph:minNCut}) yield the minimum value. For both families of costs, for any choice of $\alpha\in\left(0,1\right)$, ranges of $\tau$ exist that enable either partitioning.}
  \label{fig:toyGraph:CCB}
\end{figure}

In this paper, we show that although the B\"uhler-Hein costs provide a mechanism for interpolating between Normalized (or Ratio) Cuts and Normalized (or Ratio) Cheeger Cuts, they do not provide a path for interpolating all the way to the unnormalized Cut cost. However, a different set of infinite families of balanced cost functions \emph{can} be constructed that \emph{do} enable such interpolation. Our proposed families are called the Compassionately Conservative Normalized (CCN) Cut and the Compassionately Conservative Ratio (CCR) Cut. Both are indexed by a parameter $\tau\in\left(0,\infty\right)$; $\textrm{CCN}_{\tau}$ is equivalent to NCut for $\tau=2$ and the unnormalized Cut cost for $\tau=0$, and $\textrm{CCR}_{\tau}$ is equivalent to RCut for $\tau=2$ and the unnormalized Cut cost for $\tau=0$. As we can see from Fig. (\ref{fig:toyGraph:CCB}), when we minimize $\textrm{CCN}_{\tau}$ and $\textrm{CCR}_{\tau}$ for the graph in Fig. (\ref{fig:toyGraph}), different ranges of $\tau$ can be selected that will allow a bifurcation between solutions \emph{for any value of $\alpha$}.
  
This paper makes four novel contributions to graph partitioning in computer vision:
\begin{itemize}
	\item presenting Compassionately Conservative Balanced (CCB) Cuts: families of cut costs that enable normalizations ranging from Normalized/Ratio Cuts to unnormalized Cuts and can be naturally extended to $k$-way partitionings with $k>2$;
	\item presenting a continuous relaxation of CCB Cut minimization and illuminating its connection to computing Piecewise Flat Embeddings (PFE) \cite{yu2015pfe};
	\item presenting an efficient algorithm for solving the relaxed problem via minimizing a succession of reweighted Rayleigh quotients (IRRQ); and
	\item demonstrating empirical advantages of the CCB Cut costs and the IRRQ minimization algorithm for the application of image segmentation.
\end{itemize}

%%%%%%%%% 
\section{Balanced Cut Costs}\label{sec:BCut}

Consider an undirected weighted graph $\mathcal{G} = \left(V,E\right)$ that we wish to partition into two disjoint subgraphs $\mathcal{G}_{i} = \left(V_{i},E_{i}\right)$, $i = 1,2$, by removing the edges connecting $V_{1}$ and $V_{2}$. A standard cost of partitioning $\mathcal{G}$ is the \emph{Cut} cost, defined as the total weight of the edges that have been removed:
\begin{align} \label{eq:cutcost} 
	\textrm{Cut}\!\left(V_{1},V_{2}\right) &= \sum_{v_{i}\in V_{1}, v_{j}\in V_{2}}{w_{i,j}} \enspace ,
\end{align}
where the vertex set $V = \left\{v_{1},v_{2},\cdots,v_{n}\right\}$, and $\mathbf{W}$ is the weighted adjacency matrix (or \emph{affinity} matrix) of $\mathcal{G}$.

Minimizing the Cut cost is undesirable, however, as it can often yield partitionings that simply disconnect one vertex from the rest of the graph \cite{wu1993ogt}. More balanced partitions emerge if the Cut cost is normalized by some function of the total sizes or total degrees (volumes) of the subgraphs. Such \emph{balanced} cut costs can be expressed generally by:
\begin{align} \label{eq:bcutcost}
	\textrm{BCut}\!\left(V_{1},V_{2}\right) &= \frac{\textrm{Cut}\!\left(V_{1},V_{2}\right)}{\Theta\!\left(V_{1},V_{2}\right)} \enspace ,
\end{align}
where $\Theta\!\left(V_{1},V_{2}\right)$ is a symmetric function that decreases with increasing differences in the sizes or degrees of $V_{1}$ and $V_{2}$. A variety of balanced cut costs have been proposed in the literature, including the Normalized Cut \cite{shi1997nci,shi2000nci}, Average Cut \cite{sarkar2000sll}, Ratio Cut \cite{cheng1991itp,hagen1992nsm}, and Normalized and Ratio Cheeger Cuts \cite{buhler2009scb,cheeger1969lbs}.

In Theorem 4.1 of \cite{buhler2009scb}, B\"{u}hler and Hein show that there exists an infinite family of balanced cut costs that contains Normalized, Average, Ratio, and Cheeger Cuts. If we consider the following function:
\begin{align} \label{eq:buehlercutweight}
	\Phi_{p}\!\left(v\right) &= \frac{2^{p}}{\left(\left(\frac{1}{v}\right)^{\frac{1}{p-1}} + \left(\frac{1}{1-v}\right)^{\frac{1}{p-1}}\right)^{p-1}} \enspace ,
\end{align}
for $p>1$ and $v\in\left(0,1\right)$, then $\textrm{BCut}\!\left(V_{1},V_{2}\right)$ with $\Theta\!\left(V_{1},V_{2}\right) = \Phi_{p}\!\left(\left|V_{1}\right|\!/\!\left|V\right|\right)$ is the aforementioned $\textrm{BHR}_{p}$ Cut, and it approaches the Ratio Cheeger Cut as $p\rightarrow 1^{+}$ and equals the Average/Ratio Cuts for $p=2$. If $\Theta\!\left(V_{1},V_{2}\right) = \Phi_{p}\!\left(\textrm{Vol}\!\left(V_{1}\right)\!/\textrm{Vol}\!\left(V\right)\right)$, $\textrm{BCut}\!\left(V_{1},V_{2}\right)$ is the $\textrm{BHN}_{p}$ Cut, and it approaches the Normalized Cheeger Cut as $p\rightarrow 1^{+}$ and equals the Normalized Cut for $p=2$. (The volume of a subgraph is defined as $\textrm{Vol}\!\left(V_{\ell}\right) = \sum_{v_{j}\in V_{\ell}}{d_{j}}$, where $d_{j} = \sum_{m}{w_{j,m}}$ is the degree of vertex $v_{j}$.) Note that this is a rescaling of the balanced cut costs in \cite{buhler2009scb}, defined so that the maximum value of $\Theta\!\left(V_{1},V_{2}\right)$ is always unity.

\begin{figure}[tb]
\centering
\hspace{14pt} \includegraphics[width=2.83in]{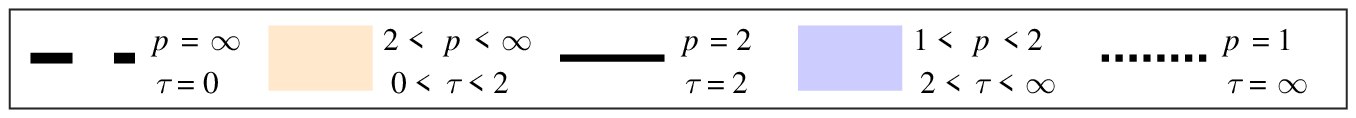} \\[0.3ex]
\includegraphics[width=1.5in]{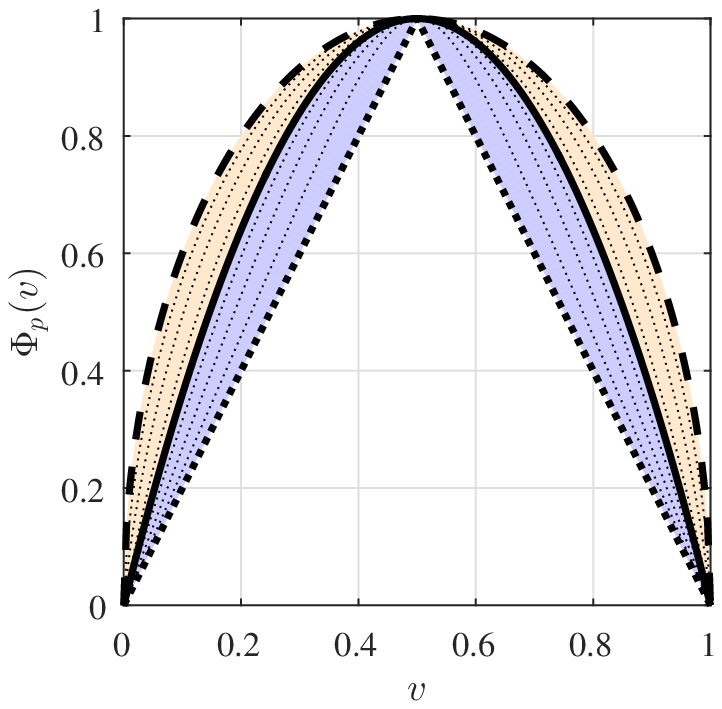} \,
\includegraphics[width=1.5in]{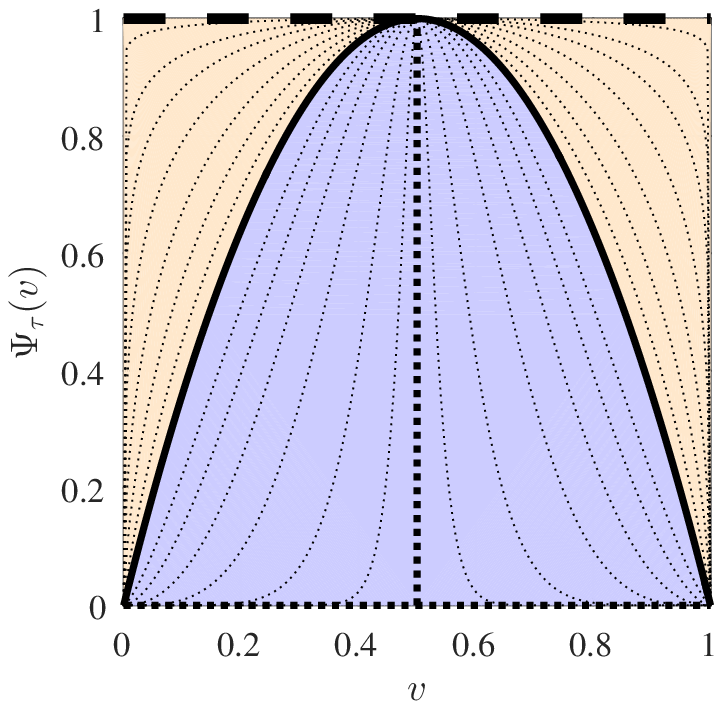}
\caption{Left: the family $\Phi_{p}\!\left(v\right)$ that enables the balanced cut costs defined by \cite{buhler2009scb}, which includes Normalized/Ratio Cuts and Cheeger Cuts but limits the extent of more conservative normalizations. Right: the family $\Psi_{\tau}\!\left(v\right)$ that enables the proposed balanced cut costs; for $0<\tau<2$, the proposed costs interpolate from Normalized/Ratio Cuts to Cuts with no normalization at all.}
\label{fig:buehlerNorm}
\end{figure}

In addition to providing a way to interpolate between Cheeger Cuts and Normalized/Ratio Cuts for $1<p<2$, (\ref{eq:buehlercutweight}) enables cut costs with normalizations that are slightly more conservative (closer to unity) than Normalized/Ratio Cuts by selecting $p>2$. It is straightforward to show this by noting that:   
\begin{align} \label{eq:phi:infty}
	\Phi_{\infty}\!\left(v\right) &= \lim_{p\rightarrow\infty}{\Phi_{p}\!\left(v\right)} = \sqrt{4v\!\left(1-v\right)} = \sqrt{\Phi_{2}\!\left(v\right)} \enspace ,
\end{align}
and so $1>\Phi_{\infty}\!\left(v\right)>\Phi_{2}\!\left(v\right)$ for all $v\in\left(0,1\right)$, $v\ne 1/2$, which can be seen in the left side of Figure \ref{fig:buehlerNorm}.

The behavior of (\ref{eq:buehlercutweight}) as $p\rightarrow\infty$ suggests that there is room to define even more conservative normalization functions. We do so in this paper by considering the following transformation of $\Phi_{p}\!\left(v\right)$:
\begin{align} 
	\Psi_{\tau}\!\left(v\right) &= \Psi_{\frac{2}{p-1}}\!\left(v\right) = \left(\Phi_{p}\!\left(v\right)\right)^{\frac{1}{p-1}} \label{eq:conservativecutweight} \\
	&= \frac{2^{p/\left(p-1\right)}}{\left(\frac{1}{v}\right)^{\frac{1}{p-1}} + \left(\frac{1}{1-v}\right)^{\frac{1}{p-1}}}
	= \frac{2^{1+\tau/2}}{\left(\frac{1}{v}\right)^{\tau/2} + \left(\frac{1}{1-v}\right)^{\tau/2}} \enspace , \nonumber
\end{align}
where $\tau = 2/\!\left(p-1\right)$. (The reason for switching parametrizations from $p$ to $\tau$ will be addressed at the end of Section \ref{sec:relaxCCBCut}.) Clearly $\Psi_{2}\!\left(v\right) = \Phi_{2}\!\left(v\right)$, and so Normalized/Ratio Cuts arise from employing (\ref{eq:conservativecutweight}) in place of (\ref{eq:buehlercutweight}) for $\tau=2$. More interestingly, however, is that $\Psi_{0}\!\left(v\right) = 1$, and so for $0<\tau<2$, defining $\textrm{BCut}\!\left(V_{1},V_{2}\right)$ with $\Theta\!\left(V_{1},V_{2}\right) = \Psi_{\tau}\!\left(\left|V_{1}\right|\!/\!\left|V\right|\right)$ or $\Psi_{\tau}\!\left(\textrm{Vol}\!\left(V_{1}\right)\!/\textrm{Vol}\!\left(V\right)\right)$ provides a way to interpolate between Normalized/Ratio Cuts and Cuts with no normalization at all. This behavior of increasing $\Psi_{\tau}\!\left(v\right)$ as $\tau\rightarrow 0^{+}$ can be seen in the right side of Figure \ref{fig:buehlerNorm}. 

Of pedagogical interest is that $\lim_{\tau\rightarrow\infty}{\Psi_{\tau}\!\left(v\right)} = \delta\!\left(v-1/2\right)$, and so as $\tau\rightarrow\infty$, employing (\ref{eq:conservativecutweight}) to define $\Theta\!\left(V_{1},V_{2}\right)$ would yield balanced cut costs that diverge whenever the sizes or degrees of $V_{1}$ and $V_{2}$ are unequal. Since our interest in this paper is to explore the impact of cut costs having more conservative normalizations than Normalized/Ratio Cuts, we will restrict our attention to the use of $\Psi_{\tau}\!\left(v\right)$ for $\tau\in\left(0,2\right)$.

In addition to enabling a greater range of conservative normalizations, the use of (\ref{eq:conservativecutweight}) in place of (\ref{eq:buehlercutweight}) yields an added benefit: the resulting Balanced Cut costs can be directly extended to form costs of partitionings into $k>2$ subgraphs. (As discussed in \cite{buhler2009scb}, it is not obvious whether such a direct extension to the B\"uhler-Hein cut costs exists.) If we now consider that we wish to partition $\mathcal{G}$ into $k$ disjoint subgraphs $\mathcal{G}_{i} = \left(V_{i},E_{i}\right)$, $i = 1,2,\ldots,k$, then an unbalanced way to measure the cost of the partitioning is the \emph{multiway} cut cost, defined in terms of pairwise cut costs as:
\begin{align} \label{eq:mcutcost} 
	\textrm{Cut}\!\left(V_{1},\ldots,V_{k}\right) &= \frac{1}{2}\sum_{\ell=1}^{k}{\textrm{Cut}\!\left(V_{\ell},V\backslash V_{\ell}\right)} \enspace .
\end{align}

The family of multiway balanced cut costs that we propose is described generally by the \emph{Compassionately \mbox{Conservative} Balanced Cut}:
\begin{align} \label{eq:ccbcutcost}
	\textrm{CCB}_{\tau}\!\left(V_{1},\ldots,V_{k}\right) &= \frac{1}{2}\sum_{\ell=1}^{k}{\frac{\textrm{Cut}\!\left(V_{\ell},V\backslash V_{\ell}\right)}{2^{\tau/2}\!\left(\sum_{v_{j}\in V_{\ell}}{\pi_{j}}\right)^{\tau/2}}} \enspace ,
\end{align}
where $\bm{\pi} = \left[\pi_{1},\ldots ,\pi_{n}\right]^{\mathbf{T}}$ is a user-defined vector of positive weights. If $\bm{\pi} = \mathbf{1}_{n}$, then $\sum_{v_{j}\in V_{\ell}}{\pi_{j}} = \left|V_{\ell}\right|$, and we refer to (\ref{eq:ccbcutcost}) as the \emph{Compassionately Conservative Ratio Cut} ($\textrm{CCR}_{\tau}$). If $\bm{\pi} = \mathbf{d}$ (the degree vector), then $\sum_{v_{j}\in V_{\ell}}{\pi_{j}} = \textrm{Vol}\!\left(V_{\ell}\right)$, and we refer to (\ref{eq:ccbcutcost}) as the \emph{Compassionately Conservative Normalized Cut} ($\textrm{CCN}_{\tau}$). With some algebraic manipulation, it is straightforward to see that (\ref{eq:ccbcutcost}) is a generalization of (\ref{eq:bcutcost}); when $k=2$, (\ref{eq:ccbcutcost}) reduces to (\ref{eq:bcutcost}) with $\Theta\!\left(V_{1},V_{2}\right) = \Psi_{\tau}\!\left(\sum_{v_{j}\in V_{1}}{\pi_{j}}\middle/\sum_{v_{j}\in V}{\pi_{j}}\right)$. Furthermore, (\ref{eq:ccbcutcost}) generalizes multiclass cut costs presented in the research literature; $\textrm{CCB}_{2}$, $\textrm{CCR}_{2}$, and $\textrm{CCN}_{2}$ are scaled versions of the Multiclass Penalized Cut \cite{zhang2008msc}, the Multiclass Ratio Cut \cite{von2007tsc}, and the Multiclass Normalized Cut \cite{yu2003msc,von2007tsc}, respectively, and so choosing $\tau\in\left(0,2\right)$ generates Multiclass Balanced Cut costs that interpolate between (\ref{eq:mcutcost}) and any of these previously developed costs.

%%%%%%%%%
\section{Relaxing the CCB Cut}\label{sec:relaxCCBCut}

As with many balanced cut costs, minimizing the CCB Cut is NP-hard. However, we can formulate a continuous relaxation of $\textrm{CCB}_{\tau}$ for all $\tau>0$. Unfortunately, the spectral clustering relaxation of $\textrm{BHN}_{p}$ and $\textrm{BHR}_{p}$ in terms of the second eigenvector of the graph $p$-Laplacian \cite{buhler2009scb} is of no help; it is only tight as $p\rightarrow 0^{+}$ (and therefore not for $\tau\in\left(0,2\right)$), and furthermore, CCB Cuts are not even members of the B\"uhler-Hein family of cuts except for when $\tau=p=2$. Hence, we must identify a different relaxation. 

To do so, we first reformulate $\textrm{CCB}_{\tau}$ to express it in terms of an $n\times k$ indicator matrix $\mathbf{X}$ so that $X_{i,j} = 1$ if $v_{i}\in V_{j}$ and $X_{i,j} = 0$ otherwise. If $\mathbf{x}_{i}$ is the $i^{\textrm{th}}$ column of $\mathbf{X}$, then $\sum_{v_{j}\in V_{i}}{\pi_{j}}$ can be written in terms of the diagonal matrix $\bm{\Pi} = \textrm{diag}\!\left(\bm{\pi}\right)$ as $\mathbf{x}_{i}^{\mathbf{T}}\bm{\Pi}\mathbf{x}_{i}$, and the pairwise cut cost between $V_{i}$ and $V\backslash V_{i}$ can be written as $\textrm{Cut}\!\left(V_{i},V\backslash V_{i}\right) = \mathbf{x}_{i}^{\mathbf{T}}\mathbf{W}\left(\mathbf{1}-\mathbf{x}_{i}\right) = \mathbf{x}_{i}^{\mathbf{T}}\mathbf{d} - \mathbf{x}_{i}^{\mathbf{T}}\mathbf{W}\mathbf{x}_{i} = \mathbf{x}_{i}^{\mathbf{T}}\mathbf{D}\mathbf{x}_{i} - \mathbf{x}_{i}^{\mathbf{T}}\mathbf{W}\mathbf{x}_{i} = \mathbf{x}_{i}^{\mathbf{T}}\mathbf{L}\mathbf{x}_{i}$, where $\mathbf{L} = \mathbf{D} - \mathbf{W}$. This allows us to express (\ref{eq:ccbcutcost}) as:
\begin{align} 
	\textrm{CCB}_{\tau}\!\left(V_{1},\ldots,V_{k}\right) &= 
		\frac{1}{2}\sum_{\ell=1}^{k}{\frac{\mathbf{x}_{\ell}^{\mathbf{T}}\mathbf{L}\mathbf{x}_{\ell}}
			{2^{\tau/2}\!\left(\mathbf{x}_{\ell}^{\mathbf{T}}\bm{\Pi}\mathbf{x}_{\ell}\right)^{\tau/2}}} \enspace . \label{eq:ccbcutEquiv} 
\end{align}
Minimizing (\ref{eq:ccbcutcost}) is equivalent to minimizing (\ref{eq:ccbcutEquiv}) subject to the constraint that $\mathbf{X^{T}X}$ is positive diagonal, which ensures that none of the $V_{i}$'s will collapse to the empty set.

To guide us towards an appropriate $\textrm{CCB}_{\tau}$ relaxation, we first consider how the Multiclass Penalized Cut ($\textrm{CCB}_{2}$) can be relaxed. Using an argument similar to Yu and Shi \cite{yu2003msc}, we see that if $\mathbf{Y} = \mathbf{X}\!\left(\mathbf{X^{T}\bm{\Pi}X}\right)^{-1/2}$, then $\mathbf{Y^{T}\bm{\Pi}Y}=\mathbf{I}$ and (\ref{eq:ccbcutcost}) is equivalent to a scalar multiple of $\textrm{tr}\!\left(\mathbf{Y^{T}LY}\right)$. Hence, the solution to minimizing a relaxed version of $\textrm{CCB}_{2}$ is $\mathbf{\tilde{Y}} = \mathbf{UQ}$, where $\mathbf{U}$ is the $n\!\times\! k$ matrix whose columns are the orthonormal eigenvectors $\mathbf{u}_{2}$, $\ldots$, $\mathbf{u}_{k+1}$ corresponding to the smallest nontrivial eigenvalues of $\bm{\Pi}^{-1/2}\mathbf{L}\bm{\Pi}^{-1/2}$, and $\mathbf{Q}$ is an arbitrary $k\!\times k$ orthogonal matrix. The optimal solution $\mathbf{\tilde{X}}$ to (\ref{eq:ccbcutcost}) with $\tau=2$ can then be approximated by $k$-means clustering \cite{ng2002osc}, nonmaximal suppression \cite{yu2003msc} or Procrustean rounding \cite{zhang2008msc} on $\mathbf{\tilde{Y}}$. 

Note that if we define $\mathbf{\hat{y}}_{i}$ to be the transpose of the $i^{\textrm{th}}$ row of $\mathbf{Y}$, then minimizing the relaxed version of $\textrm{CCB}_{2}$ is equivalent to solving the constrained minimization problem:
\begin{align}
	\min_{\mathbf{Y}\in\mathbb{R}^{n\!\times\! k}} &\quad \mathcal{J}_{2}\!\left(\mathbf{Y}\right) := \sum_{i=1}^{n}{\sum_{j=1}^{n}{w_{i,j}\!\left\|\mathbf{\hat{y}}_{i}-\mathbf{\hat{y}}_{j}\right\|_{2}^{2}}} \label{eq:LE:objective} \\
	\textrm{subject to:} &\quad \mathbf{Y^{T}\bm{\Pi}Y}=\mathbf{I} \enspace , 
	\quad \mathbf{Y^{T}\bm{\Pi}1}=\mathbf{0} \enspace , \nonumber
\end{align}
which, when $\bm{\pi}=\mathbf{d}$, is identical to the Laplacian Eigenmaps (LE) problem \cite{belkin2003led} for computing embeddings of data that are assumed to lie on a manifold. The \emph{balance constraint} $\mathbf{Y^{T}\bm{\Pi}1}=\mathbf{0}$ is necessary to avoid eigenvectors of $\bm{\Pi}^{-1/2}\mathbf{L}\bm{\Pi}^{-1/2}$ corresponding to the trivial eigenvalue.

Turning our attention now to the more general case where $\tau>0$, if we define $\alpha_{\ell} = \left(\sum_{i}{\pi_{i}x_{i,\ell}^{2}}\right)^{-1/2}$ for $\ell = 1, \ldots , k$, and we note that $\left(\mathbf{X^{T}\bm{\Pi}X}\right)^{-1/2} = \textrm{diag}\!\left(\bm{\alpha}\right)$, we can write:
\begin{align}
	\textrm{CCB}_{\tau}\!&\left(V_{1},\ldots,V_{k}\right) 
	= \frac{1}{2}\sum_{\ell=1}^{k}{\frac{\sum_{i,j}{w_{i,j}\left(x_{i,\ell}-x_{j,\ell}\right)^{2}}}{2^{\tau/2}\!\left(\sum_{i}{\pi_{i}x_{i,\ell}^{2}}\right)^{\tau/2}}} \nonumber \\
	&= \frac{1}{2^{1+\tau/2}}\sum_{\ell=1}^{k}{\alpha^{\tau}_{\ell}\sum_{i,j}{w_{i,j}\left|x_{i,\ell}-x_{j,\ell}\right|^{\tau}}} \nonumber \\
	&= \frac{1}{2^{1+\tau/2}}\sum_{i,j}{w_{i,j}\!\left\|\left(\mathbf{X^{T}DX}\right)^{-1/2}\!\left(\mathbf{\hat{x}}_{i}-\mathbf{\hat{x}}_{j}\right)\right\|^{\tau}_{\tau}} \nonumber \\
	&= \frac{1}{2^{1+\tau/2}}\sum_{i,j}{w_{i,j}\!\left\|\mathbf{\hat{y}}_{i}-\mathbf{\hat{y}}_{j}\right\|^{\tau}_{\tau}} \enspace . \label{eq:ccNCut:relax}
\end{align}
Hence, the relaxation of (\ref{eq:ccbcutEquiv}) is obtained by dropping the condition that $\mathbf{Y} = \mathbf{X}\!\left(\mathbf{X^{T}\bm{\Pi}X}\right)^{-1/2}$ and solving the constrained minimization problem:
\begin{align}
	\min_{\mathbf{Y}\in\mathbb{R}^{n\!\times\! k}} &\quad \mathcal{J}_{\tau}\!\left(\mathbf{Y}\right) := \sum_{i=1}^{n}{\sum_{j=1}^{n}{w_{i,j}\!\left\|\mathbf{\hat{y}}_{i}-\mathbf{\hat{y}}_{j}\right\|^{\tau}_{\tau}}} \label{eq:CCBp:relaxed} \\
	\textrm{subject to:} &\quad \mathbf{Y^{T}\bm{\Pi}Y}=\mathbf{I} \enspace . \nonumber
\end{align}
Note that the balance constraint $\mathbf{Y^{T}\bm{\Pi}1}=\mathbf{0}$ is only necessary for the case $\tau=2$.

Remarkably, in the special case where $\bm{\pi} = \mathbf{d}$ and $\tau=1$, (\ref{eq:CCBp:relaxed}) is exactly the Piecewise Flat Embedding (PFE) problem \cite{yu2015pfe}, whose solution is an embedding in which the data are naturally clustered due to the promotion of sparsity in the differences between rows of $\mathbf{Y}$. It is for this reason that we choose to parametrize CCB Cuts in terms of $\tau$ and not $p$: solutions to the relaxed versions of $\textrm{CCB}_{\tau}$ for $\tau\in\left(0,1\right]$ can be expected to be piecewise constant, consistent with the sparse nature of solutions to $\ell_{\tau}$-minimization problems for $\tau\in\left(0,1\right]$.

%%%%%%%%%
\section{IRRQ Minimization Algorithm}

While (\ref{eq:CCBp:relaxed}) with $\tau = 2$ has a solution that can be written in terms of the eigenvectors corresponding to the smallest $k$ nontrivial eigenvalues of $\bm{\Pi}^{-1/2}\mathbf{L}\bm{\Pi}^{-1/2}$, no analytical form of the solution is known for $\tau\neq 2$. In \cite{yu2015pfe}, Yu et al. proposed approximating the solution to (\ref{eq:CCBp:relaxed}) with $\tau = 1$ via the Splitting Orthogonality Constraint (SOC) algorithm \cite{lai2014smo}, which requires an initial estimate of $\mathbf{Y}$ and performs a nested iteration with parameters at each level that must be tuned. While it is possible to partially generalize this approach to handle (\ref{eq:CCBp:relaxed}) with $\tau > 1$, $\mathcal{J}_{\tau}\!\left(\mathbf{Y}\right)$ is non-convex for $\tau\in\left(0,1\right)$, and so the SOC algorithm is not applicable in this regime. Furthermore, even in the regime in which it is applicable, the use of splitting in the algorithm formulation means that the solutions will not strictly satisfy the orthogonality constraint $\mathbf{Y^{T}DY}=\mathbf{I}$. 

Here, we propose an alternative algorithm for solving (\ref{eq:CCBp:relaxed}) that can be applied for \emph{all} $\tau\in\left(0,2\right]$, that \emph{does not} require an initial estimate of $\mathbf{Y}$, and that \emph{does} strictly satisfy the orthogonality constraint. Our alternative algorithm is motivated from the Iteratively Reweighted Least Squares (IRLS) algorithms commonly used to solve $\ell_{1}$-minimization problems \cite{daubechies2010irl}. In IRLS, $\ell_{1}$-minimization is performed by iteratively solving a succession of weighted least-squares ($\ell_{2}$-minimization) problems, with the weights updated at each iteration so as to decrease the impact of large residual errors. IRLS algorithms do require initialization, but it is the \emph{weights} that must be initialized as opposed to the \emph{solution}. Weights are typically initialized to unity (although they can be initialized differently by an expert user), and in specific cases \cite{daubechies2010irl}, IRLS algorithms have provable convergence guarantees.

In our relaxed problem (\ref{eq:CCBp:relaxed}), the presence of the orthogonality constraint renders IRLS algorithms invalid. However, solutions to (\ref{eq:CCBp:relaxed}) can be approximated by iteratively solving a series of \emph{constrained} weighted $\ell_{2}$-minimization problems, each of the form:
\begin{align}
	\min_{\mathbf{Y}\in\mathbb{R}^{n\!\times\! k}} &\quad \mathcal{J}^{\mathbf{\Gamma}}_{2}\!\left(\mathbf{Y}\right) := \sum_{i=1}^{n}{\sum_{j=1}^{n}{w_{i,j}\gamma_{i,j}\!\left\|\mathbf{\hat{y}}_{i}-\mathbf{\hat{y}}_{j}\right\|_{2}^{2}}} \label{eq:IRRQ:objective} \\
	\textrm{subject to:} &\quad \mathbf{Y^{T}\bm{\Pi}Y}=\mathbf{I} \enspace , 
	\quad \mathbf{Y^{T}\bm{\Pi}1}=\mathbf{0} \enspace , \nonumber
\end{align}
where $\mathbf{\Gamma}$ is the $n\times n$ matrix of weights (with entries $\gamma_{i,j}$) that is updated at each iteration in a manner similar to IRLS. 

To establish a connection between (\ref{eq:CCBp:relaxed}) and (\ref{eq:IRRQ:objective}), we first eliminate the balance constraint from (\ref{eq:IRRQ:objective}) using the result of the following Lemma, which is proved in Appendix \ref{app:proof:simpleWeights}:
\begin{lemma} \label{lemma:YBG}
	Let
\begin{align}
	\mathcal{Y} &= \left\{\mathbf{Y}\in\mathbb{R}^{n\times k}|\mathbf{Y^{T}\bm{\Pi}Y}=\mathbf{I} ,\mathbf{Y^{T}\bm{\Pi}1}=\mathbf{0}\right\} \enspace , \nonumber \\
	\mathcal{G} &= \left\{\mathbf{G}\in\mathbb{R}^{\left(n-1\right)\times k}|\mathbf{G^{T}G}=\mathbf{I}\right\} \enspace , \nonumber
\end{align}
and define $\bm{\aleph}\!\left(\mathbf{G}\right) = \bm{\Pi}^{-1/2}\mathbf{BG}$ for $\mathbf{G}\in\mathcal{G}$, where $\mathbf{B} = \mathbf{M}\!\left(\mathbf{M^{T}M}\right)^{-1/2}$, $\mathbf{M}\in\mathbb{R}^{n\times\left(n-1\right)}$ is a full rank matrix with $\textrm{null}\!\left(\mathbf{M^{T}}\right) = \textrm{span}\!\left(\mathbf{q}\right)$, and $\mathbf{q} = \bm{\Pi}^{1/2}\mathbf{1}\!/\!\left\|\bm{\Pi}^{1/2}\mathbf{1}\right\|$. Then $\bm{\aleph}$ is a bijection from $\mathcal{G}$ to $\mathcal{Y}$.
\end{lemma} 

Using this Lemma allows us to solve (\ref{eq:IRRQ:objective}) by first solving:
\begin{align} \label{eq:irrq:stepa:Ghat}
	\mathbf{\hat{G}} := \arg\min\limits_{\mathbf{G^{T}G} = \mathbf{I}} \mathcal{J}_{2}^{\mathbf{\Gamma}}\!\left(\bm{\aleph}\!\left(\mathbf{G}\right)\right) \enspace , 
\end{align}
and then computing $\mathbf{Y} = \bm{\aleph}\!\left(\mathbf{\hat{G}}\right)$.

An analogy to IRLS suggests that under the assumption that component-wise differences in the embedding do not vanish, the best choice of weights for (\ref{eq:IRRQ:objective}) would be $\gamma_{i,j} = \left\|\mathbf{\hat{y}}_{i}^{\ast}-\mathbf{\hat{y}}_{j}^{\ast}\right\|^{\tau}_{\tau}/\left\|\mathbf{\hat{y}}_{i}^{\ast}-\mathbf{\hat{y}}_{j}^{\ast}\right\|^{2}_{2}$ for $i\ne j$, where $\mathbf{Y}^{\ast}$ is the solution to (\ref{eq:CCBp:relaxed}). This choice of weights would yield $\mathcal{J}_{2}^{\mathbf{\Gamma}}\!\left(\mathbf{Y}^{\ast}\right) = \mathcal{J}_{\tau}\!\left(\mathbf{Y}^{\ast}\right)$. In practice, however, $\mathbf{Y}^{\ast}$ is unknown, and many of the component-wise differences in the embedding will vanish. Hence, we propose using regularized weights as suggested in \cite{daubechies2010irl}:
\begin{align}
	\gamma_{i,j} := \left[w_{i,j}\!\left\|\mathbf{\hat{y}}_{i}-\mathbf{\hat{y}}_{j}\right\|_{2}^{2} + \varepsilon^{2}\right]^{-1+\tau/2} \enspace , \label{eq:gamma:optimal}
\end{align}
and we update $\varepsilon$ according to the schedule prescribed by \cite{daubechies2010irl}, which suggests:
\begin{align}
	\varepsilon \leftarrow \min\!\left(\varepsilon,n^{-1}r\!\left(\mathbf{Y}\right)_{\kappa+1}\right) \enspace , \label{eq:varepsilon}
\end{align}
where $r\!\left(\mathbf{Y}\right)_{\kappa}$ is the $\kappa^{\textrm{th}}$ largest element of $\left\{w^{1/2}_{i,j}\!\left\|\mathbf{\hat{y}}_{i}-\mathbf{\hat{y}}_{j}\right\|_{2}, \forall i,j = 1, \ldots , n \right\}$. 

Combining the steps of solving (\ref{eq:IRRQ:objective}) and updating (\ref{eq:gamma:optimal})--(\ref{eq:varepsilon}) into a sequence of iterations yields Algorithm \ref{alg:irrq:main} for computing the solution to (\ref{eq:CCBp:relaxed}). Since (\ref{eq:IRRQ:objective}) is equivalent to (\ref{eq:irrq:stepa:Ghat}), and (\ref{eq:irrq:stepa:Ghat}) can be transformed into an unconstrained minimization of a Rayleigh Quotient, we term this algorithm \emph{Iteratively Reweighted Rayleigh Quotient} (IRRQ) Minimization. 

\begin{algorithm}[t]
\caption{IRRQ Algorithm for Solving (\ref{eq:CCBp:relaxed})}
\label{alg:irrq:main}
\begin{algorithmic}
\Procedure{\texttt{IRRQ}}{$\mathbf{W}$, $k$, $\tau$, $\kappa$}
	\State $\gamma_{i,j}^{\left(0\right)} := 1$, $\varepsilon_{0} := 1$, $m := 0$, $n := \textrm{size}\!\left(\mathbf{W},1\right)$
	\While{$\varepsilon_{m}>0$} 
		\State (a) {$\mathbf{Y}^{\left(m+1\right)} := \arg\min\limits_{\substack{\mathbf{Y^{T}\bm{\Pi}Y} = \mathbf{I} \\ \mathbf{Y^{T}\bm{\Pi}1} = \mathbf{0} \\ \mathbf{Y}\in\mathbb{R}^{n\times k}}} \mathcal{J}_{2}^{\mathbf{\Gamma}^{\left(m\right)}}\!\left(\mathbf{Y}\right)$}
		\State (b) {$\varepsilon_{m+1} := \min\!\left(\varepsilon_{m},n^{-1}r\!\left(\mathbf{Y}^{\left(m+1\right)}\right)_{\kappa+1}\right)$}
		\State (c) {$\gamma_{i,j}^{\left(m+1\right)} := \left[w_{i,j}\!\left\|\mathbf{\hat{y}}_{i}-\mathbf{\hat{y}}_{j}\right\|_{2}^{2} + \varepsilon_{m+1}^{2}\right]^{-1+\tau/2}$}
	\EndWhile
	\State \textbf{return} $\mathbf{Y}^{\left(m\right)}$
\EndProcedure
\end{algorithmic}
\end{algorithm}

Unlike SOC, IRRQ requires tuning of only a single hyperparameter $\kappa$, it can be used for any $\tau\in\left(0,2\right)$, and it guarantees a solution in which the orthogonality constraint is strictly enforced. Furthermore, IRRQ does not require a preprocessing step to estimate an initial clustering; rather, only the weights $\gamma_{i,j}$ must be initialized, and they can all be initialized to unity. Interestingly, this initialization is equivalent to implicitly using an initial clustering that corresponds to the solution of the relaxed NCut problem. This is because $\mathcal{J}_{2}^{\mathbf{11^{T}}}\!\left(\mathbf{Y}\right)$ is equivalent to the LE objective function $\mathcal{J}_{2}\!\left(\mathbf{Y}\right)$ in (\ref{eq:LE:objective}). If different initializations are desired, for instance, by computing an initial embedding $\mathbf{Y}^{\left(0\right)}$ using a Gaussian Mixture Model as in \cite{yu2015pfe}, these can be incorporated by setting the initial weights to be $\gamma_{i,j}^{\left(0\right)} = \left\|\mathbf{\hat{y}}_{i}^{\left(0\right)}-\mathbf{\hat{y}}_{j}^{\left(0\right)}\right\|_{\tau}^{\tau}/\left\|\mathbf{\hat{y}}_{i}^{\left(0\right)}-\mathbf{\hat{y}}_{j}^{\left(0\right)}\right\|_{2}^{2}$ or $\gamma_{i,j}^{\left(0\right)} = \left[w_{i,j}\!\left\|\mathbf{\hat{y}}_{i}^{\left(0\right)}-\mathbf{\hat{y}}_{j}^{\left(0\right)}\right\|_{2}^{2}+\varepsilon_{0}^{2}\right]^{-1+\tau/2}$.

\subsection*{Solving IRRQ Step (a)}

Computing $\mathbf{Y}^{\left(m+1\right)}$ in step (a) of the IRRQ minimization algorithm is nontrivial. From the relationship between (\ref{eq:IRRQ:objective})--(\ref{eq:irrq:stepa:Ghat}), we can see that computing $\mathbf{Y}^{\left(m+1\right)}$ is equivalent to solving
\begin{align} \label{eq:IRRQ:GhatIter}
	\mathbf{\hat{G}}^{\left(m+1\right)} := \arg\min\limits_{\mathbf{G^{T}G} = \mathbf{I}} \mathcal{J}_{2}^{\mathbf{\Gamma}^{\left(m\right)}}\!\left(\bm{\aleph}\!\left(\mathbf{G}\right)\right) \enspace , 
\end{align}
and then computing $\mathbf{Y}^{\left(m+1\right)} = \bm{\aleph}\!\left(\mathbf{\hat{G}}^{\left(m+1\right)}\right)$. Problem (\ref{eq:IRRQ:GhatIter}) can be expressed as the Rayleigh quotient minimization:
\begin{align} \label{eq:IRRQ:GhatRQ}
	\mathbf{\hat{G}}^{\left(m+1\right)} &:= \arg\min \textrm{tr}\!\left(\mathbf{G^{T}B^{T}\bm{\Pi}}^{-1/2}\mathbf{\hat{L}}^{\left(m\right)}\mathbf{\bm{\Pi}}^{-1/2}\mathbf{BG}\right. \nonumber \\
	&\quad\quad\quad\quad\quad\quad\quad\quad \cdot\left.\left(\mathbf{G^{T}G}\right)^{-1}\right) \enspace ,
\end{align}
where $\mathbf{\hat{L}}^{\left(m\right)}$ is the Laplacian of the graph having weight matrix $\mathbf{W}\!\odot\!\mathbf{\Gamma}^{\left(m\right)}$ and $\odot$ denotes Hadamard product. The solution to (\ref{eq:IRRQ:GhatRQ}) is given by $\mathbf{\hat{G}}^{\left(m+1\right)} = \mathbf{UH}$, where $\mathbf{U}\in\mathbb{R}^{\left(n-1\right)\!\times k}$ is the matrix whose columns are the orthonormal eigenvectors $\mathbf{u}_{1}$, $\ldots$, $\mathbf{u}_{k}$ corresponding to the smallest eigenvalues of $\mathbf{B^{T}\bm{\Pi}}^{-1/2}\mathbf{\hat{L}}^{\left(m\right)}\bm{\Pi}^{-1/2}\mathbf{B}$, and $\mathbf{H}\in\mathbb{R}^{k\times k}$ is an arbitrary orthogonal matrix. (Note that by eliminating the balance constraint, we also eliminate the possibility of a trivial eigenvalue of $\mathbf{B^{T}}\bm{\Pi}^{-1/2}\mathbf{\hat{L}}^{\left(m\right)}\bm{\Pi}^{-1/2}\mathbf{B}$. Such an eigenvalue would have eigenvector $\mathbf{p}$ for which $\mathbf{Bp}$ is in the direction of $\mathbf{q}$; however, this is contradicted by the fact that $\textrm{range}\!\left(\mathbf{B}\right) = \textrm{range}\!\left(\mathbf{M^{T}}\right)$.)

The ability of this solution to scale to large $n$ depends critically on the structure of $\mathbf{M}$, and on whether or not $\mathbf{B}$ must explicitly be constructed. Since we are free to choose \emph{any} full-rank $n\!\times\!\left(n-1\right)$ matrix $\mathbf{M}$ such that $\mathbf{M^{T}q}=\mathbf{0}$, we make the particular choice $\mathbf{M^{T}} = \left[\mathbf{\hat{q}}\,|-q_{1}\mathbf{I}_{n-1} \right]$, where $\mathbf{\hat{q}} = \left[q_{2},q_{3},\ldots ,q_{n} \right]^{\mathbf{T}}$ and $\mathbf{I}_{n-1}$ is the $\left(n-1\right)\!\times\!\left(n-1\right)$ identity matrix. This particular choice of $\mathbf{M}$ is sparse, and therefore, as shown in Appendix \ref{app:proof:BTLB}, the multiplication of an arbitrary vector by $\mathbf{B^{T}}\bm{\Pi}^{-1/2}\mathbf{\hat{L}}^{\left(m\right)}{\bm{\Pi}}^{-1/2}\mathbf{B}$ can be performed efficiently without explicitly constructing $\mathbf{B}$.

A final note is that the solution to step (a) is not actually unique: $\mathbf{Y}^{\left(m+1\right)}$ can be postmultiplied by $\mathbf{H^{T}}$ and still be a valid solution (recalling that $\mathbf{H}$ is orthogonal). This is not a problem for steps (b) and (c) of Algorithm \ref{alg:irrq:main}, because $r$ and $\gamma_{i,j}$ are invariant to such transformations of $\mathbf{Y}^{\left(m+1\right)}$. As a consequence, IRRQ minimization could yield an entire family of solutions to the PFE problem. This could be problematic because the $\ell_{\tau}$-norms/pseudo-norms are not invariant under orthogonal transformations for $\tau\neq 2$. In practice, however, we have found that the $\ell_{\tau}$-norms/pseudo-norms are minimized for the choice $\mathbf{H}=\mathbf{I}$ and we therefore suggest this choice. Proof that this is the best choice remains an open problem.

\subsection*{Choosing $\kappa$ for Rapid Convergence}

In IRLS algorithms for $\ell_{\tau}$-minimization, linear convergence can typically be achieved for $\tau\geq 1$ and super-linear convergence for $\tau\in\left(0,1\right)$ if $\kappa$ is chosen large enough so that if the resulting solution is $\theta$-sparse, then $\kappa>\theta$. (Actually, there are much more sophisticated convergence results in \cite{daubechies2010irl}, but this is a good rule-of-thumb.) Without attempting to prove convergence results for the IRRQ algorithm, we use a similar strategy in choosing $\kappa$. In practice, the main difficulty in choosing $\kappa$ is that $\theta$ is not known exactly until the problem is solved. To approximate $\theta$, we use an estimate $\hat{\theta}$ equal to twice the number of graph edges that connect different clusters from a $k$-means clustering performed on the initial embedding $\mathbf{Y}^{\left(0\right)}$. For ``scale-free" behavior, we introduce the hyperparameter $\tilde{\kappa} \in \left(0,1\right)$ that can then be mapped to $\kappa$ by $\kappa = \hat{\theta} + \tilde{\kappa}\!\left(2\left|E\right|-\hat{\theta}\right)$, where $\left|E\right|$ is the total number of edges in the graph.  

%%%%%%%%%
\section{CCB Cuts for Segmentation}

\begin{figure*}[t]
	\begin{minipage}[t]{\linewidth}
	\begin{center}
		\begin{tabular}{P{0.58in} P{0.58in} P{0.58in} P{0.58in} P{0.58in} P{0.58in} P{0.58in} P{0.58in} P{0.58in}}
				\includegraphics[width=0.7in]{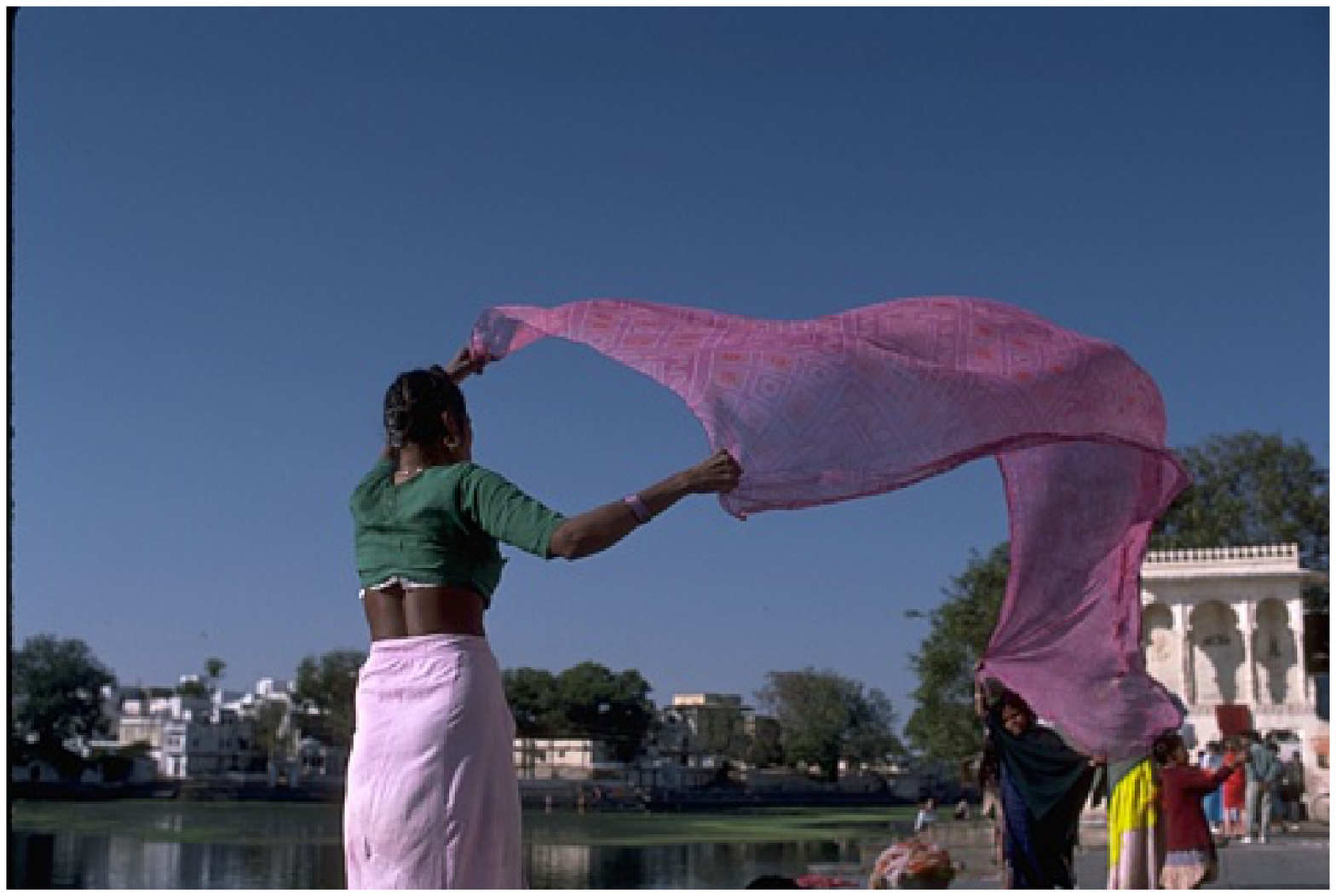} &
				\includegraphics[width=0.7in]{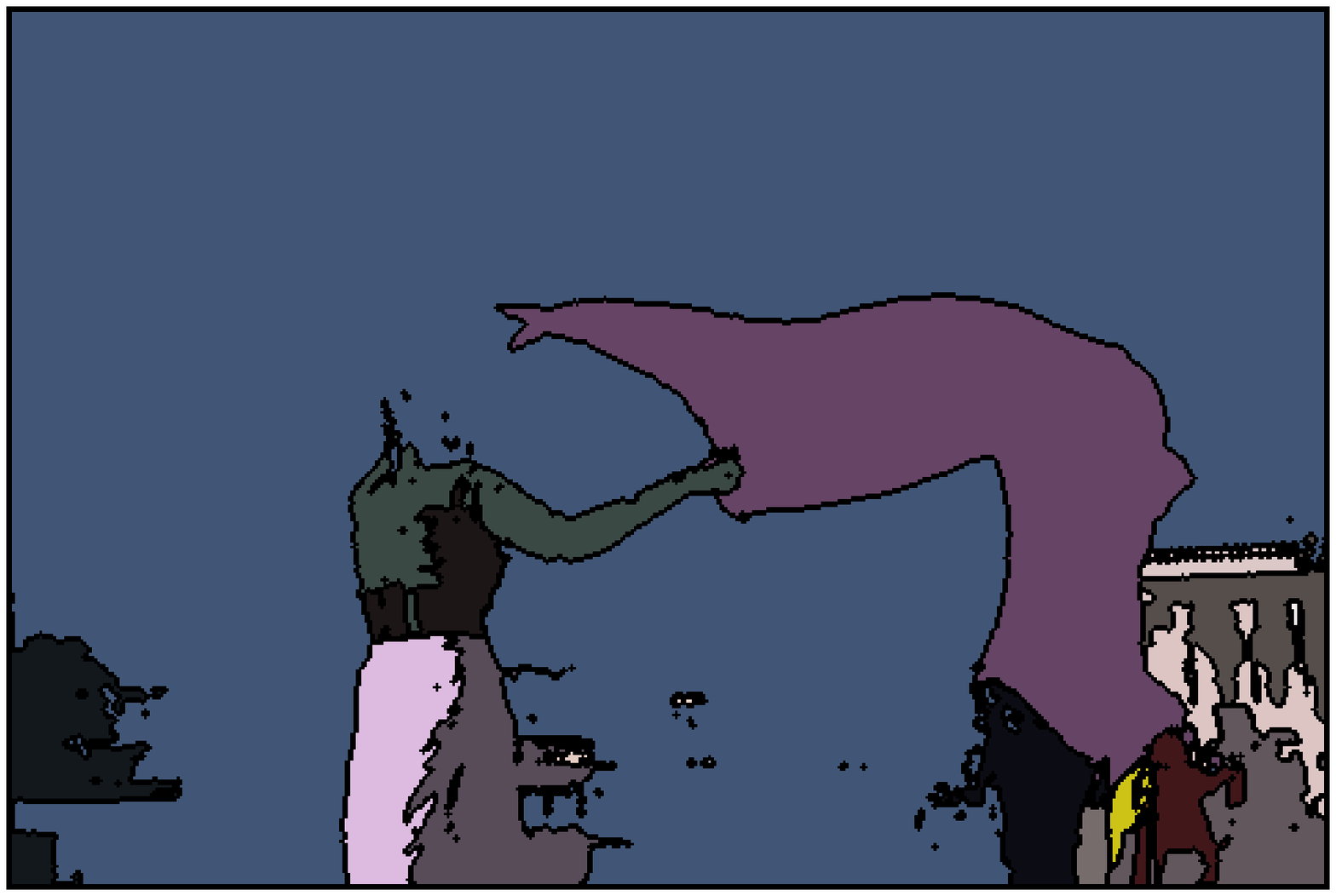} &
				\includegraphics[width=0.7in]{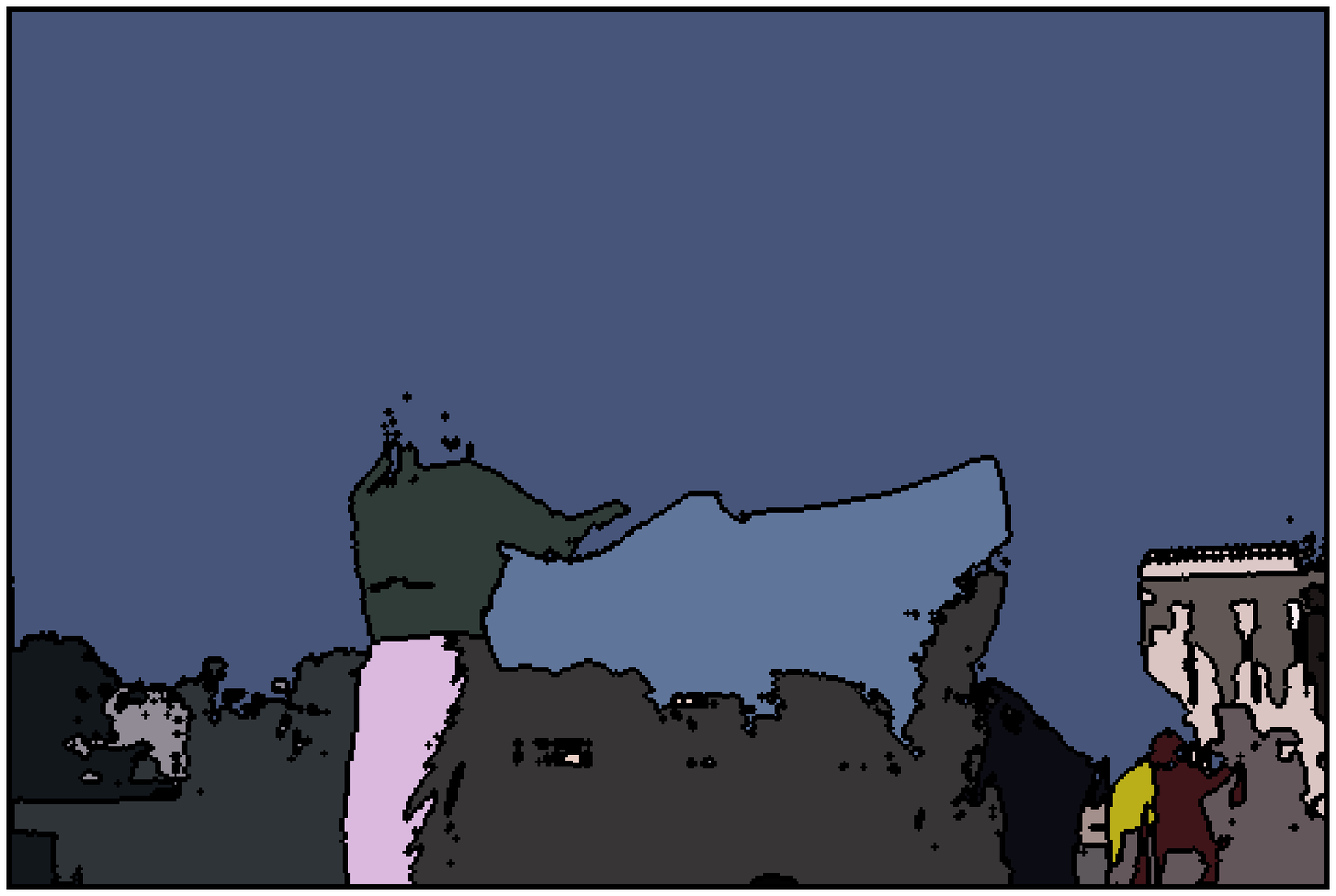} &
				\includegraphics[width=0.7in]{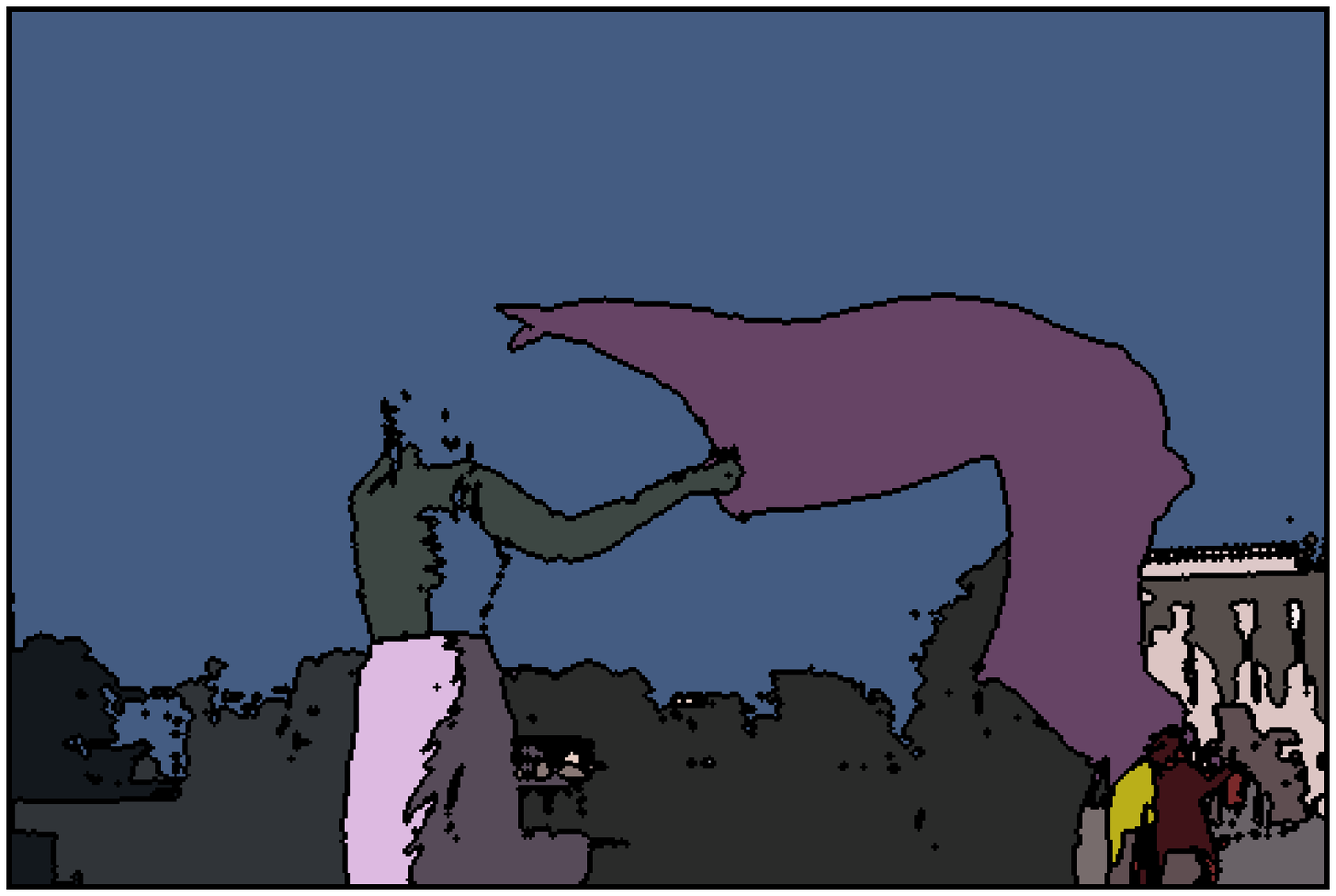} &
				\includegraphics[width=0.7in]{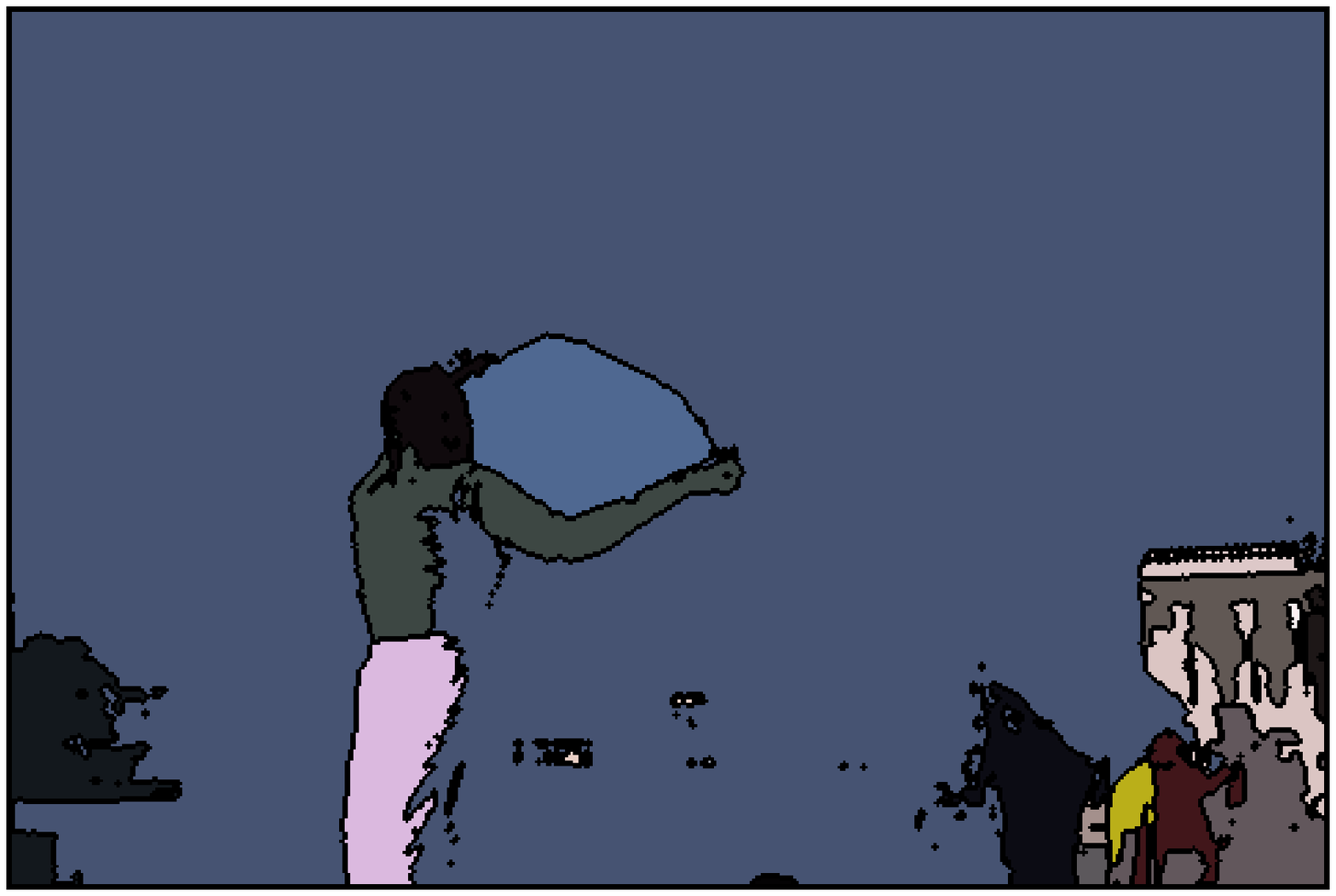} &
				\includegraphics[width=0.7in]{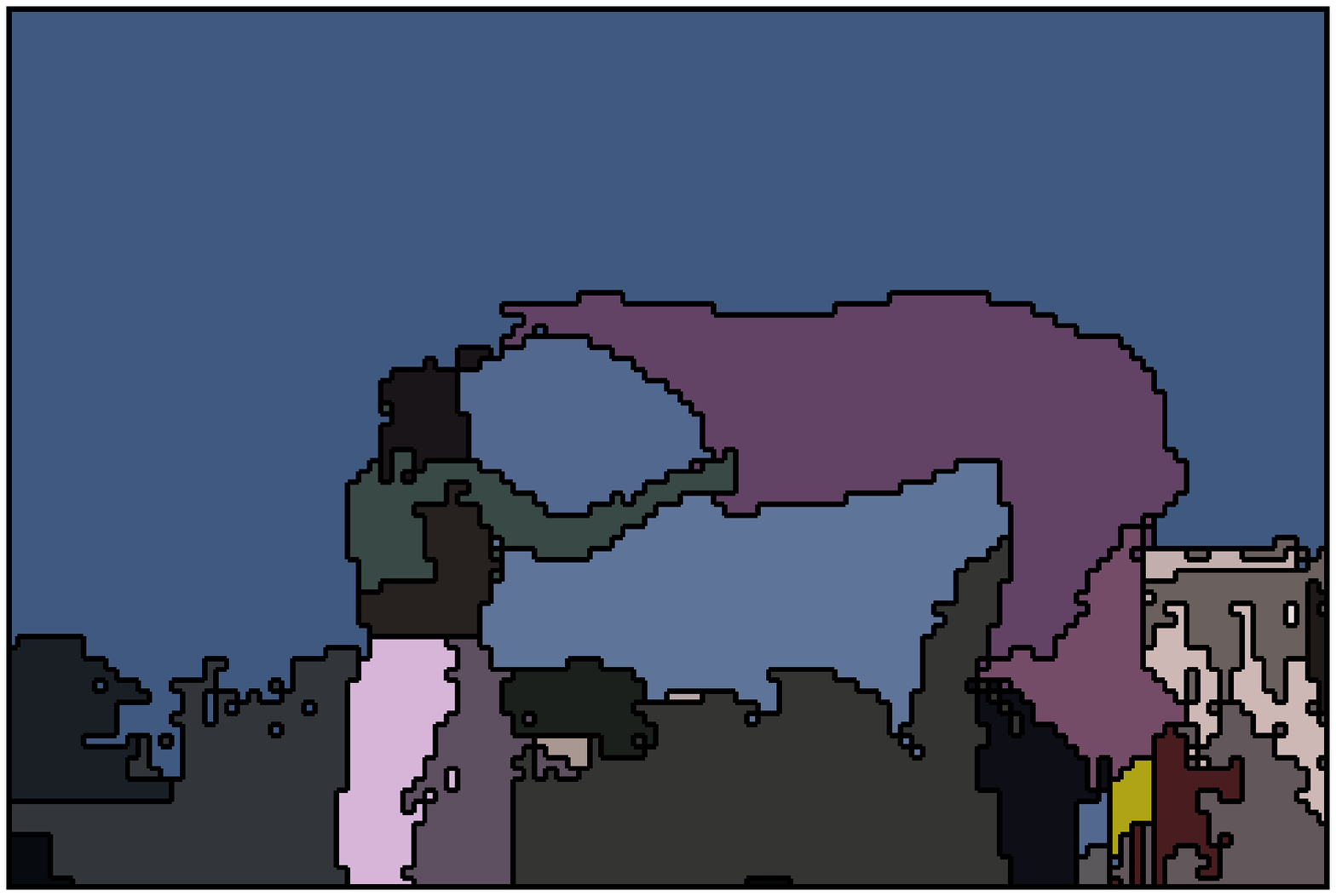} &
				\includegraphics[width=0.7in]{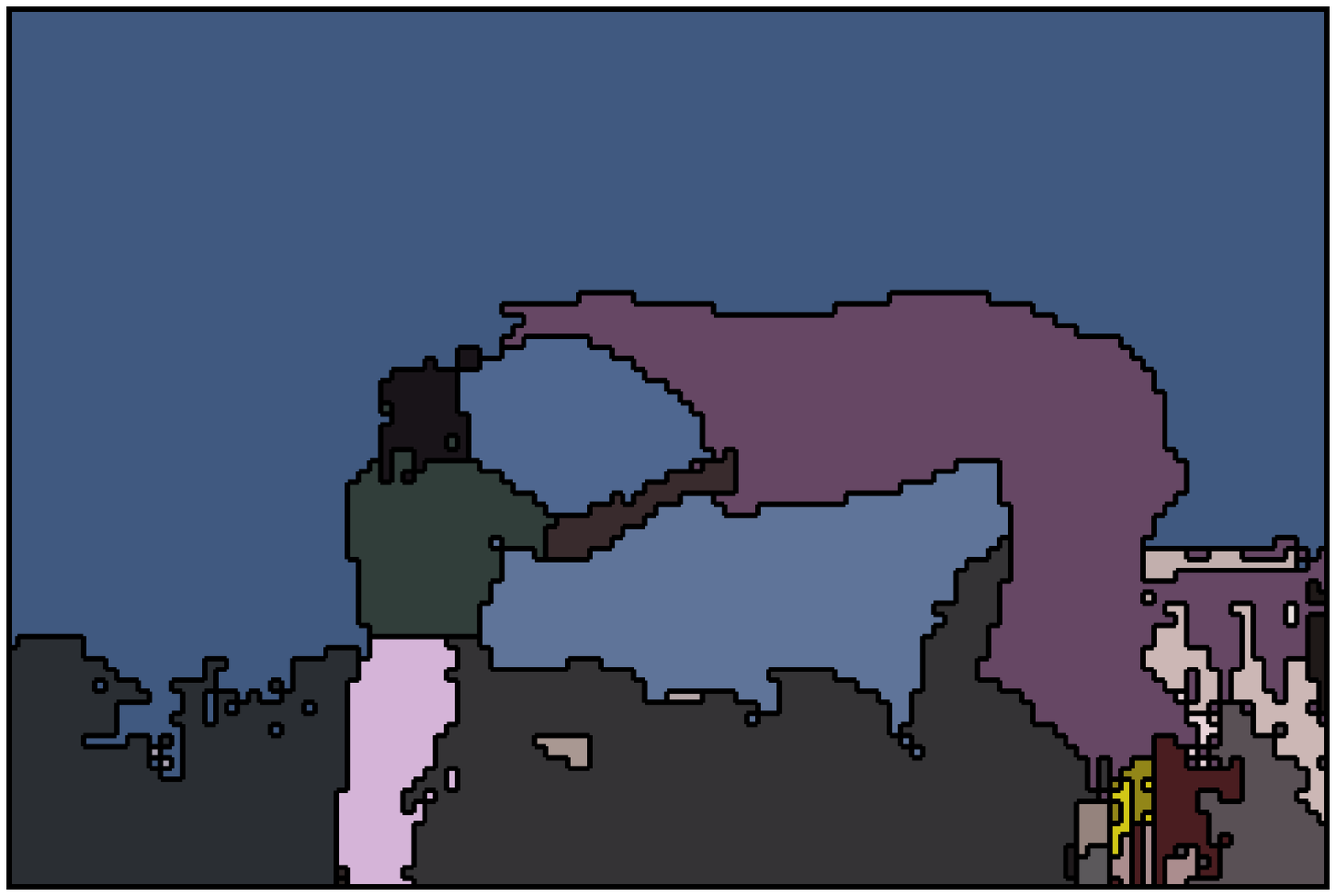} &
				\includegraphics[width=0.7in]{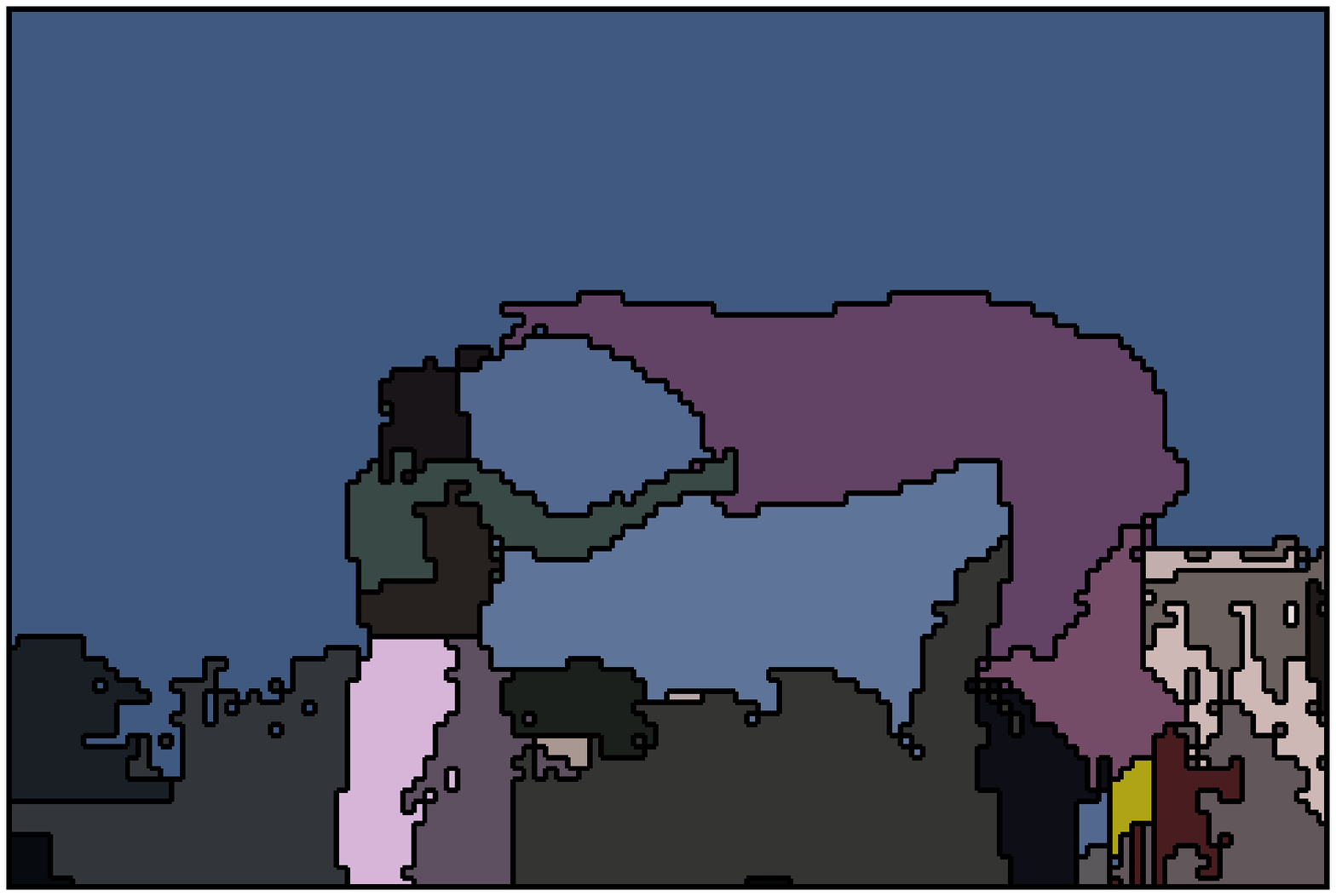} &
				\includegraphics[width=0.7in]{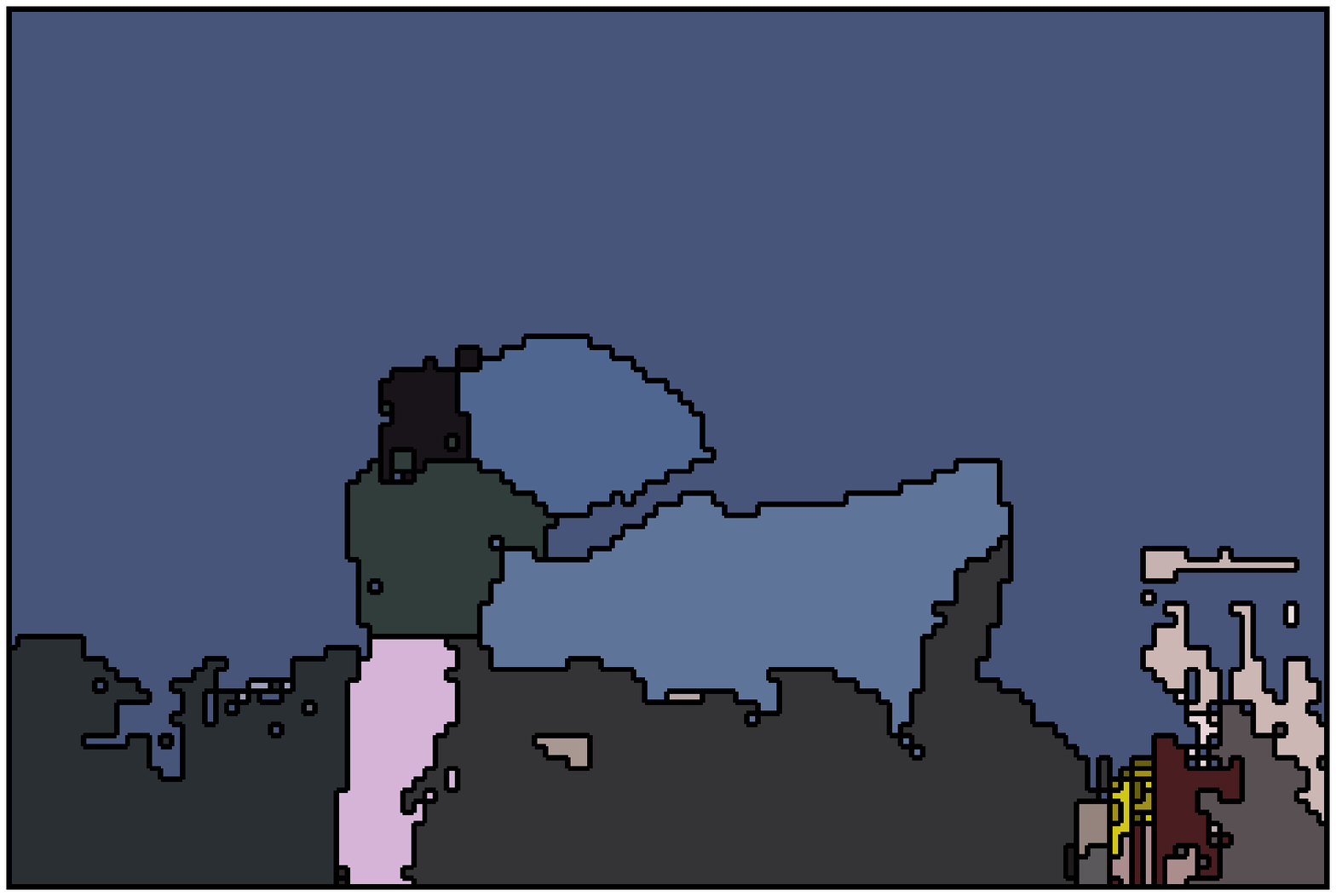} \\				
				&
				\includegraphics[width=0.7in]{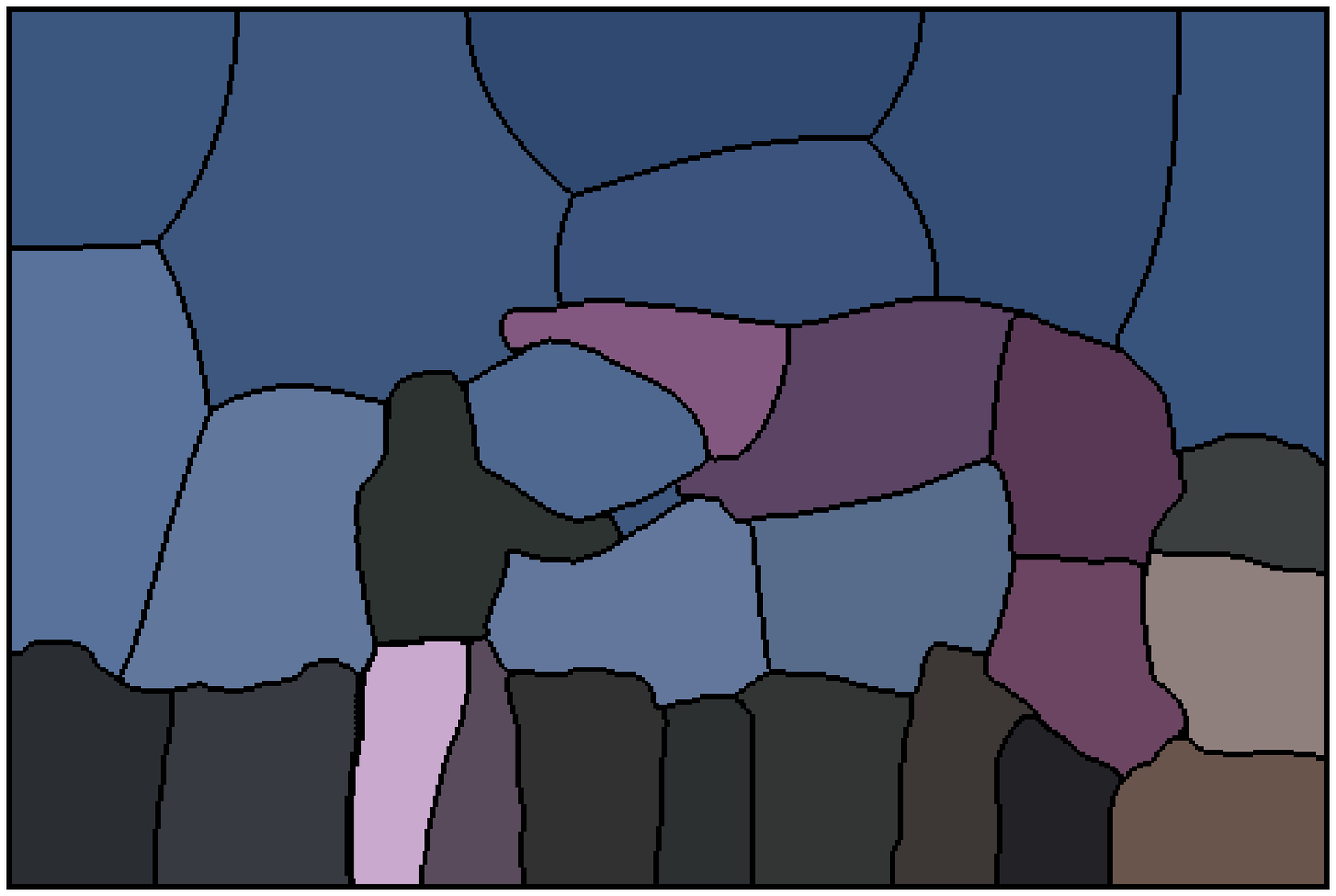} &
				\includegraphics[width=0.7in]{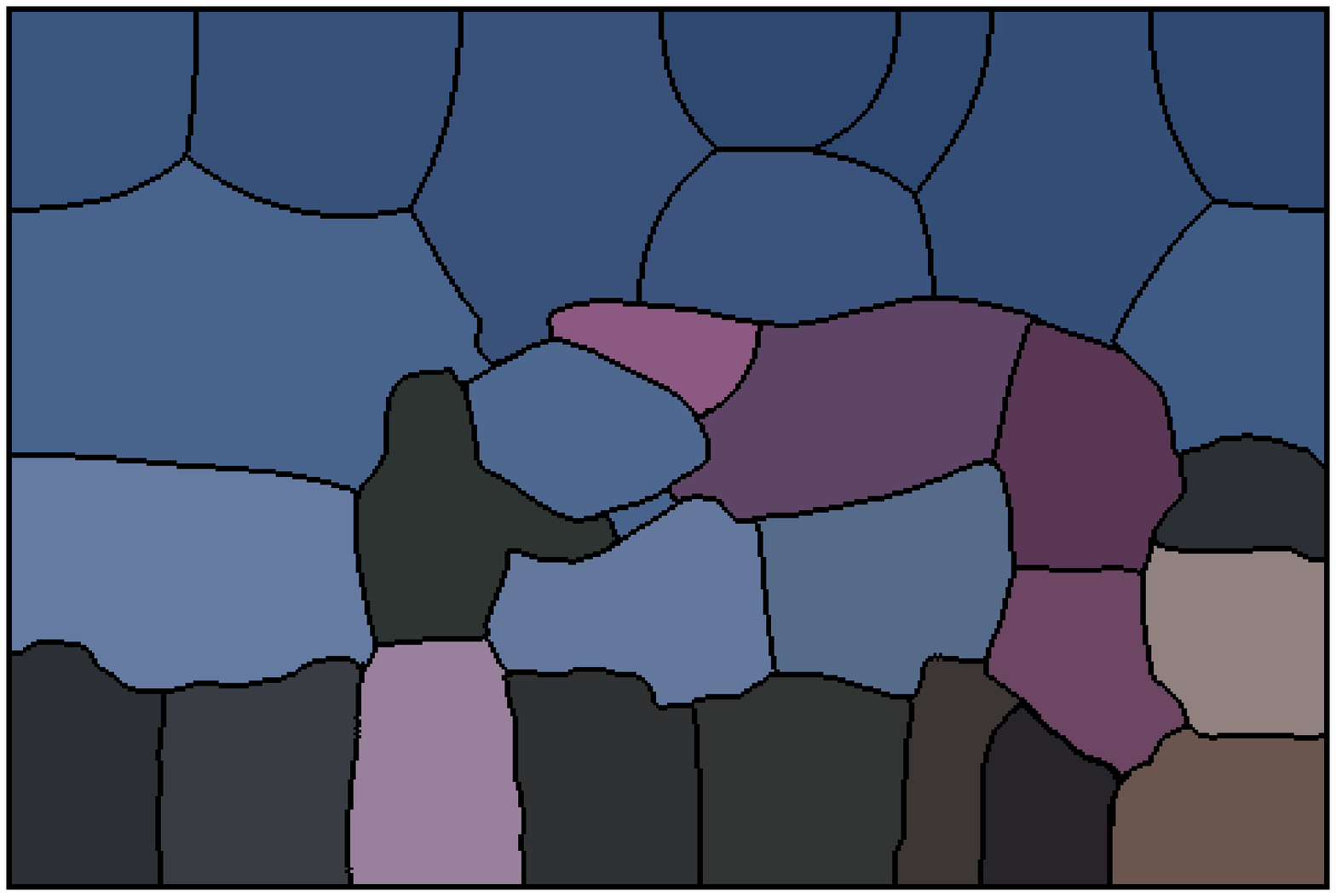} &
				\includegraphics[width=0.7in]{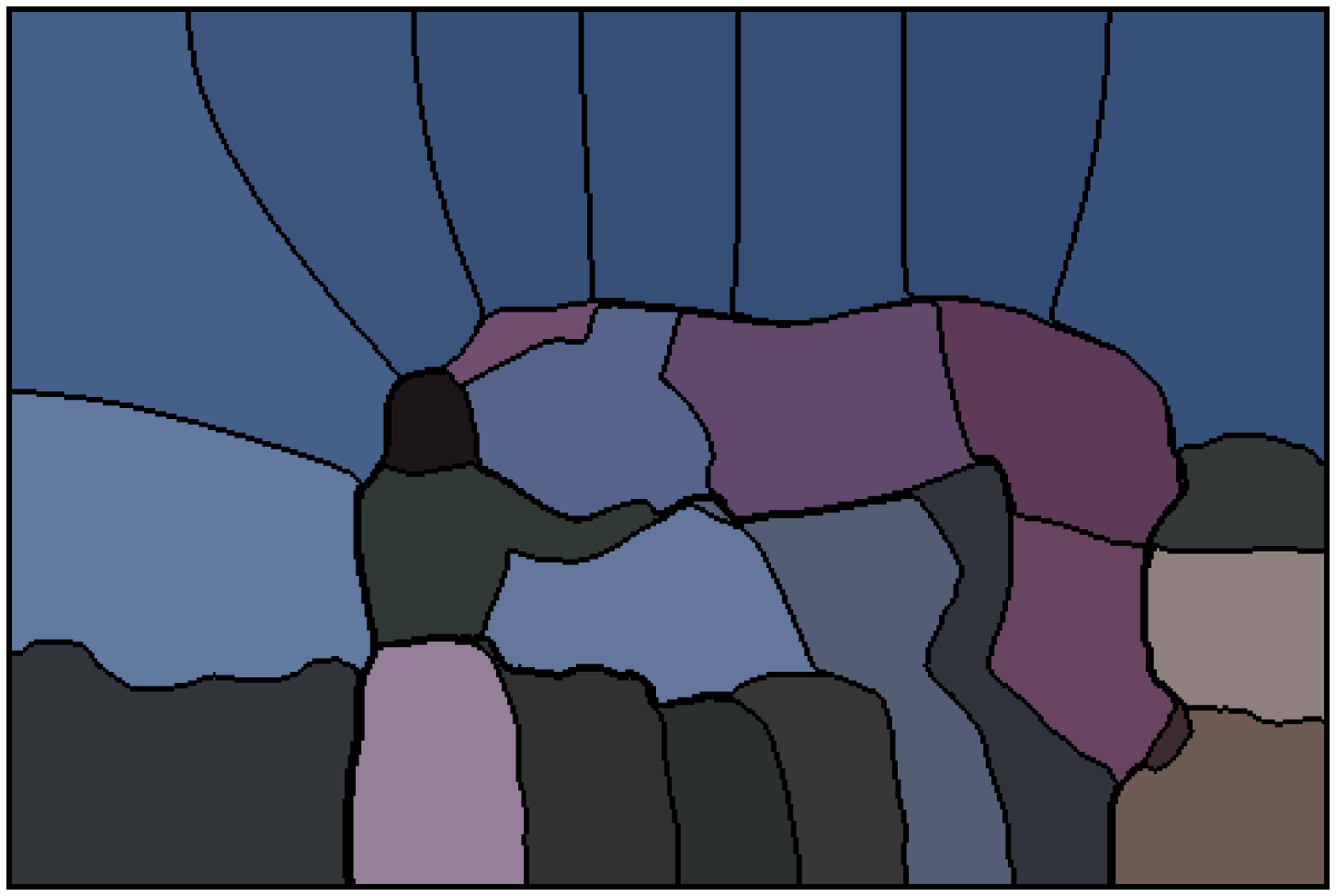} &
				\includegraphics[width=0.7in]{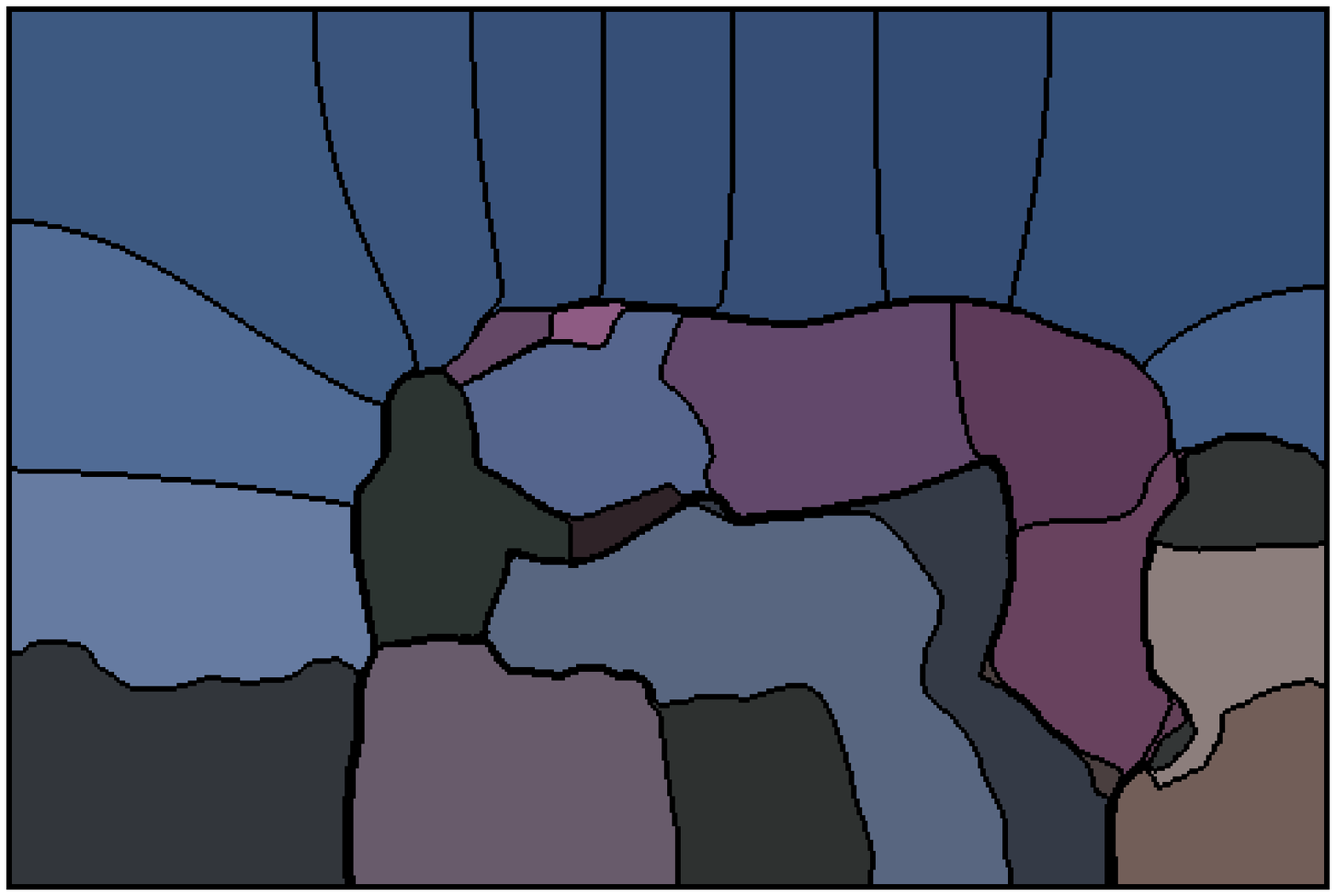} &
				\includegraphics[width=0.7in]{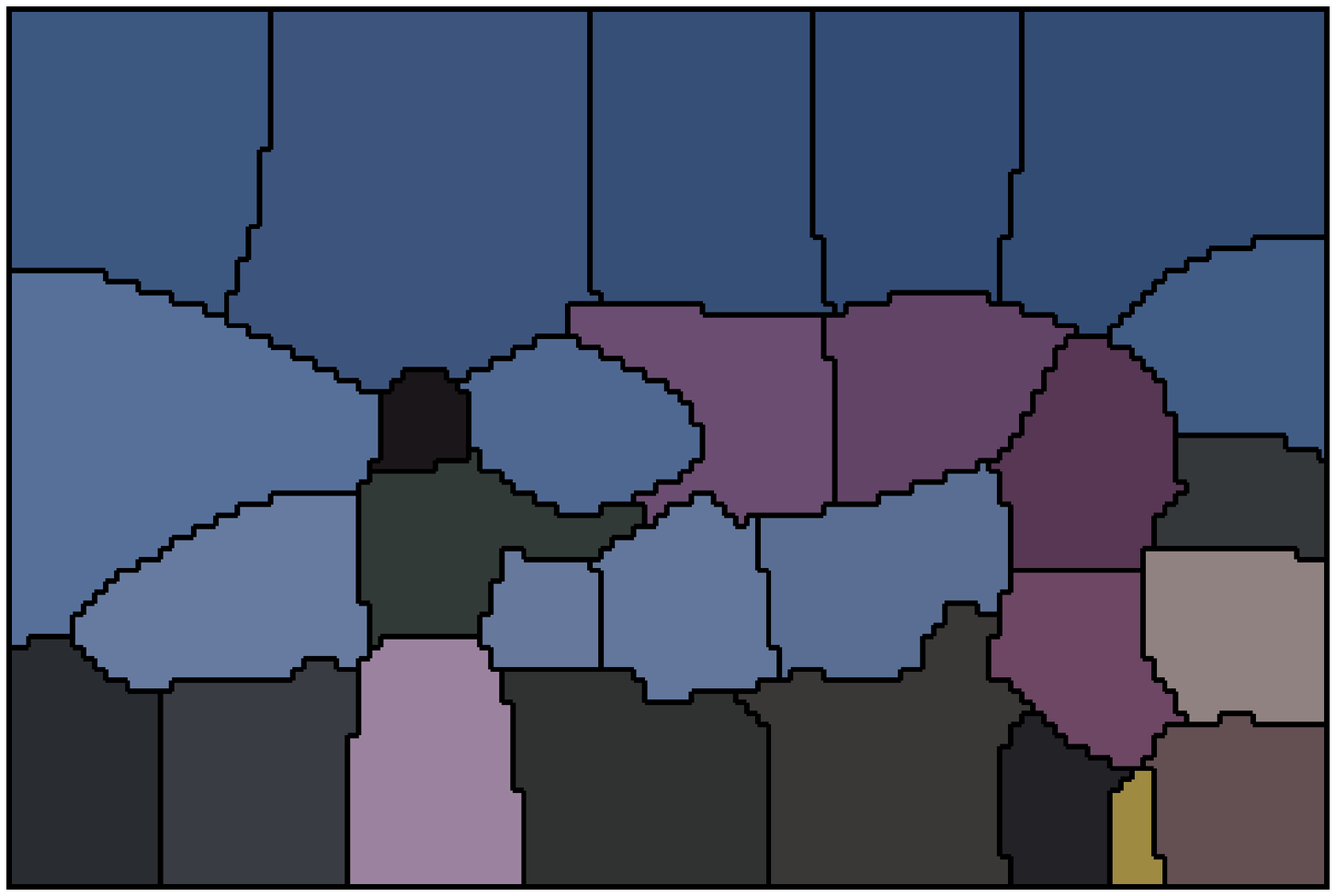} &
				\includegraphics[width=0.7in]{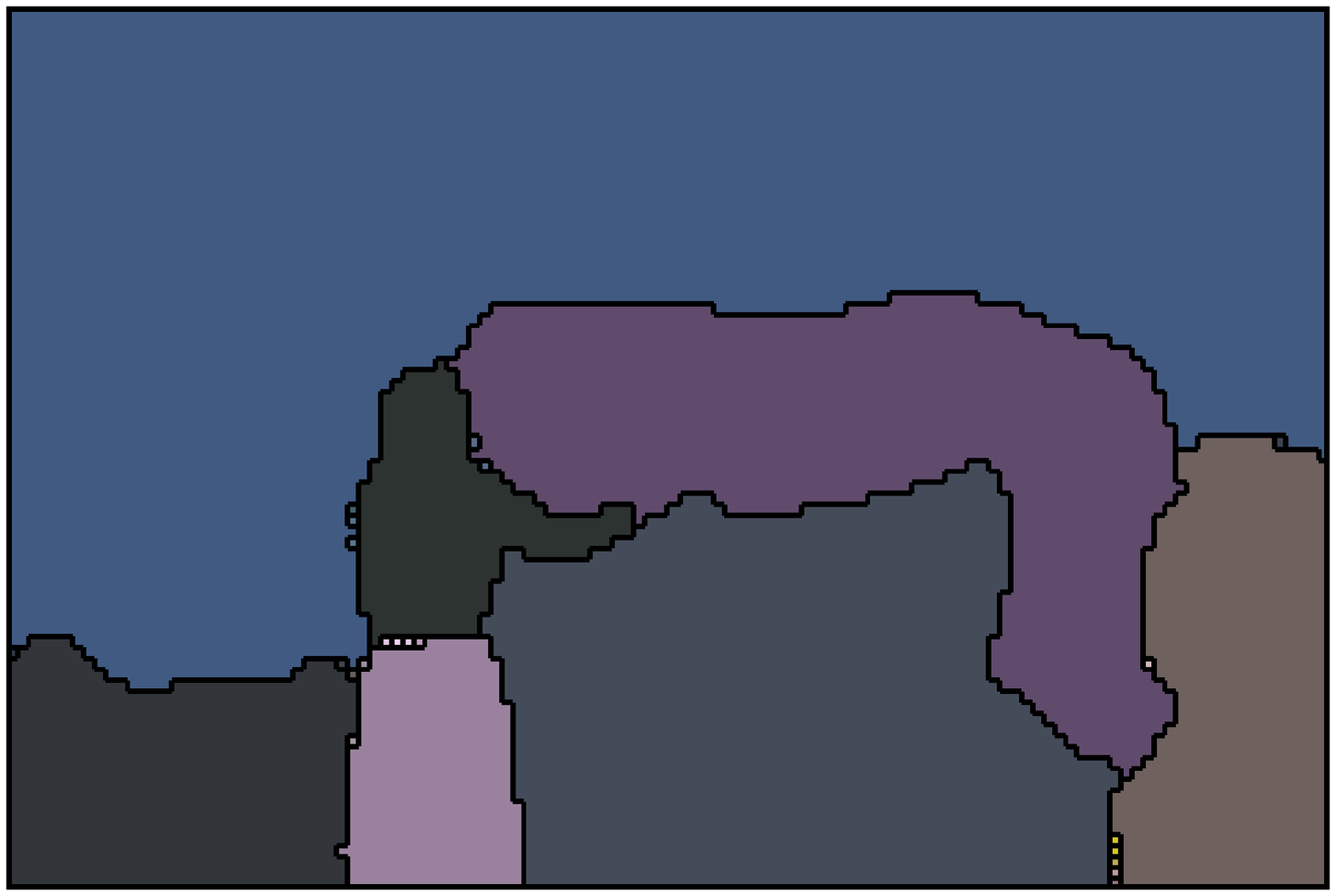} &
				\includegraphics[width=0.7in]{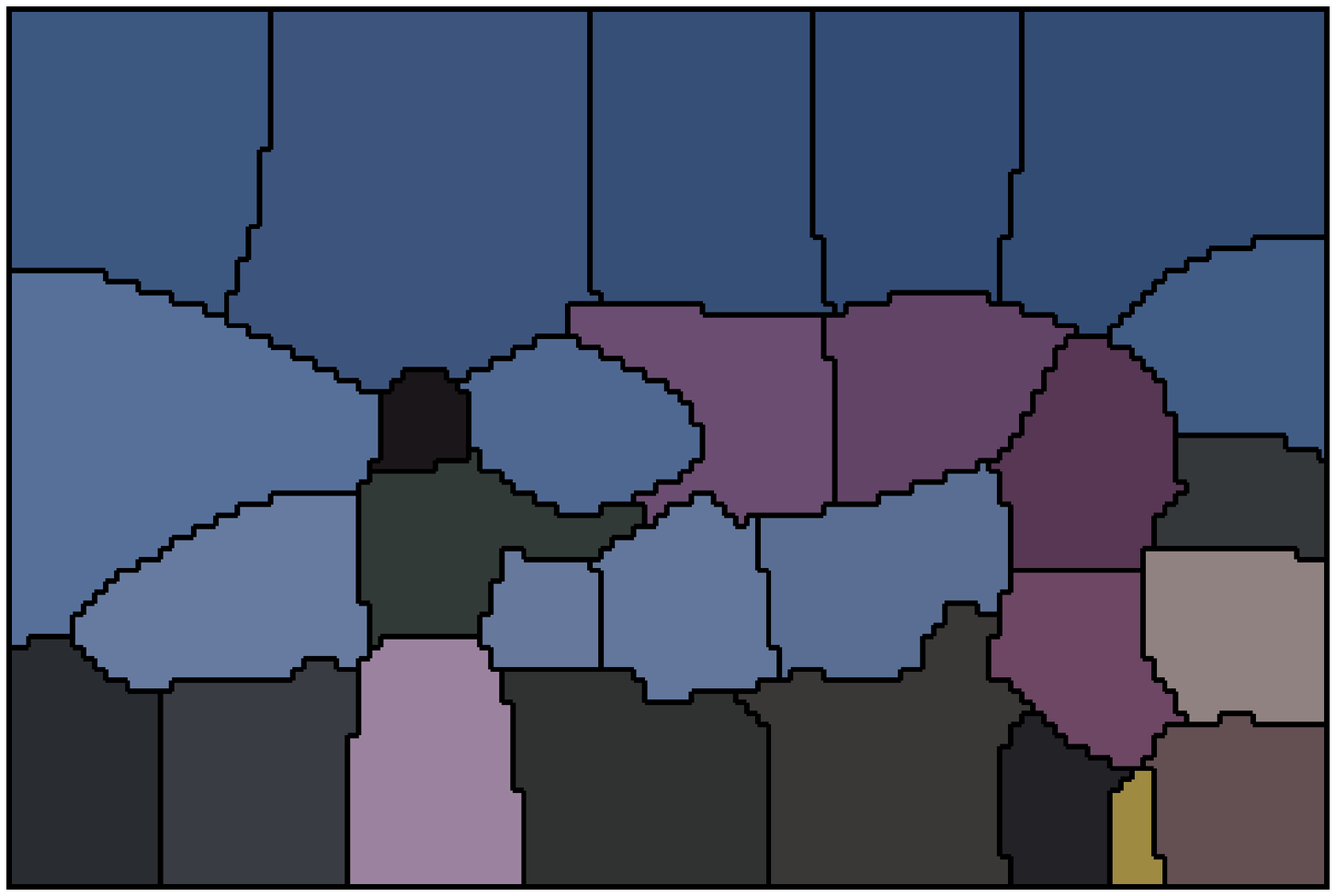} &
				\includegraphics[width=0.7in]{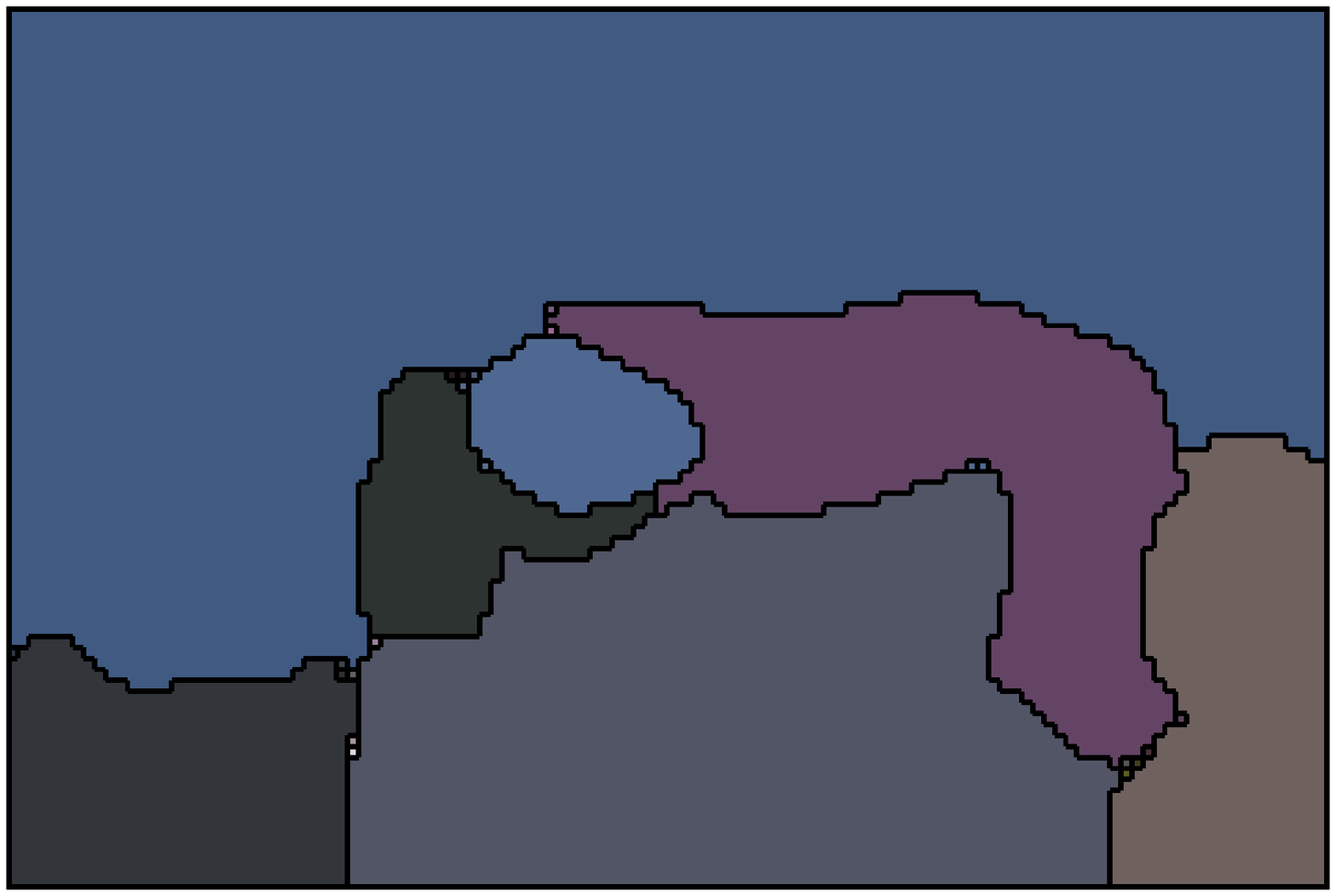} \\				
				&
				\includegraphics[width=0.7in]{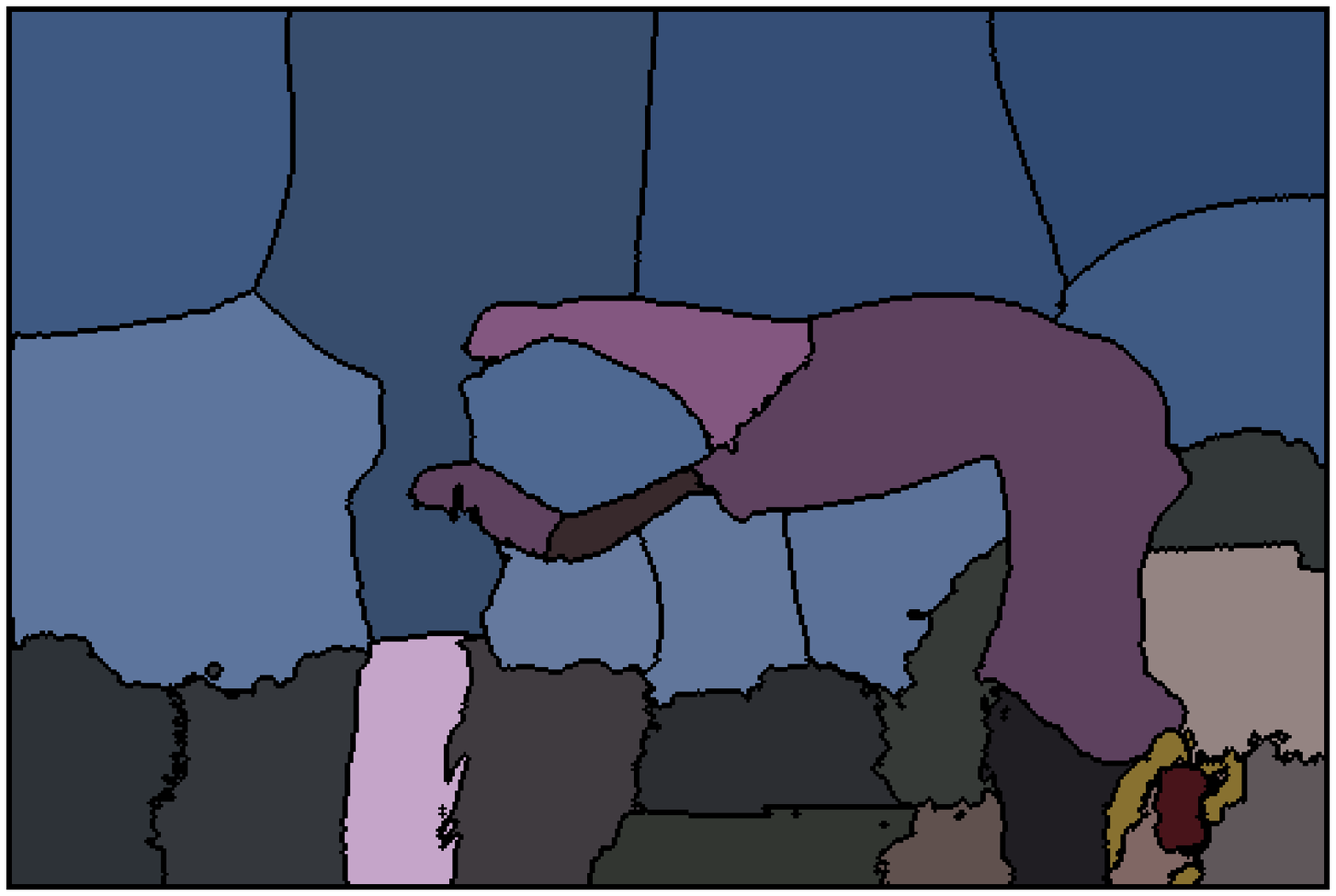} &
				\includegraphics[width=0.7in]{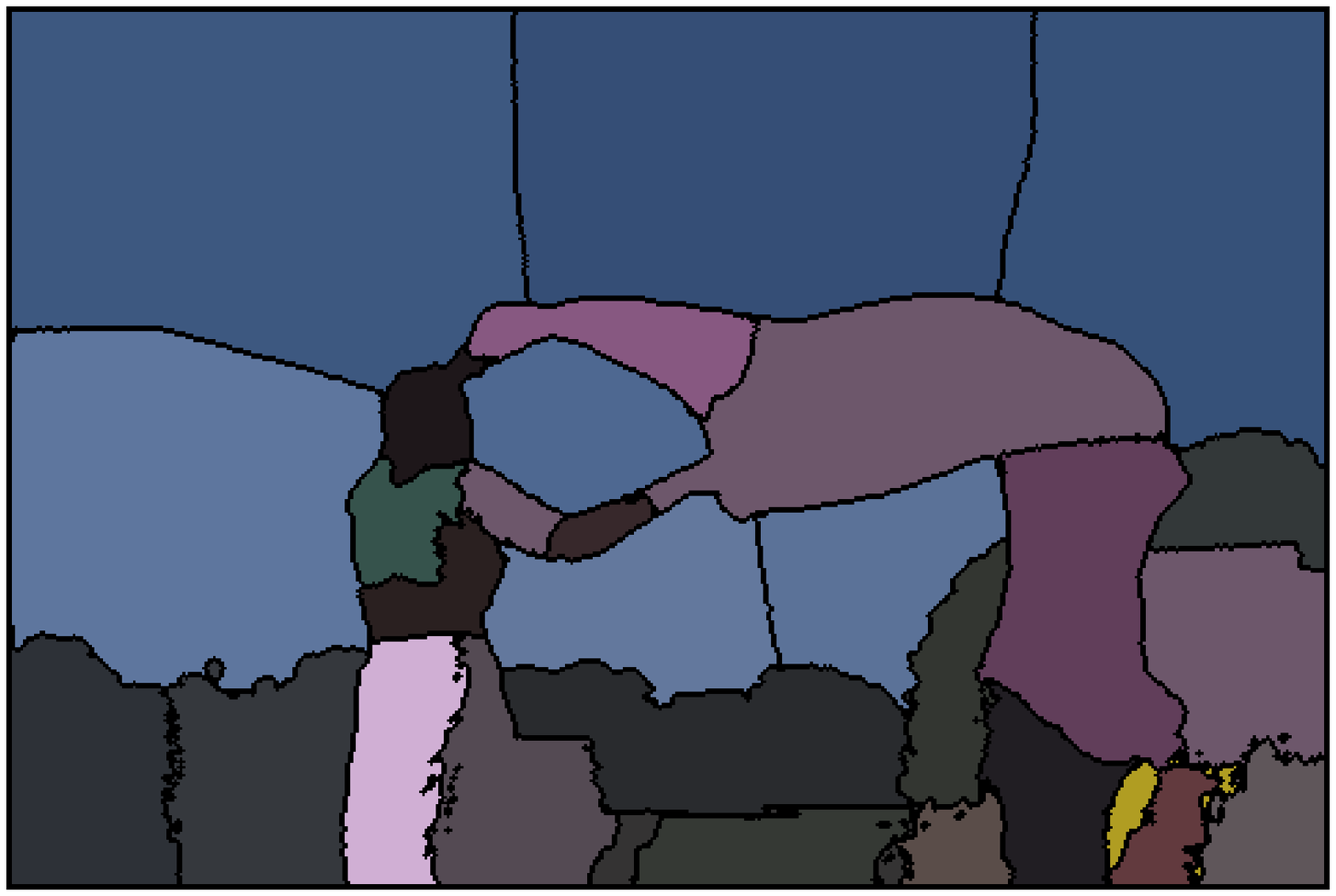} &
				\includegraphics[width=0.7in]{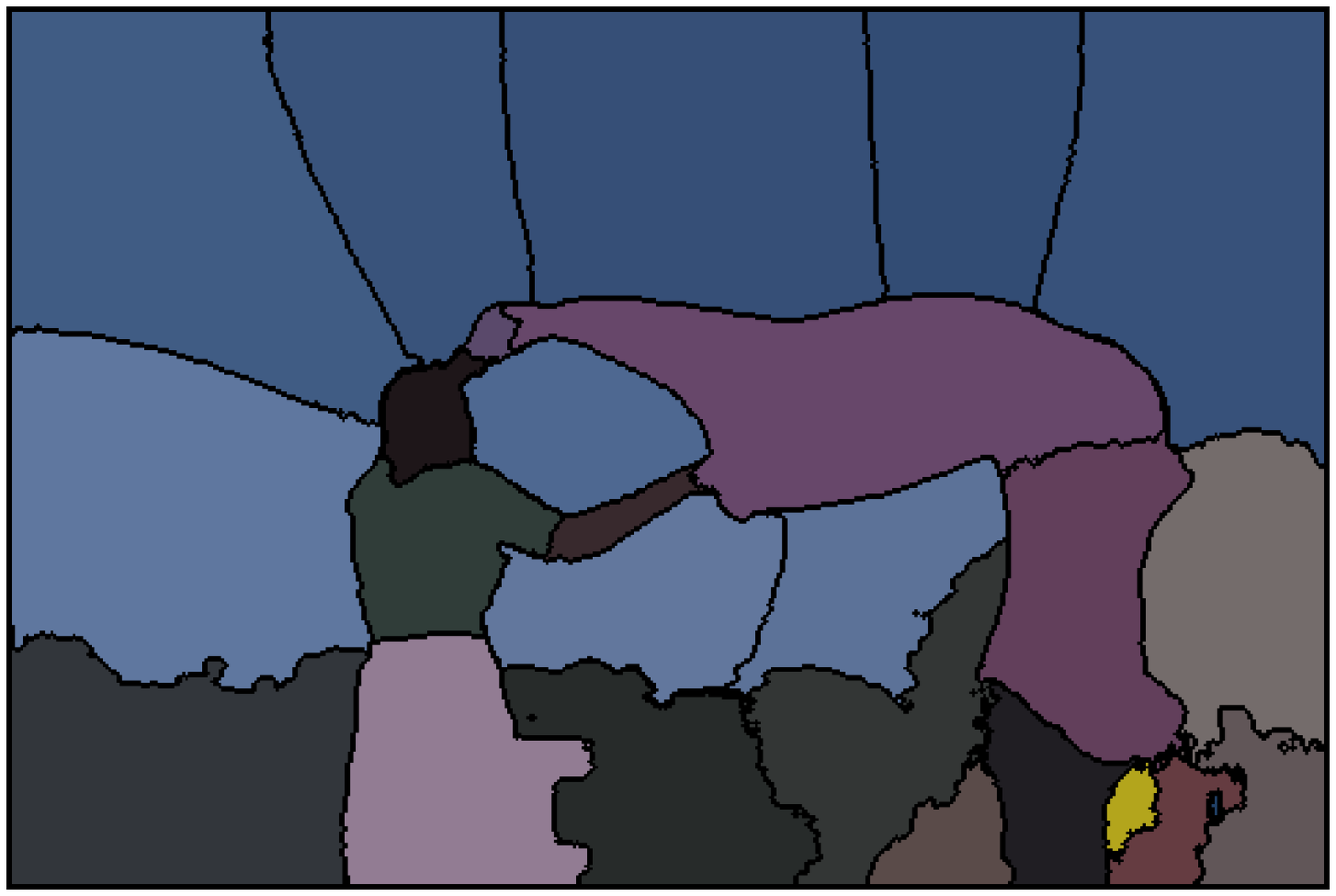} &
				\includegraphics[width=0.7in]{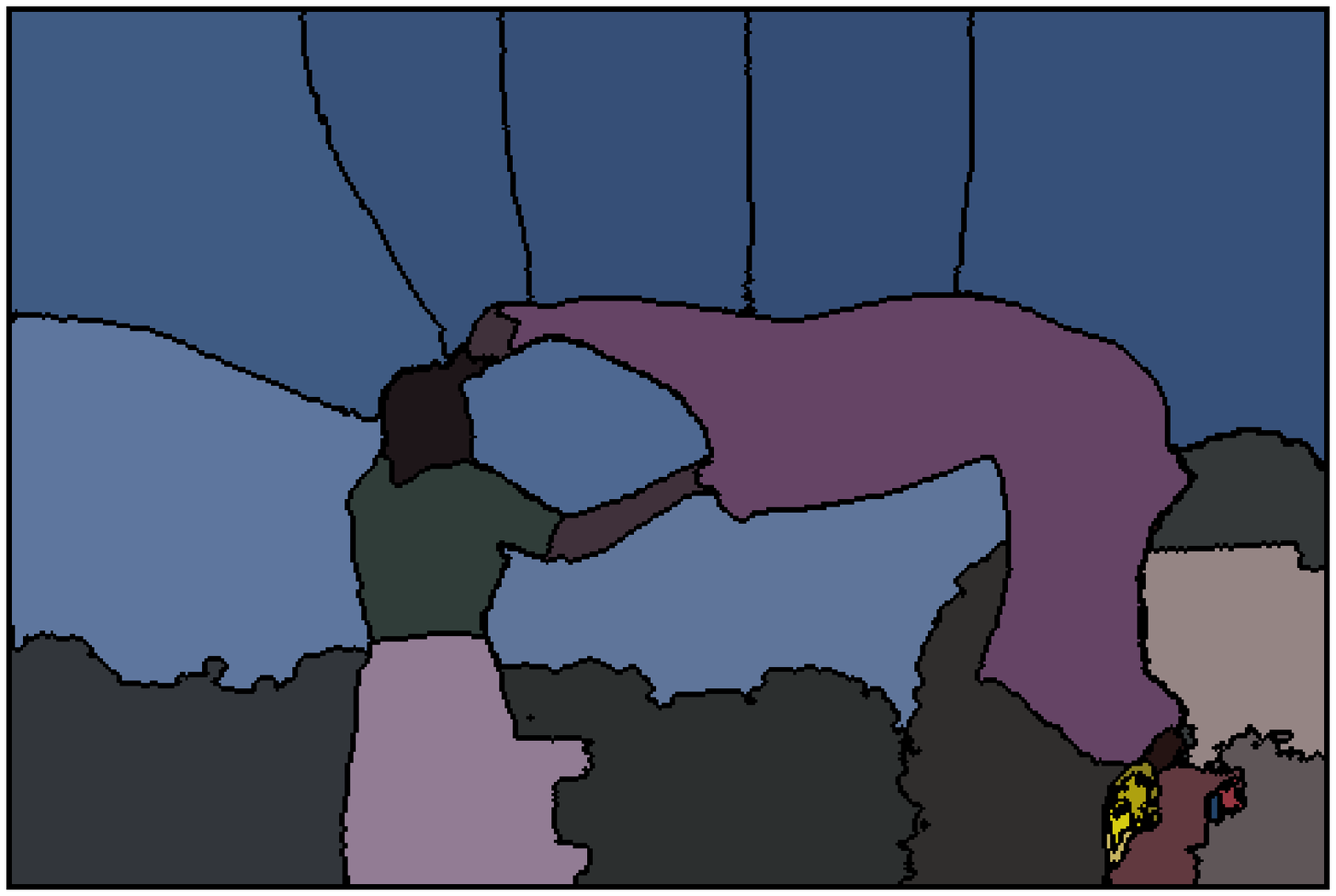} &
				\includegraphics[width=0.7in]{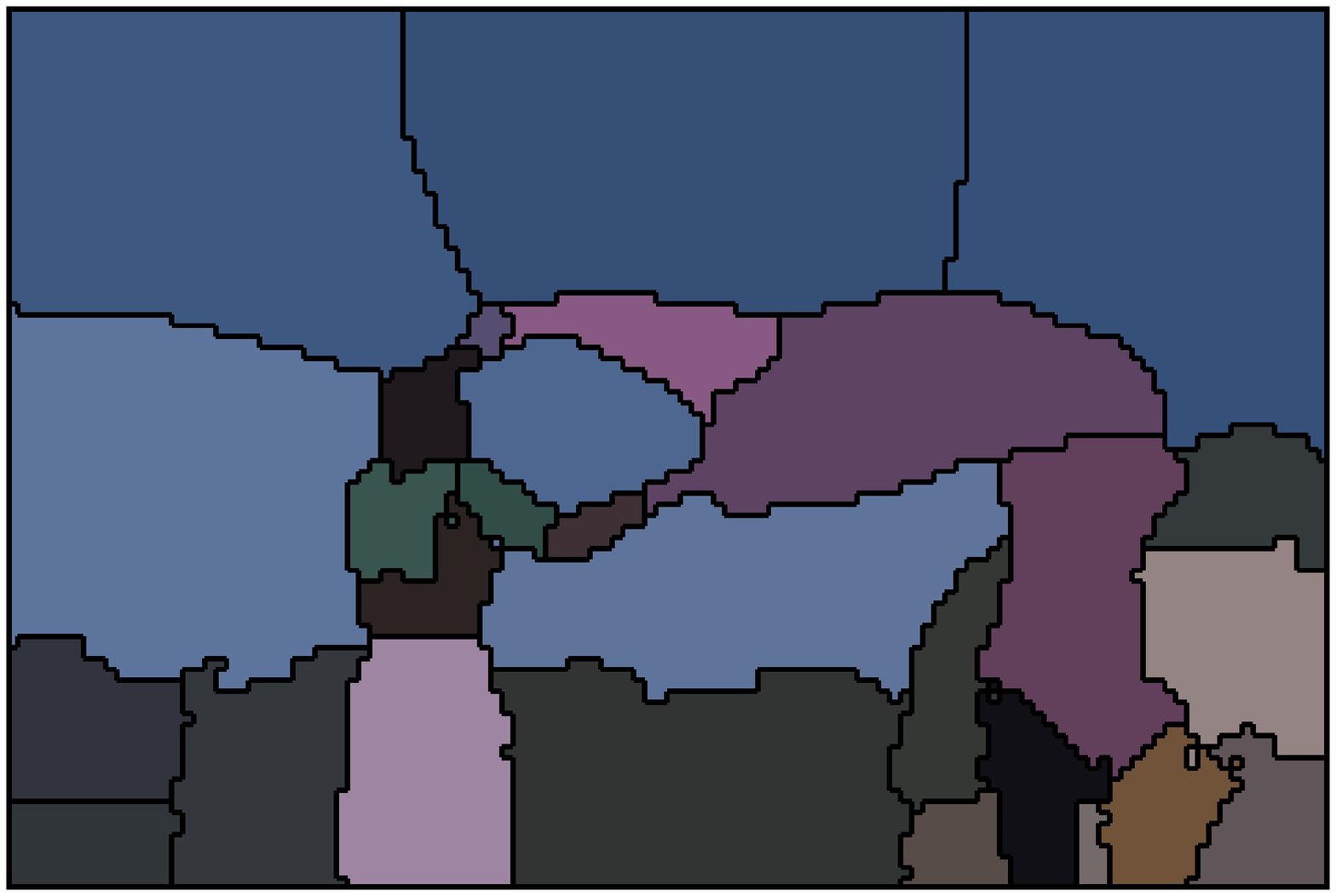} &
				\includegraphics[width=0.7in]{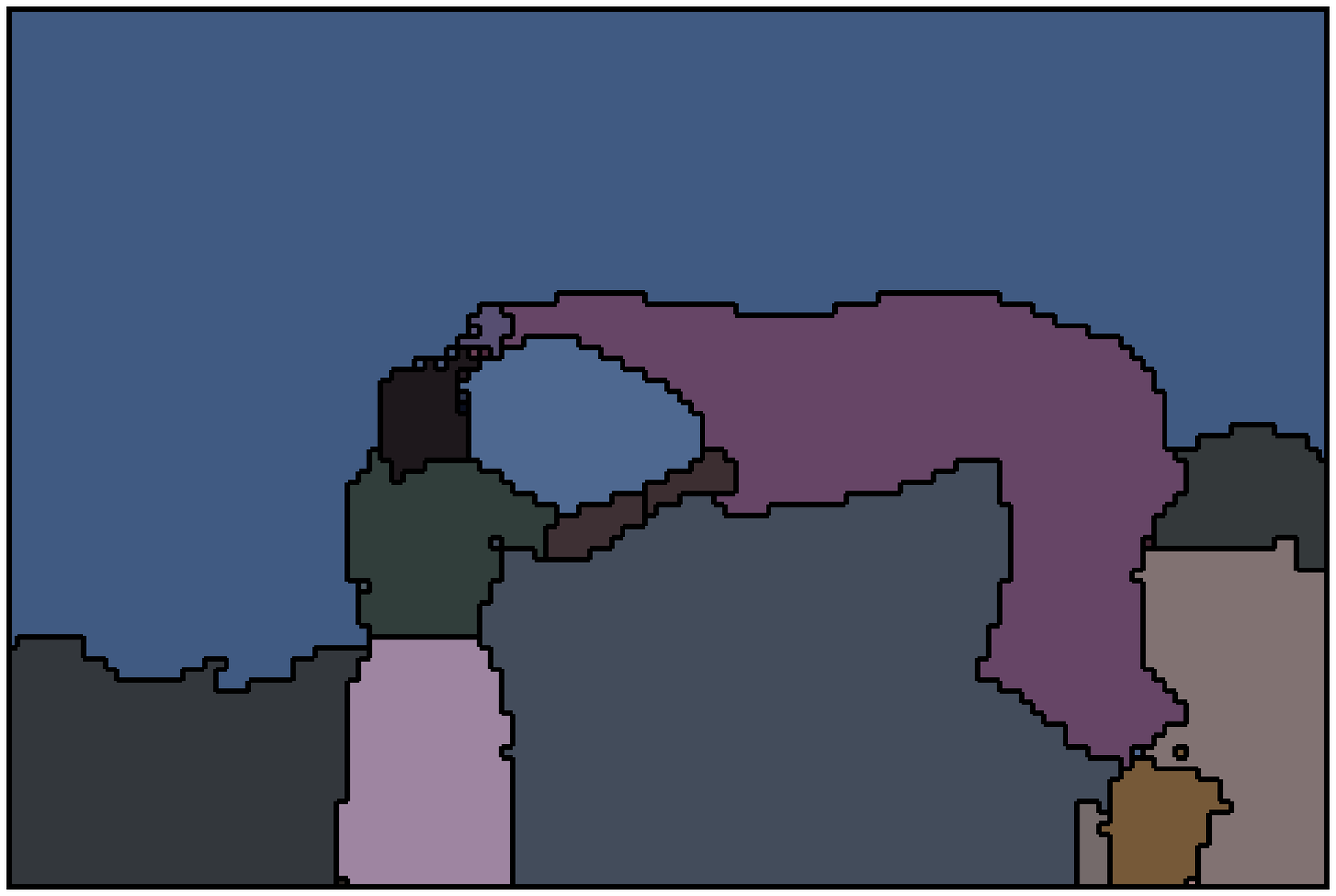} &
				\includegraphics[width=0.7in]{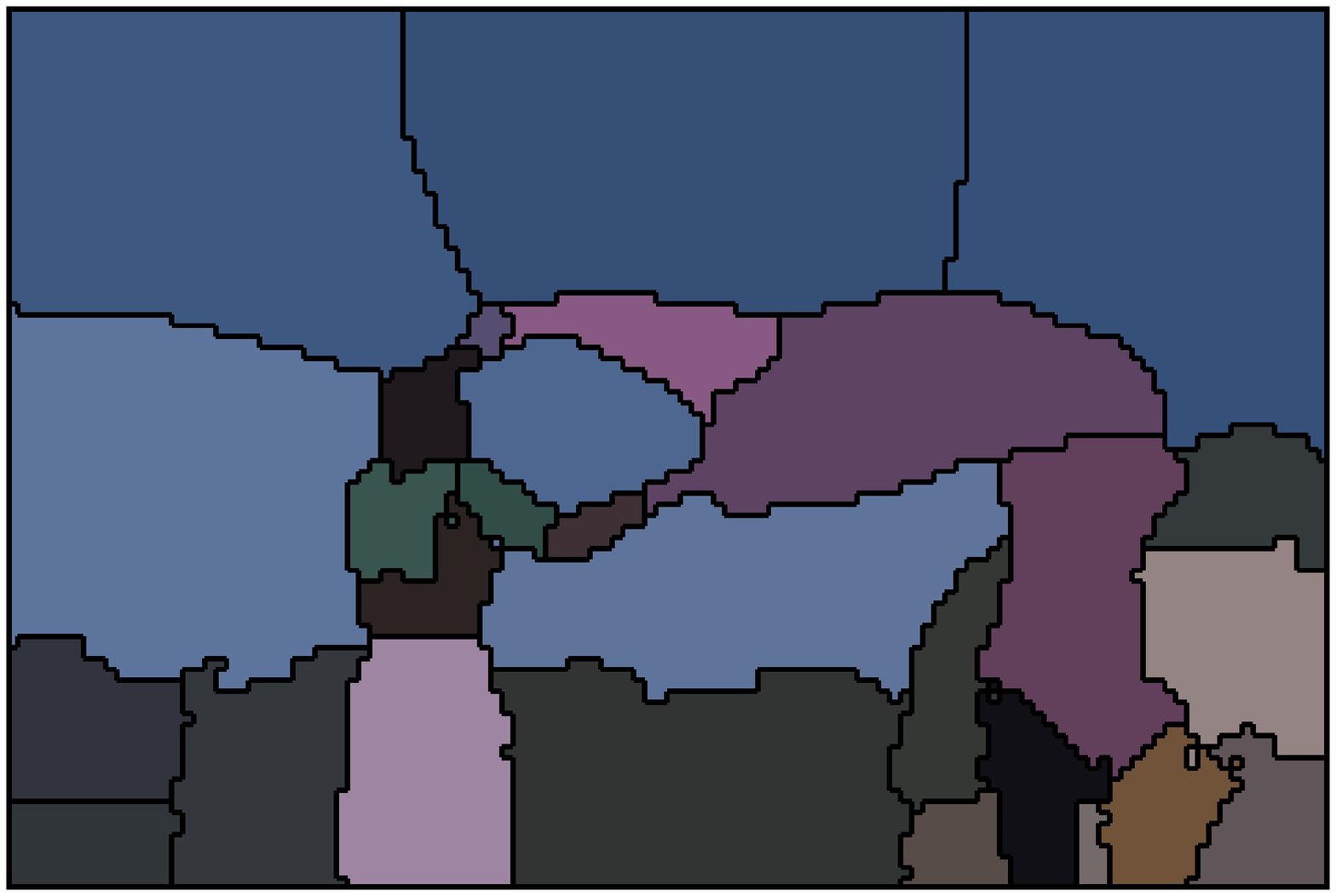} &
				\includegraphics[width=0.7in]{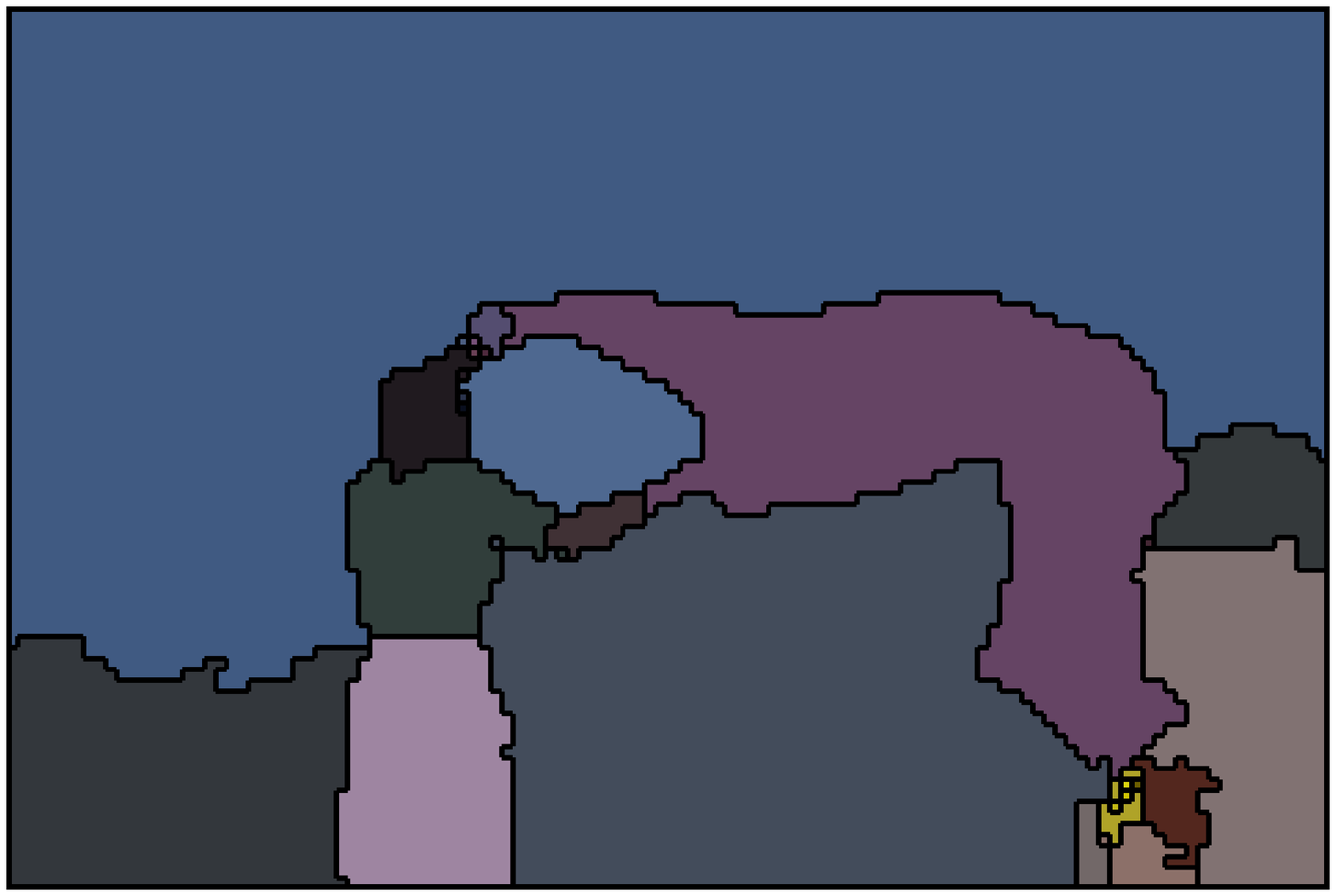} \\				
				{\small Original} & {\small NCut$-$M} & {\small $\textrm{CCN}_{1}-\textrm{M}$} & {\small RCut$-$M} & {\small $\textrm{CCR}_{1}-\textrm{M}$} & {\small NCut$-$H} & {\small $\textrm{CCN}_{1}-\textrm{H}$} & {\small RCut$-$H} & {\small $\textrm{CCR}_{1}-\textrm{H}$}
		\end{tabular}
	\end{center}
		\vspace{-10pt}
		\caption{Example BSDS500 test image and segmentation results. First row: Lab affinities; second row: mPb affinities; third row: PMI affinities. *$-$M indicates multiway segmentation; *$-$H indicates hierarchical 2-way segmentation. $k=28$ clusters specified for each case. Images appearing to have fewer segments actually have a number of small/singleton segments.}
\label{fig:BSDS:Segmentation}
	\vspace{8pt}
	\end{minipage}
	\begin{minipage}[t]{\linewidth}
	\begin{center}
	\scalebox{0.95}{
		\begin{tabular}{P{0.62in} P{0.62in} P{0.62in} P{0.62in} P{0.62in} P{0.62in} P{0.62in} P{0.62in} P{0.62in}}
				\includegraphics[width=0.74in]{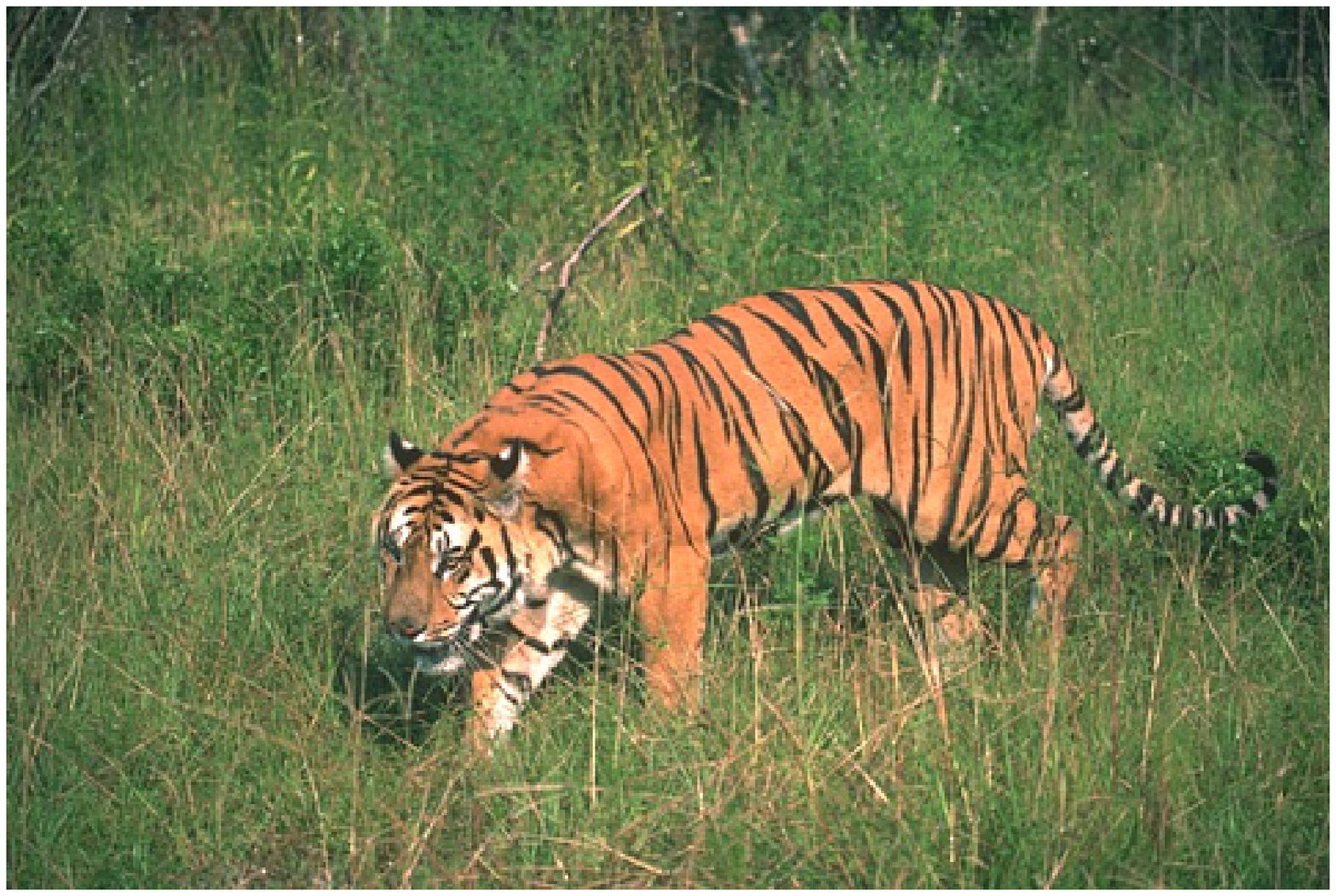} &
				\includegraphics[width=0.74in]{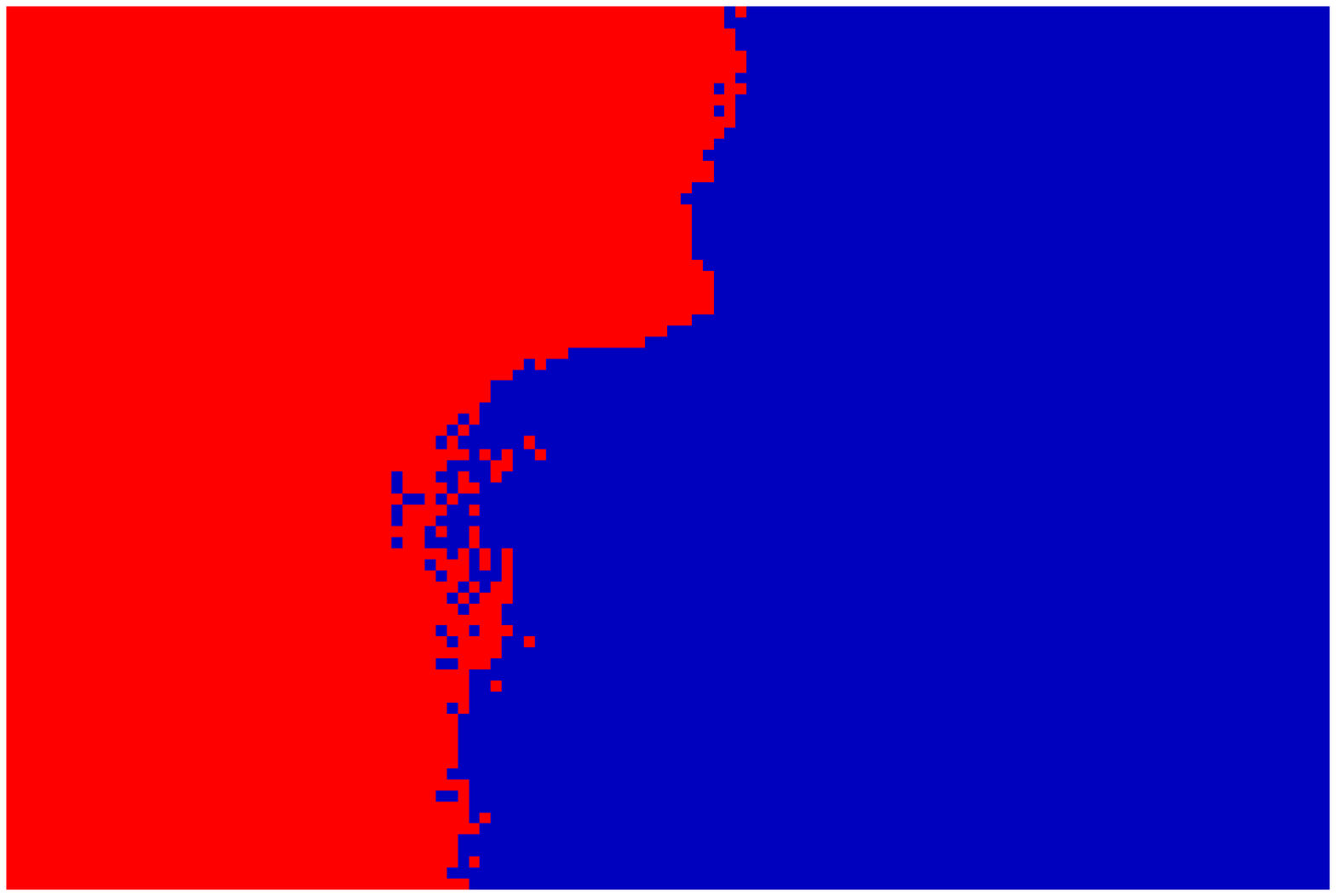} &
				\includegraphics[width=0.74in]{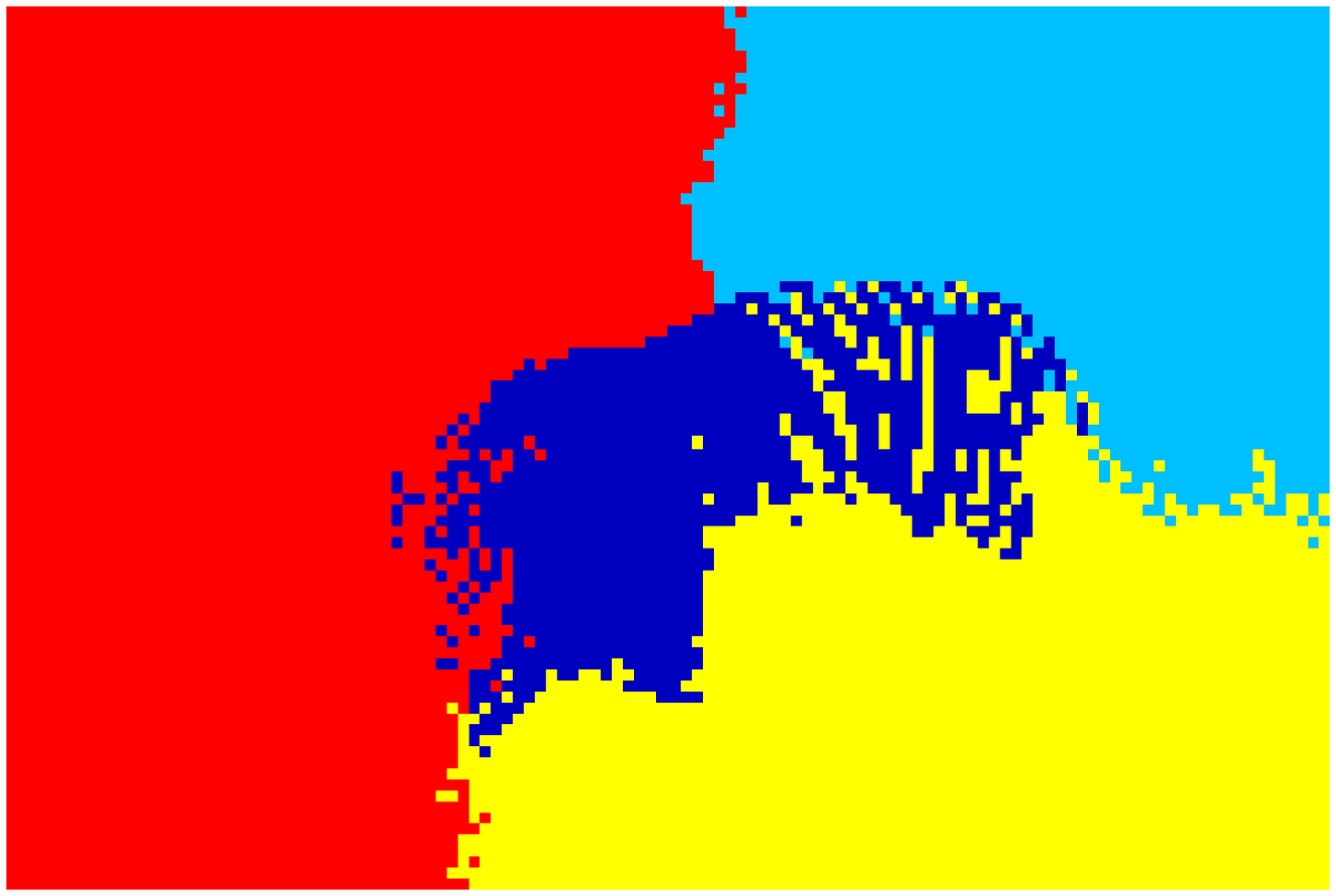} &
				\includegraphics[width=0.74in]{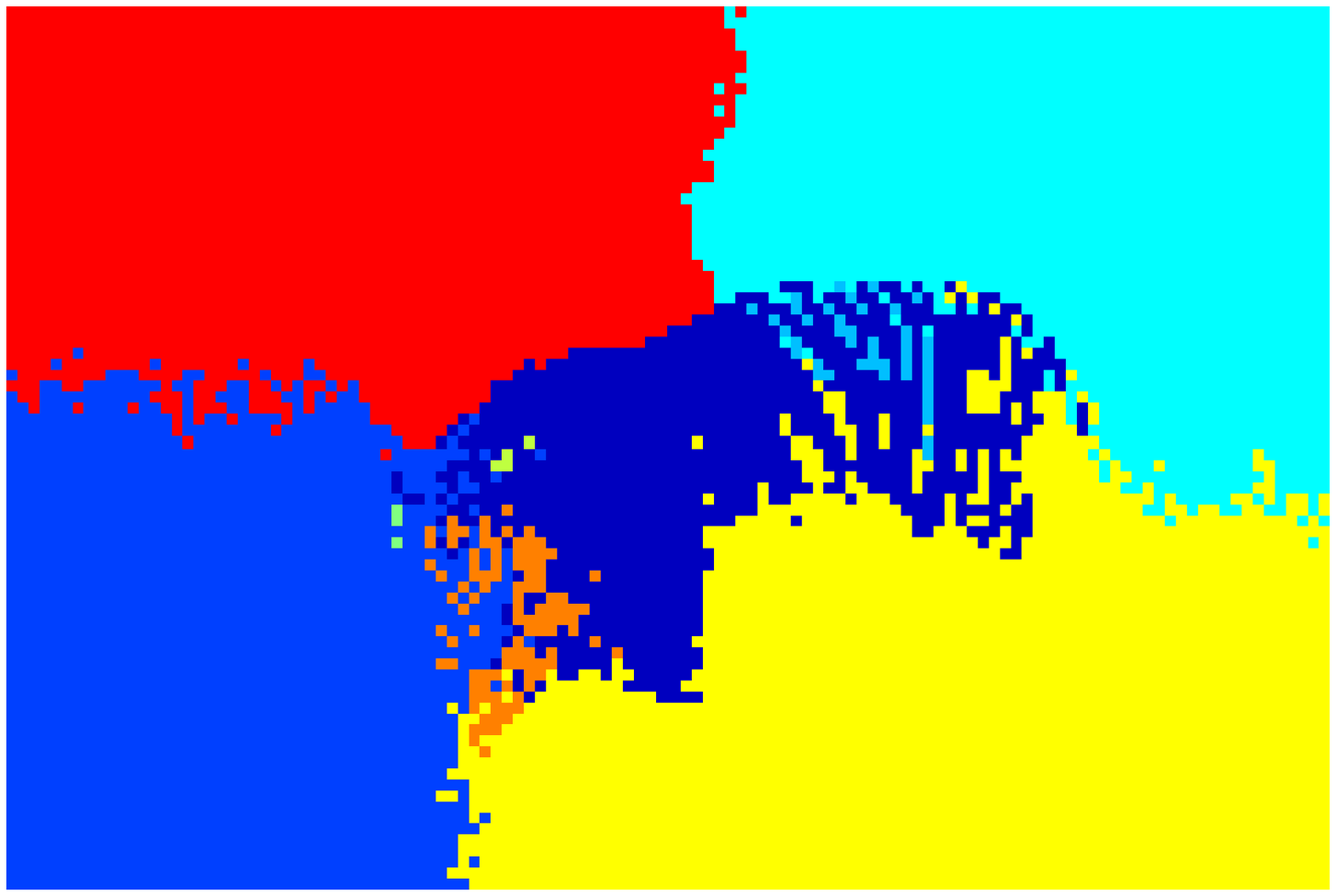} &
				\includegraphics[width=0.74in]{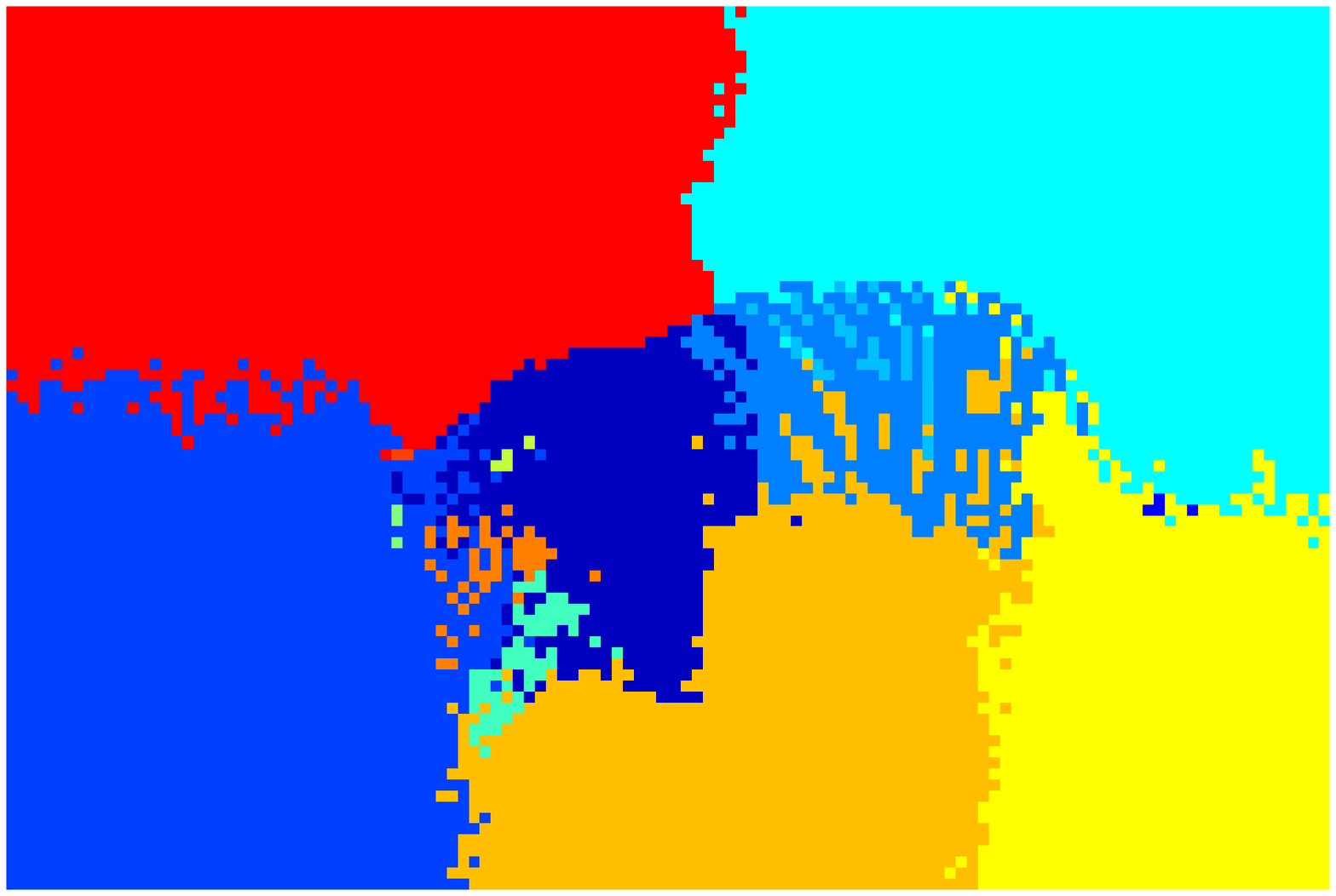} &
				\includegraphics[width=0.74in]{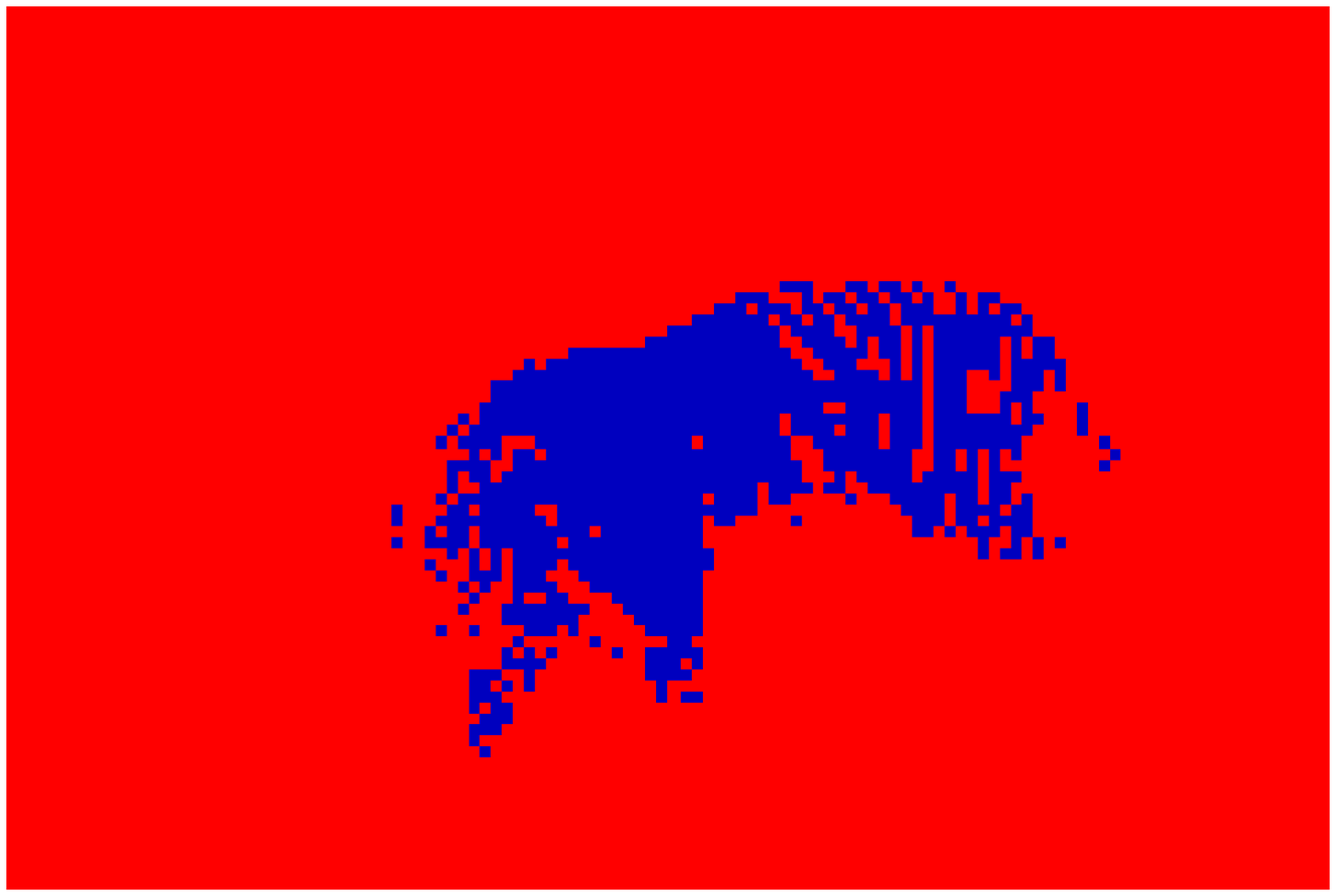} &
				\includegraphics[width=0.74in]{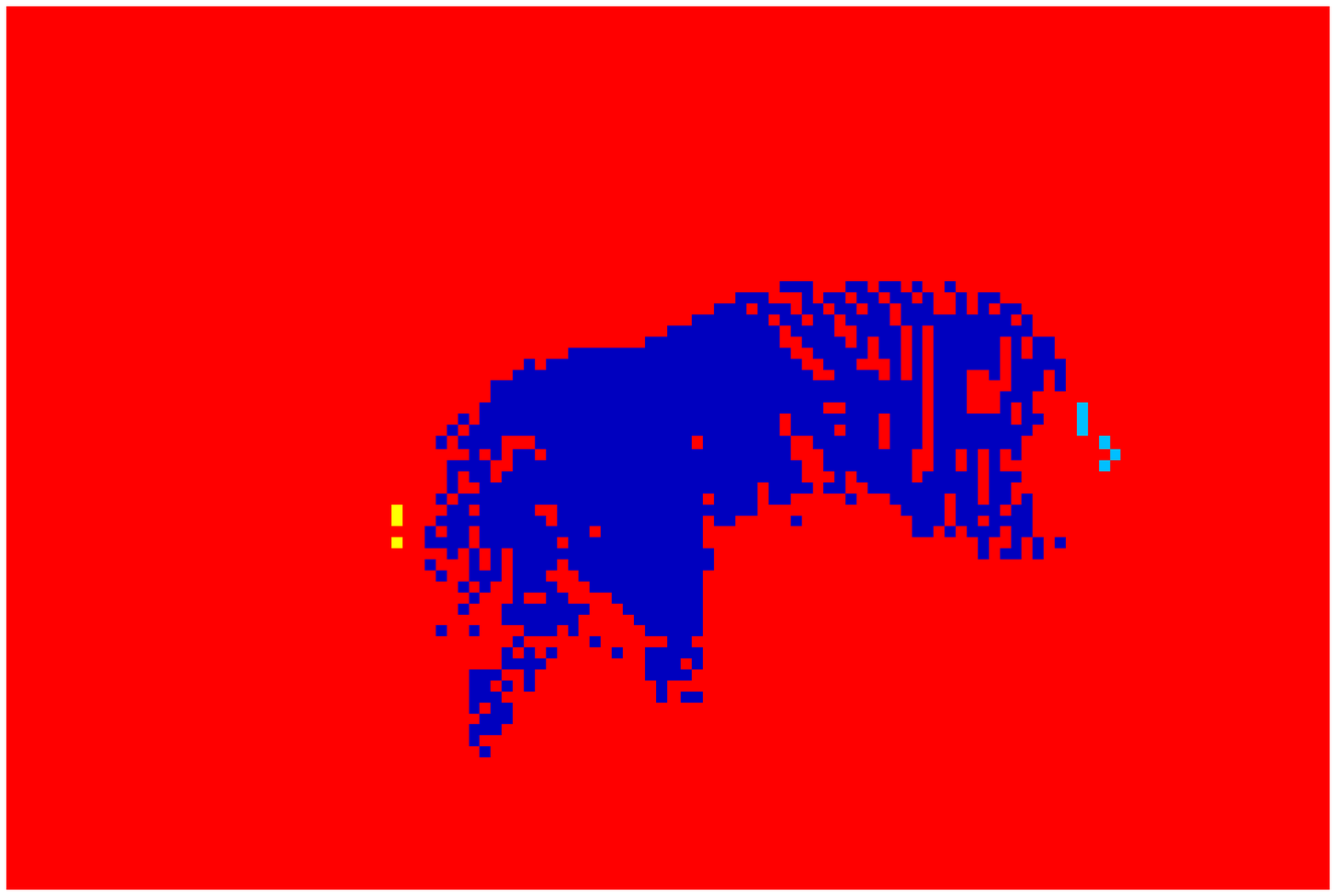} &
				\includegraphics[width=0.74in]{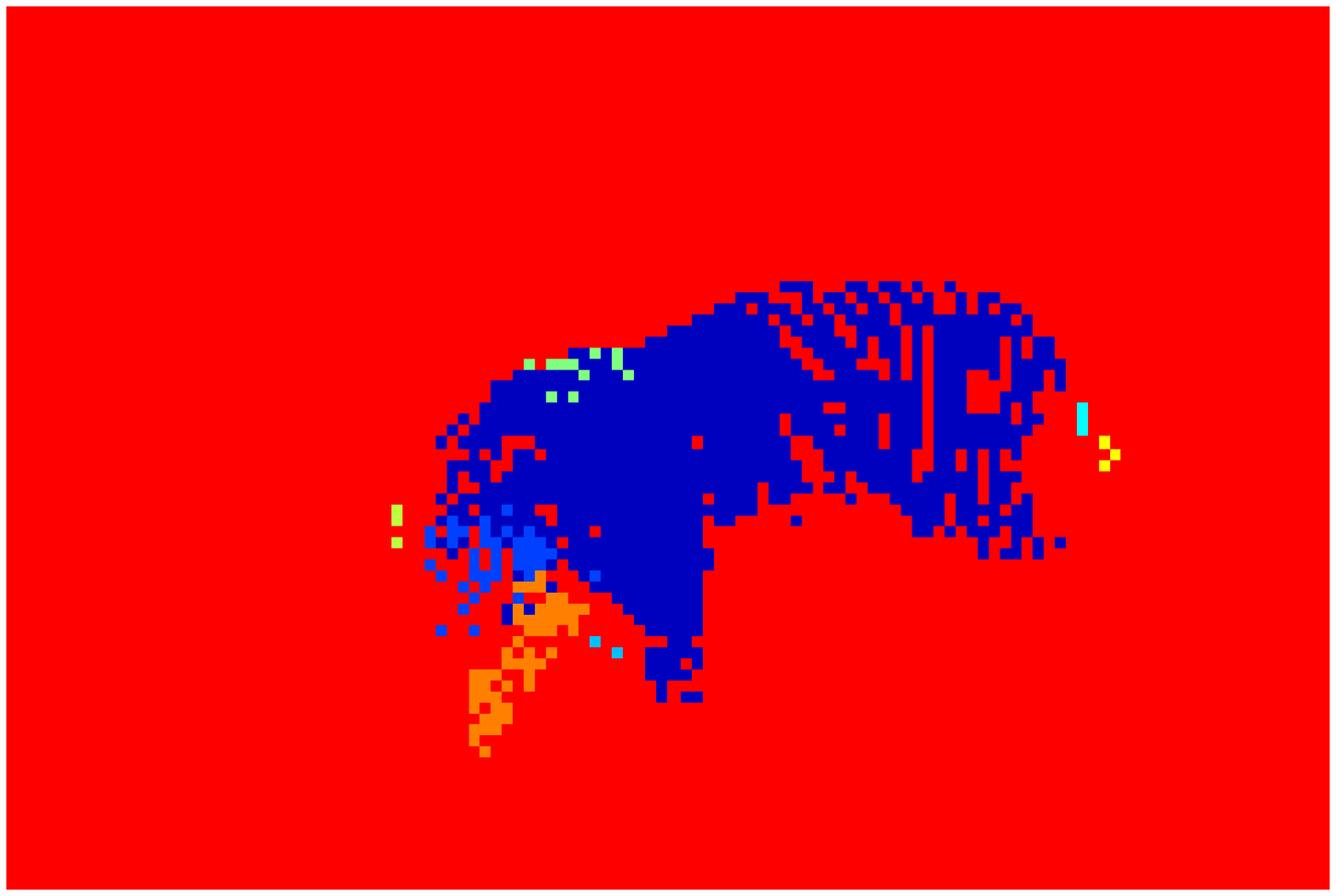} &
				\includegraphics[width=0.74in]{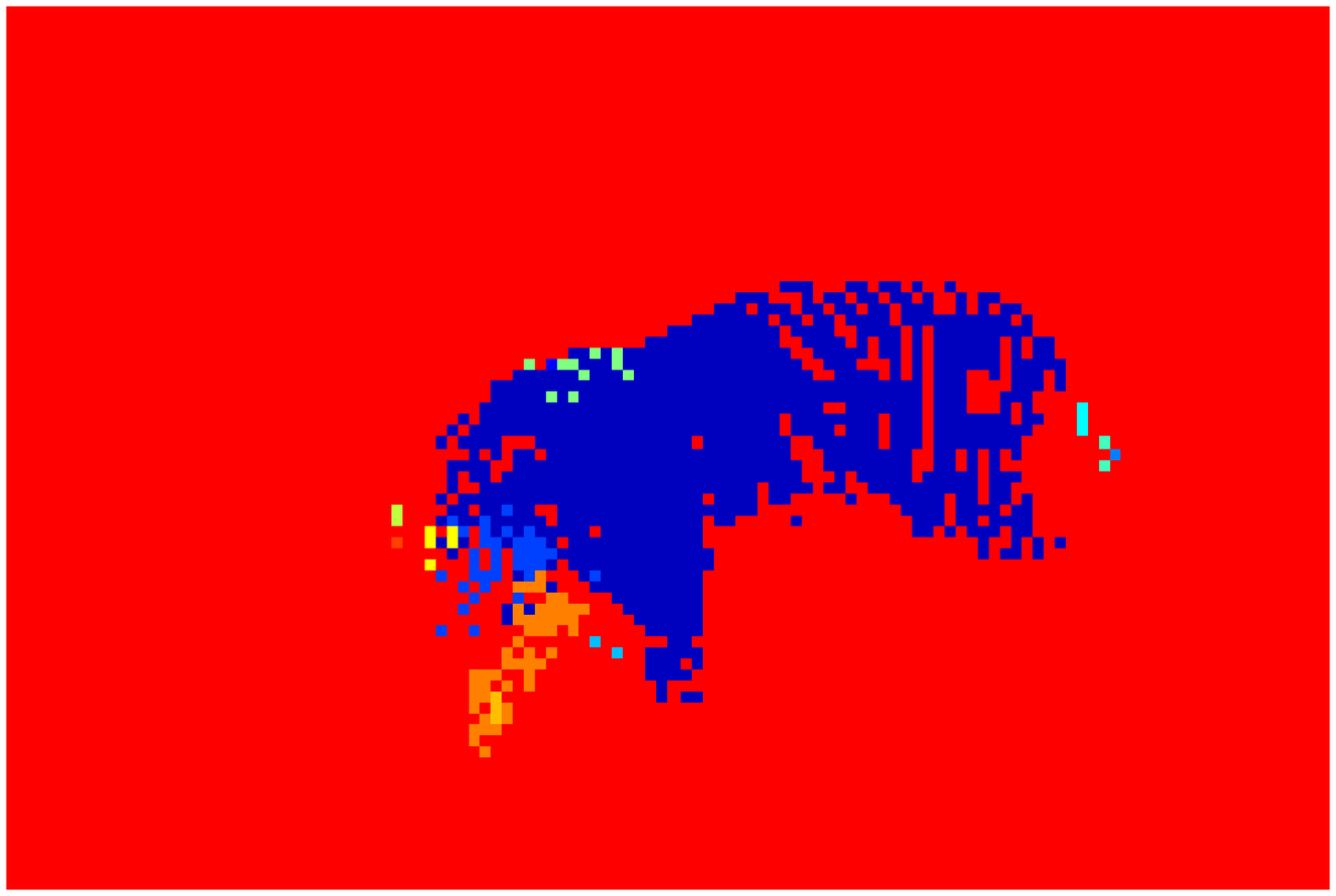} \\
				& \includegraphics[width=0.74in]{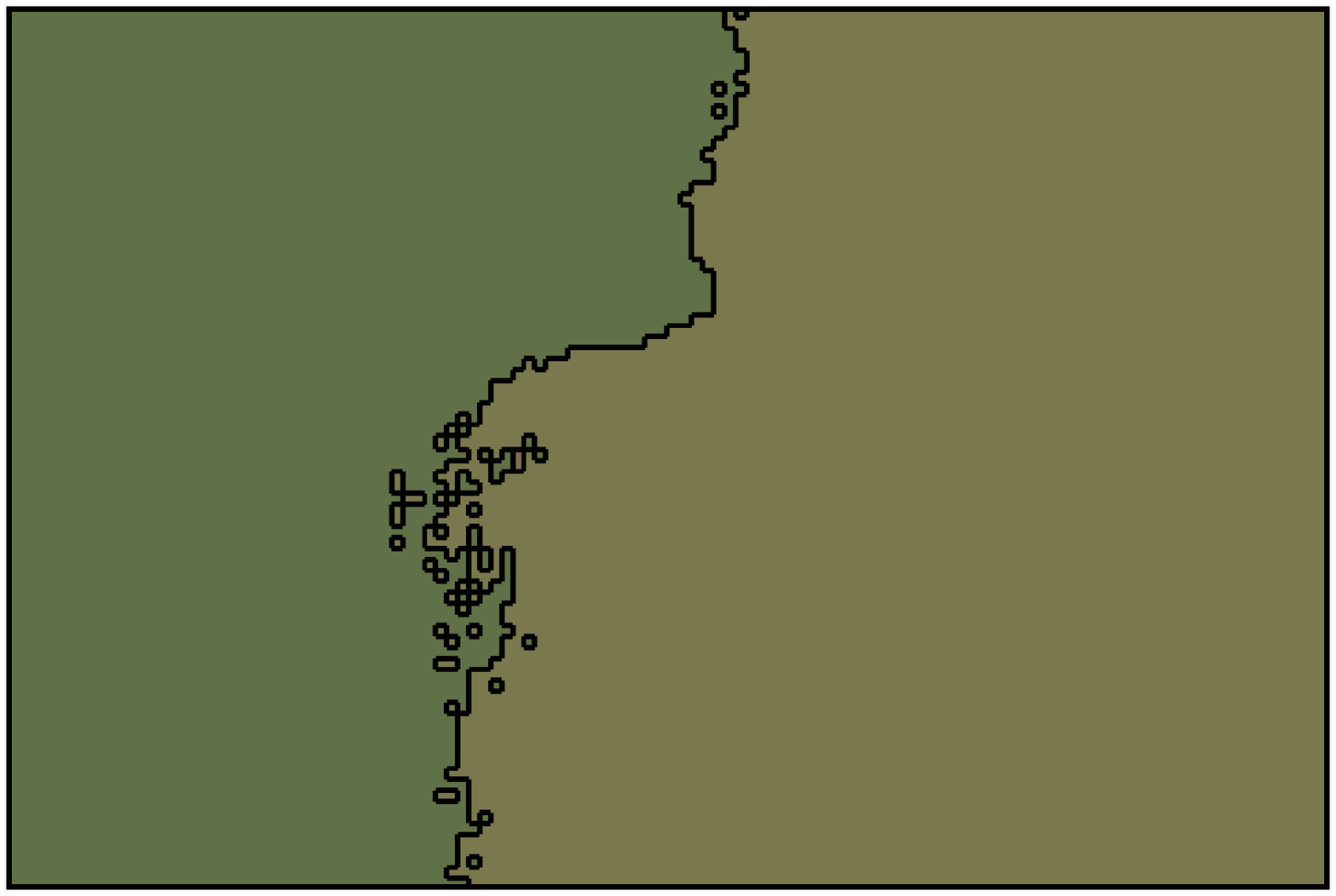} &
				\includegraphics[width=0.74in]{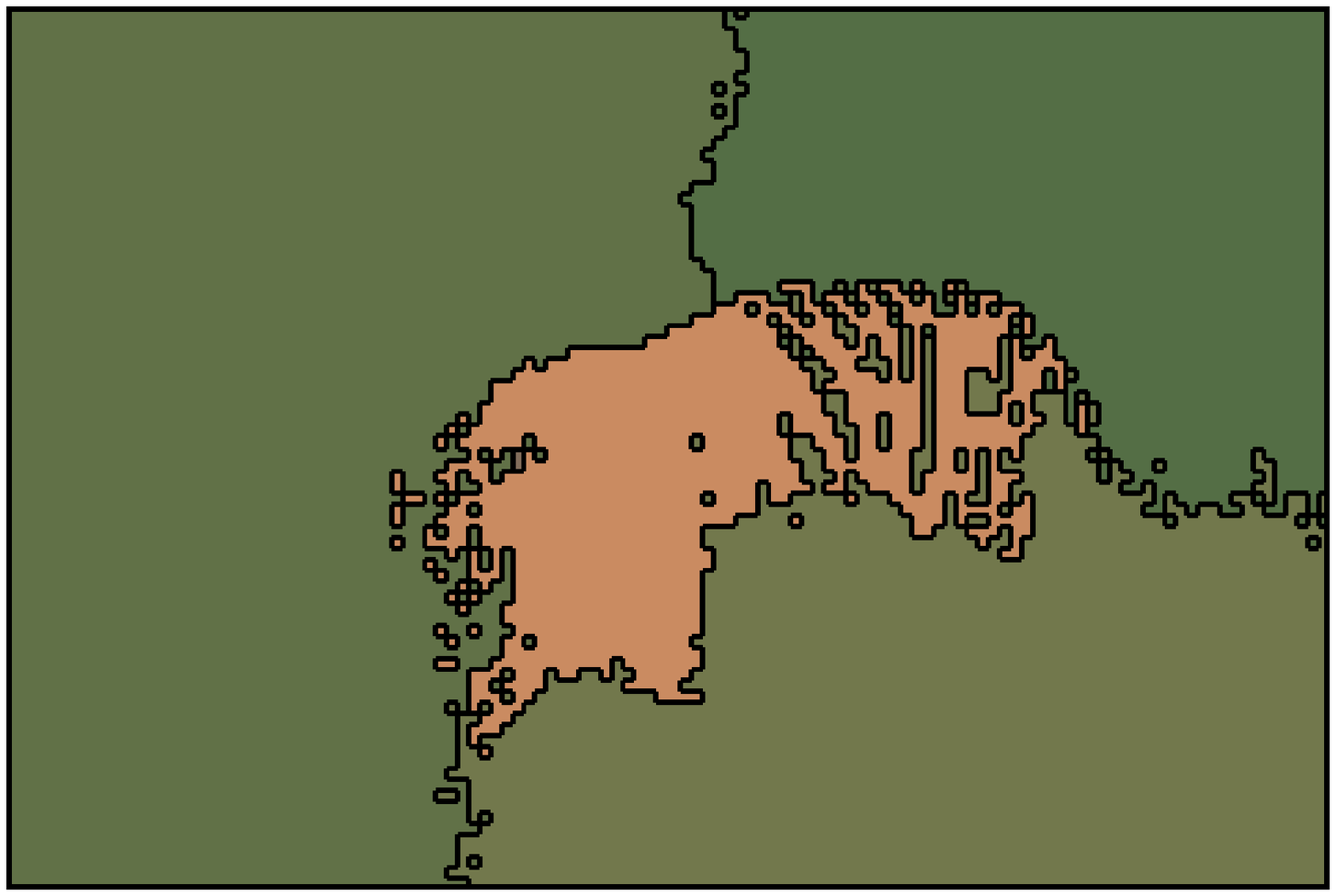} &
				\includegraphics[width=0.74in]{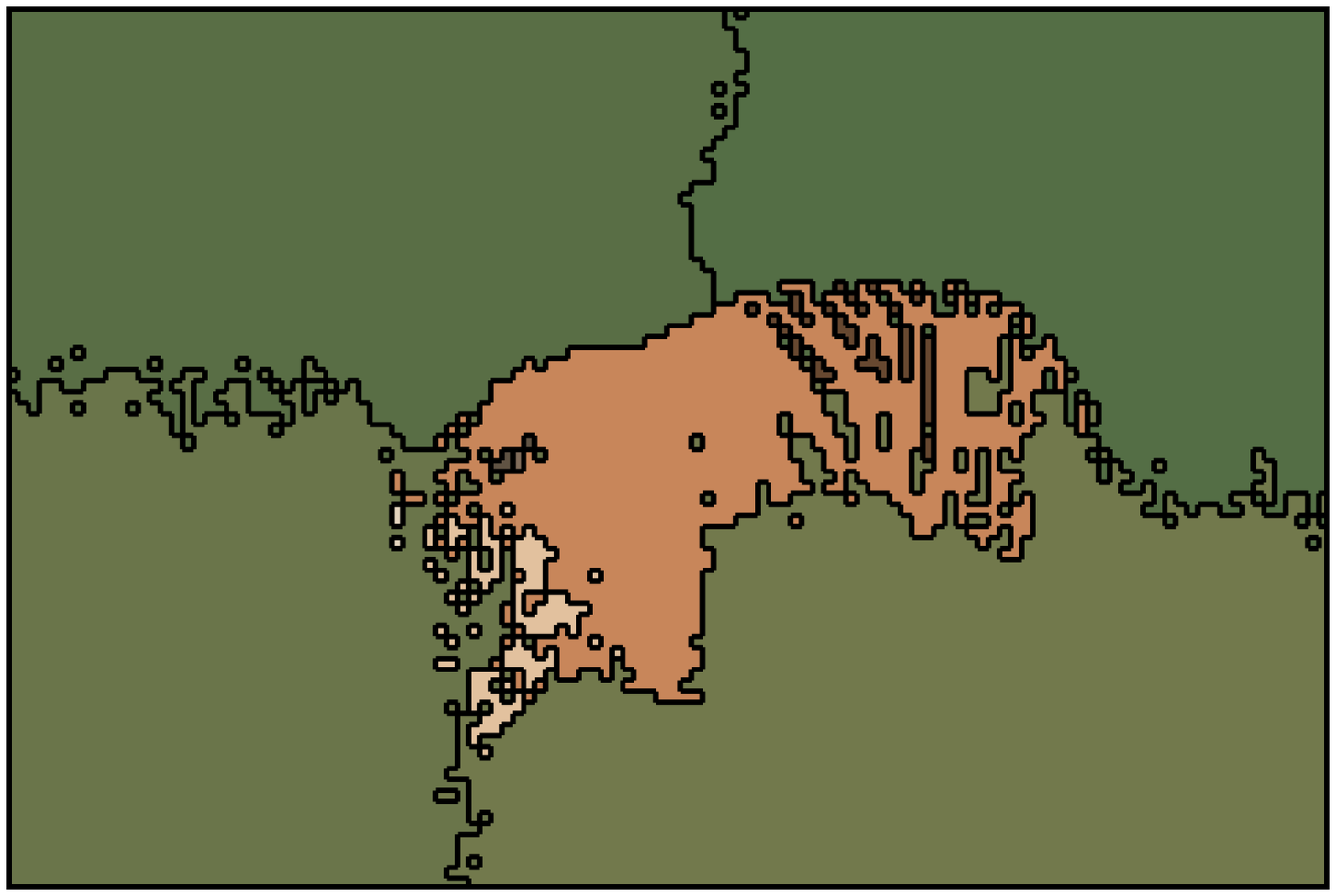} &
				\includegraphics[width=0.74in]{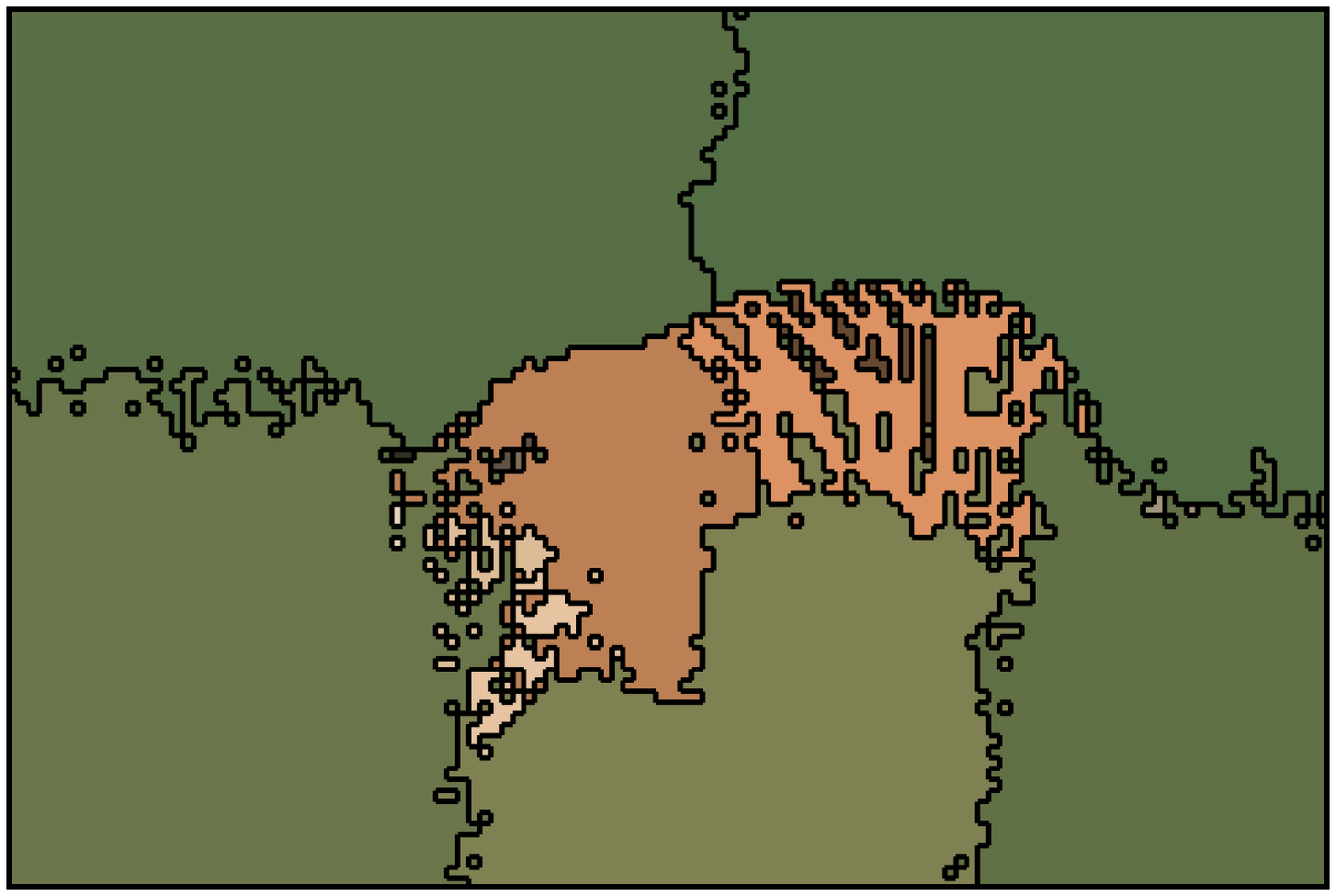} &
				\includegraphics[width=0.74in]{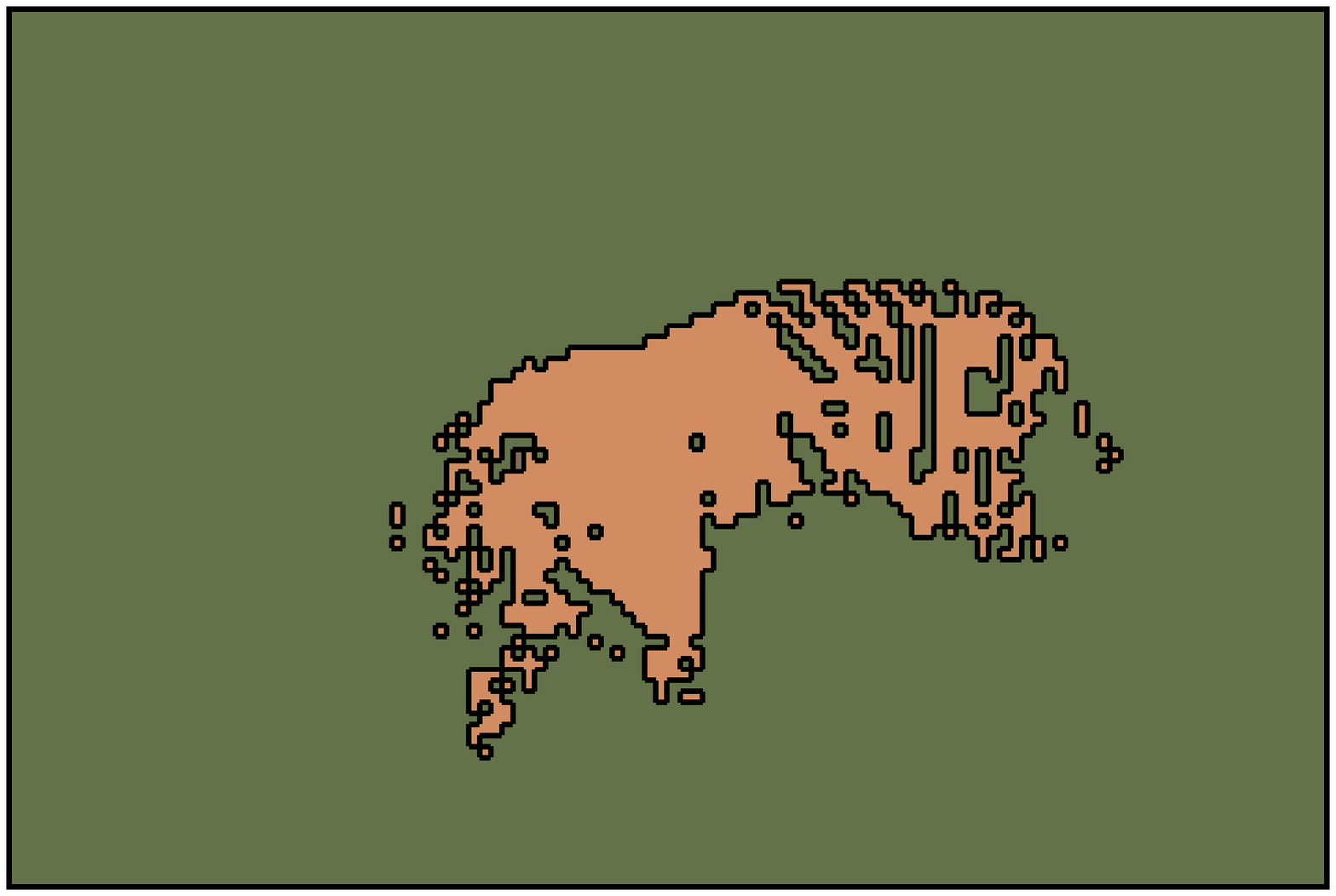} &
				\includegraphics[width=0.74in]{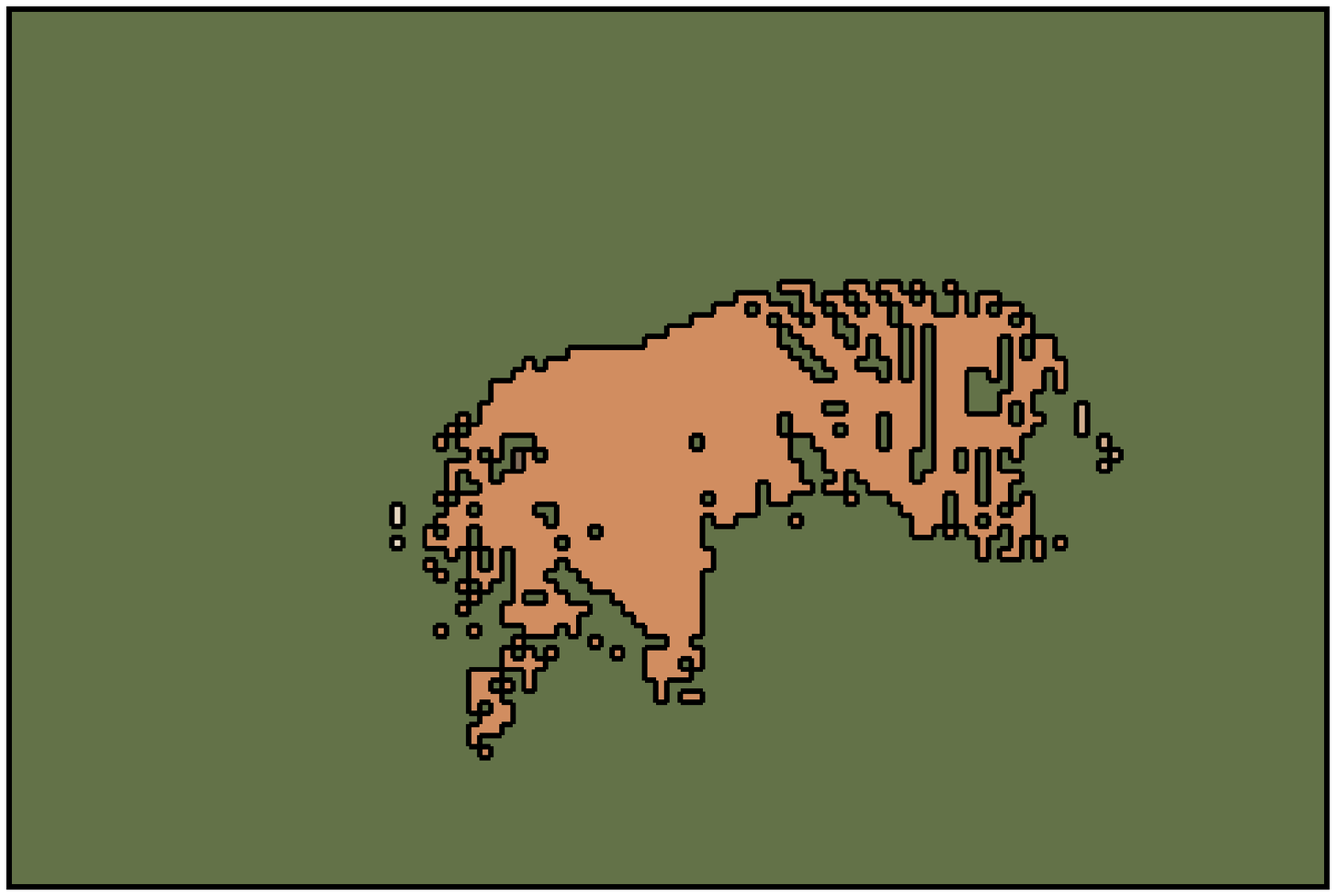} &
				\includegraphics[width=0.74in]{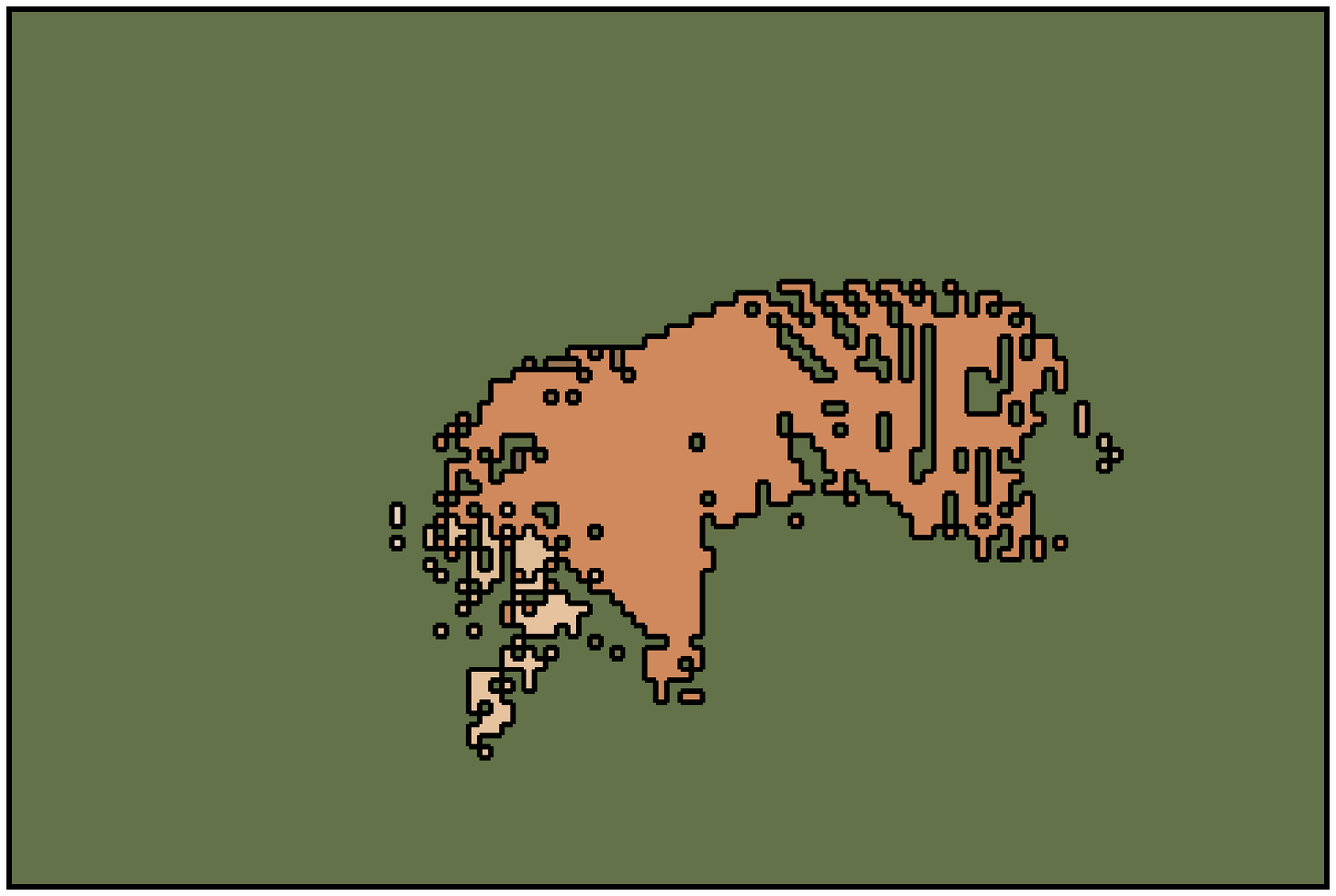} &
				\includegraphics[width=0.74in]{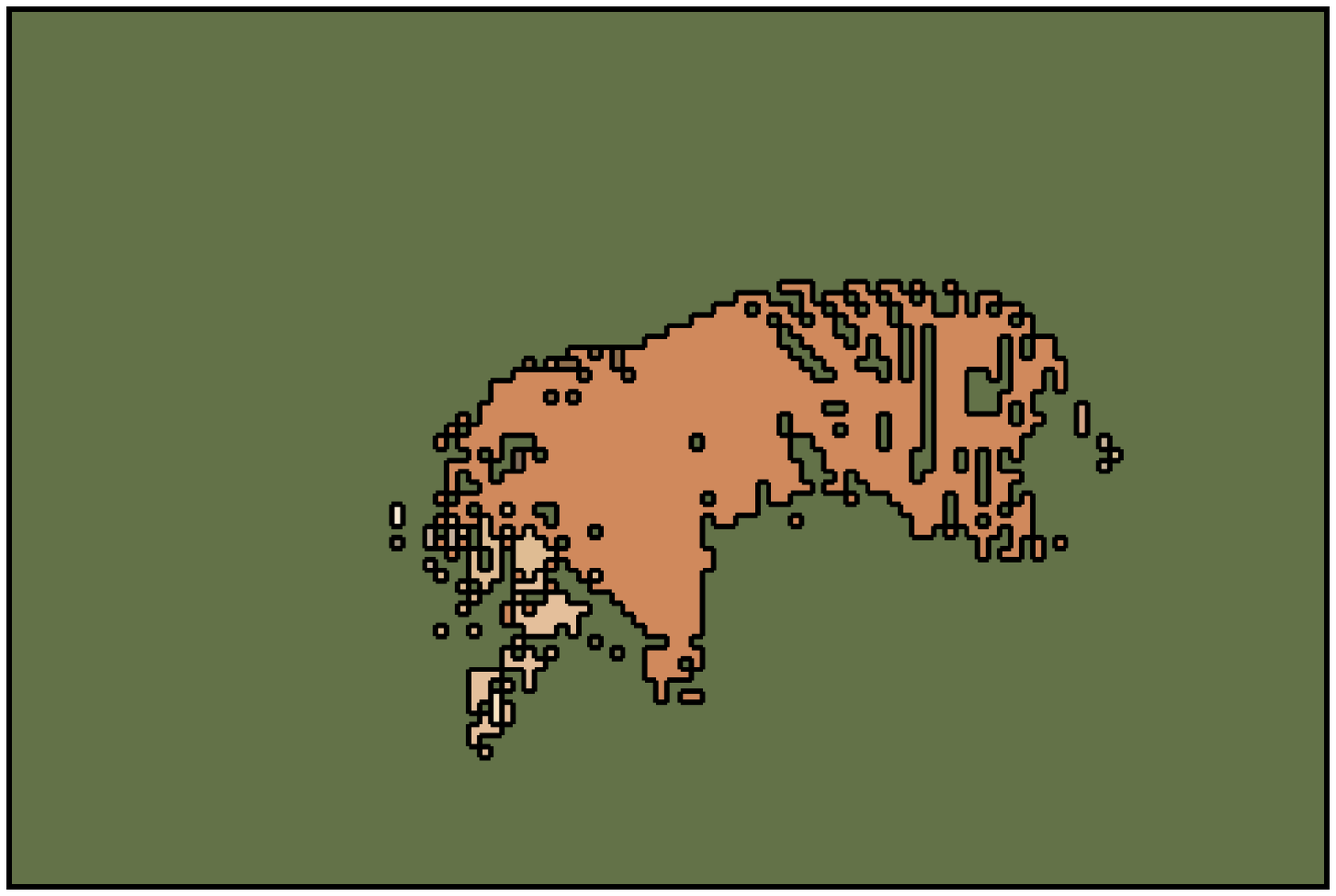} \\
				{\small Original} & {\small $\textrm{NCut:}\,k\!=\!2$} & {\small $k\!=\!4$} & {\small $k\!=\!9$} & {\small $k\!=\!14$} & {\small $\textrm{CCN}_{1}\textrm{:}k\!=\!2$} & {\small $k\!= 4$} & {\small $k\!= 9$} & {\small $k\!=\!14$} 
	\end{tabular}
	}
	\end{center}
		\vspace{-16pt}
		\caption{Hierarchical clustering results for NCuts ($\textrm{CCN}_{2}$ Cuts) and $\textrm{CCN}_{1}$ Cuts for different numbers of clusters $k$ and Lab affinities. Top row: segmentation maps; bottom row: colorized segmentation maps.}
		\label{fig:BSDS:Hierarchical}
	\end{minipage}
\end{figure*}

To investigate the performance of CCB Cuts for image segmentation, we segment the 200 test images in the BSDS500 dataset \cite{arbelaez2011cdh}, each of which has a variety of manually-labeled segmentations with different numbers of segments that can be used as ground truth. We explore three different affinity constructions for each image:
\begin{itemize}
	\item $\textbf{Lab}$: a Gaussian kernel on squared Euclidean distances in $\textrm{L}^{*}\textrm{a}^{*}\textrm{b}^{*}$ color space;
	\item $\textbf{mPb}$: an exponentiated negative maximum of the \emph{multiscale probability of boundary} (mPb) using the code provided with the BSDS500 dataset \cite{arbelaez2011cdh}; and, 
	\item $\textbf{PMI}$: an exponentiated version of a statistic related to \emph{pointwise mutual information} (PMI) used in crisp boundary detection \cite{isola2014cbd}.
\end{itemize}
We use a $10$-pixel radius to define neighborhoods for affinity construction. For efficiency, we downsample each affinity matrix by two scale levels (and subsequently upsample the computed embeddings) using the strategy in \cite{arbelaez2014mcg,pont2017mcg}.  

Figure \ref{fig:BSDS:Segmentation} illustrates segmentation results on a BSDS500 test image for both multi-way and hierarchical 2-way segmentation using various types of balanced cuts. Although different segmentations appear to have different numbers of regions, many of the regions are small/singleton, and all segmentation results have the same number of regions. Figure \ref{fig:BSDS:Hierarchical} shows how hierarchical 2-way segmentation using CCN Cuts with $\tau = 1$ does not exhibit the iterative ``chopping'' behavior common to hierarchical NCuts minimization.

\subsection*{Quantitative Validation}

\begin{figure}[tb]
\centering
\includegraphics[width=3in]{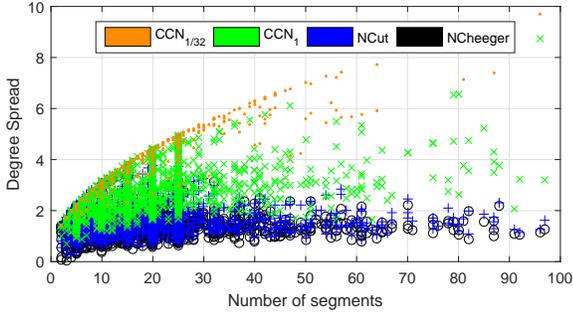}
\caption{Ratio of the standard deviation of partition volumes to the mean partition volume for $\textrm{CCN}_{\tau}$, NCut and Normalized Cheeger Cut-based segmentation of BSDS images for various numbers of segments.}
\label{fig:BSDS:DegreeSpread}
\end{figure}

Drawing any conclusions based on visual assessment of segmentation maps can be problematic in the absence of quantitative validation. First, to validate that CCB segmentation yields less balanced partitions than NCut and Cheeger Cut-based segmentation, we show in Figure \ref{fig:BSDS:DegreeSpread} the ``degree spread" for segmentation results from various methods across all test images, as measured by the ratio of the standard deviation of the partition volumes to the mean partition volume.

Next, we evaluate segmentation performance with respect to ground truth using the three criteria discussed in \cite{arbelaez2011cdh} and used in \cite{yu2015pfe}: Segmentation covering, Probabilistic Rand Index (PRI), and Variation of Information (VI). Covering and PRI increase and VI decreases as segmentations become closer to the ground truth. We segment each BSDS500 test image multiple times by minimizing CCB Cuts for various values of $\tau\in\left(0,2\right]$, once corresponding to each value of $k$ (number of segments) reflected in one of the ground-truth segmentations, and once for each of $k = 5$, $10$, $15$, $20$, and $25$, if any of these are not reflected in the ground truth. Each multi-way segmentation algorithm proceeds by solving/approximating the corresponding embedding and then subsequently performing $k$-means clustering on the result. We also perform segmentation using hierarchical 2-way cuts; in this case, clustering is performed at each step by identifying the threshold that best minimizes the balanced cut cost in question. For comparison purposes, we also apply the inverse power method of \cite{hein2010ipm}; this method hierarchically approximates 2-way cuts that minimize the Normalized/Ratio Cheeger costs.

In CCB Cut minimization (for both multi-way and hierarchical 2-way segmentation), we iterate until the relative change in cost between two iterations falls below $1\%$, or until a maximum of 50 iterations are performed, whichever occurs first. The hyperparameter $\tilde{\kappa}$ is set to $10^{-4}$ for Lab affinities and to $0.2$ for mPb and PMI affinities. Following the strategy of \cite{yu2015pfe}, we report results for both a \emph{fixed} scheme, in which we run the algorithm repeatedly with $k$ corresponding to each number of segments in the ground-truth and average the performance results from the multiple runs, and the \emph{dynamic} scheme, in which we choose the value of $k$ from $k = 5$, $10$, $15$, $20$, and $25$ that yields the best performance for a particular image. Table \ref{tab:CPRIVI:mean} shows the performance results for a subset of the cut costs; to save space, we excluded results from smaller values of $\tau$, which typically performed worse than $\tau=1$ but better than $\tau=2$, and we excluded results from ratio-costs, which are similar to those of normalized-costs.

\begin{table}[t] 
\begin{center}
\scalebox{0.65}{
\begin{tabular}{c|l|cccccc}
    \multirow{2}{*}{Affinity} & \multirow{2}{*}{Method} & \multicolumn{2}{c}{Covering} & \multicolumn{2}{c}{PRI} & \multicolumn{2}{c}{VI} \\
		&& fixed & dynamic & fixed & dynamic & fixed & dynamic \\
    \hline\hline
		Lab & NCut$-$M & 0.48 & 0.65 & 0.70 & 0.83 & 2.17 & 1.50 \\
		\hline
		& $\textrm{CCN}_{1}-\textrm{M}$ & 0.49 & 0.66 & 0.69 & 0.83 & 2.13 & 1.47 \\
		\hline
		& NCut$-$H & 0.51 & 0.67 & \textbf{0.78} & \textbf{0.87} & 1.98 & 1.43 \\
		\hline
		& $\textrm{CCN}_{1}-\textrm{H}$ & \textbf{0.57} & \textbf{0.69} & 0.69 & 0.78 & \textbf{1.81} & \textbf{1.38} \\
		\hline
		& NCheeger& 0.46 & 0.62 & \textbf{0.78} & \textbf{0.87} & 2.21 & 1.66 \\
		\hline\hline
		mPb & NCut$-$M & 0.29 & 0.48 & 0.75 & 0.84 & 2.86 & 2.10 \\
		\hline
		& $\textrm{CCN}_{1}-\textrm{M}$ & 0.32 & 0.51 & 0.75 & 0.83 & 2.75 & 1.96 \\
		\hline
		& NCut$-$H & 0.30 & 0.48 & \textbf{0.83} & \textbf{0.89} & 2.61 & 2.00 \\
		\hline
		& $\textrm{CCN}_{1}-\textrm{H}$ & \textbf{0.38} & \textbf{0.52} & 0.73 & 0.78 & \textbf{2.35} & \textbf{1.83} \\
		\hline
		& NCheeger& 0.27 & 0.48 & \textbf{0.83} & 0.88 & 2.90 & 2.31 \\
		\hline\hline
		PMI & NCut$-$M & 0.34 & 0.53 & 0.76 & 0.85 & 2.62 & 1.86 \\
		\hline
		& $\textrm{CCN}_{1}-\textrm{M}$ & 0.40 & 0.58 & 0.77 & 0.86 & 2.34 & 1.65 \\
		\hline
		& NCut$-$H & 0.37 & 0.54 & \textbf{0.87} & \textbf{0.90} & 2.37 & 1.75 \\
		\hline
		& $\textrm{CCN}_{1}-\textrm{H}$ & \textbf{0.47} & \textbf{0.60} & 0.78 & 0.81 & \textbf{1.99} & \textbf{1.55} \\
		\hline
		& NCheeger& 0.33 & 0.51 & 0.84 & 0.89 & 2.65 & 2.04 \\
		\hline		
\end{tabular}}
\end{center}
\vspace{-10pt}
\caption{Comparison of various segmentation methods on BSDS500 test set, averaged across images. For each affinity construction, the best results for each performance measure are highlighted in bold. *$-$M indicates multiway segmentation; *$-$H indicates hierarchical 2-way segmentation.}
\label{tab:CPRIVI:mean}
\end{table}

A few interesting conclusions can be drawn. First, the use of simple Lab affinities surprisingly yields superior results compared to the more complicated mPb and PMI affinities. Second, PRI values appear to indicate that greater degrees of balance are better, whereas VI and Covering values seem to indicate that CCN with $\tau = 1$ provides the best results. Third, better results are found when hierarchical 2-way segmentation is performed as opposed to simultaneous multiway segmentation. However, in our MATLAB implementation, multiway segmentation is significantly faster than hierarchical 2-way segmentation, as shown in Table \ref{tab:timing}. In the future, it would be interesting to compare these results to those of hierarchical 2-way segmentation via minimizing ratios of differences of set functions \cite{hein2011bsc,buhler2013cfs}.

\begin{table}[t] 
\begin{center}
\scalebox{0.8}{
\begin{tabular}{l|cccc}
    Cost Function & Quartile I & Median & Quartile III & Max \\
    \hline\hline
		$\textrm{CCN}_{\tau}-\textrm{Multiway}$ & 2.0 & 2.9 & 4.2 & 14.9 \\
		\hline
		$\textrm{CCN}_{\tau}-\textrm{Hierarchical}$ & 13.0 & 23.4 & 41.6 & 258.2 \\
		\hline
		NCheeger & 2.4 & 4.0 & 9.4 & 434.6 \\
		\hline
		$\textrm{CCR}_{\tau}-\textrm{Multiway}$ & 1.0 & 1.4 & 1.9 & 8.0 \\
		\hline
		$\textrm{CCR}_{\tau}-\textrm{Hierarchical}$ & 10.0 & 18.6 & 31.6 & 180.7 \\
		\hline
		RCheeger & 1.4 & 2.6 & 9.5 & 378.1 \\
		\hline
\end{tabular}}
\end{center}
\vspace{-10pt}
\caption{Statistics of computation times (minutes) required for segmentation, excluding affinity construction. Statistics are computed across all values of $\tau$, all numbers of clusters, all images, and all affinity types.}
\label{tab:timing}
\end{table}

%%%%%%%%%%%%%%%%%%%%%%%%%%%%%%%%%%%%%%%%%%%%%%%%%%%%%%%%%%%%%
%%%%%%%%%%%%%%%%%%%%%%%%%%%%%%%%%%%%%%%%%%%%%%%%%%%%%%%%%%%%%
\section{Conclusion}

Compassionately Conservative Balanced Cuts enable normalizations ranging from Normalized/Ratio Cuts to unnormalized Cuts, allowing a tradeoff between generating partitions that are too similar in size/degree and avoiding all singletons. They can be directly applied to the $k$-way partitioning problem for any $k\ge 2$, and their minimization can be relaxed into a problem that can be solved by minimizing a succession of Rayleigh quotients. Image segmentation experiments show that minimizing the CCB Cut yields more accurate results than minimizing NCuts/RCuts and generates regions having  greater variability in size/degree.

%%%%%%%%% Appendices
\appendix
\section{Code}
Prototype implementations of algorithms in this paper are available at the MATLAB Central File Exchange under File ID \#66158.

%%%%%%%%%%%%%%%%%%%%%%%%%%%%%%%%%%%%%%%%%%%%%%%%%%%%%%%%%%%%%
%%%%%%%%%%%%%%%%%%%%%%%%%%%%%%%%%%%%%%%%%%%%%%%%%%%%%%%%%%%%%
\section{Acknowledgements}

NDC thanks Thomas B{\"u}hler and Matthias Hein for helpful discussions.

%%%%%%%%%%%%%%%%%%%%%%%%%%%%%%%%%%%%%%%%%%%%%%%%%%%%%%%%%%%%%
%%%%%%%%%%%%%%%%%%%%%%%%%%%%%%%%%%%%%%%%%%%%%%%%%%%%%%%%%%%%%
\section{Proof of Lemma \ref{lemma:YBG}} \label{app:proof:simpleWeights}

Let $\mathbf{G}\in\mathcal{G}$. First, note that $\left(\bm{\aleph}\!\left(\mathbf{G}\right)\right)^{\mathbf{T}}\bm{\Pi}\mathbf{1} = \mathbf{G^{T}B^{T}\bm{\Pi}}^{1/2}\mathbf{1} = \mathbf{G^{T}}\!\left(\mathbf{M^{T}M}\right)^{-1/2}\!\mathbf{M^{T}\bm{\Pi}}^{1/2}\mathbf{1}$, which vanishes because $\mathbf{M^{T}}$ annihilates vectors in the direction of $\mathbf{q}$. Second, note that $\left(\bm{\aleph}\!\left(\mathbf{G}\right)\right)^{\mathbf{T}}\bm{\Pi}\bm{\aleph}\!\left(\mathbf{G}\right) = \mathbf{G^{T}B^{T}BG} = \mathbf{G^{T}}\!\left(\mathbf{M^{T}M}\right)^{-1/2}\!\mathbf{M^{T}M}\!\left(\mathbf{M^{T}M}\right)^{-1/2}\mathbf{G} = \mathbf{G^{T}G} = \mathbf{I}$. From these two properties, we see that $\bm{\aleph}\!\left(\mathbf{G}\right)\in\mathcal{Y}$, and so $\bm{\aleph}:\mathcal{G}\rightarrow\mathcal{Y}^{\prime}\subseteq\mathcal{Y}$. 

Now, define $\bm{\beth}\!\left(\mathbf{Y}\right) = \mathbf{B}^{\dagger}\bm{\Pi}^{1/2}\mathbf{Y}$ for $\mathbf{Y}\in\mathcal{Y}$, where $\mathbf{B}^{\dagger} = \left(\mathbf{B^{T}B}\right)^{-1}\!\mathbf{B^{T}} = \left(\mathbf{M^{T}M}\right)^{-1/2}\!\mathbf{M^{T}}$. Then $\left(\bm{\beth}\!\left(\mathbf{Y}\right)\right)^{\mathbf{T}}\!\bm{\beth}\!\left(\mathbf{Y}\right) = \mathbf{Y^{T}\bm{\Pi}}^{1/2}\left(\mathbf{B}^{\dagger}\right)^{\mathbf{T}}\!\mathbf{B}^{\dagger}\bm{\Pi}^{1/2}\mathbf{Y} = \mathbf{Y^{T}\bm{\Pi}}^{1/2}\mathbf{M}\!\left(\mathbf{M^{T}M}\right)^{-1}\!\mathbf{M}^{\mathbf{T}}\bm{\Pi}^{1/2}\mathbf{Y}$. Noting that $\mathbf{M}\!\left(\mathbf{M^{T}M}\right)^{-1}\!\mathbf{M}^{\mathbf{T}}$ is the orthogonal projector onto the subspace orthogonal to $\mathbf{q}$, and that this orthogonal projector can be expressed alternatively as $\mathbf{I} - \mathbf{qq^{T}}$, we can write $\left(\bm{\beth}\!\left(\mathbf{Y}\right)\right)^{\mathbf{T}}\!\bm{\beth}\!\left(\mathbf{Y}\right) = \mathbf{Y^{T}\bm{\Pi}Y} - \mathbf{Y^{T}\bm{\Pi}}^{1/2}\mathbf{qq^{T}\bm{\Pi}}^{1/2}\mathbf{Y} = \mathbf{I}_{k}$ because $\mathbf{Y^{T}\bm{\Pi}}^{1/2}\mathbf{q} = \mathbf{Y^{T}\bm{\Pi}1}\!/\!\left\|\mathbf{\bm{\Pi}}^{1/2}\mathbf{1}\right\| = \mathbf{0}$ for $\mathbf{Y}\in\mathcal{Y}$. Hence, $\bm{\beth}:\mathcal{Y}\rightarrow\mathcal{G}^{\prime}\subseteq\mathcal{G}$. 

Now, suppose $\exists\mathbf{Y}^{\ast}\in\mathcal{Y}\backslash\mathcal{Y}^{\prime}$. Then $\exists\mathbf{G}^{\ast}\in\mathcal{G}$ such that $\mathbf{G}^{\ast} = \bm{\beth}\!\left(\mathbf{Y}^{\ast}\right)$. Note that $\bm{\aleph}\!\left(\mathbf{G}^{\ast}\right) = \bm{\aleph}\!\circ\!\bm{\beth}\!\left(\mathbf{Y}^{\ast}\right) = \bm{\Pi}^{-1/2}\mathbf{BB}^{\dagger}\bm{\Pi}^{1/2}\mathbf{Y}^{\ast} = \bm{\Pi}^{-1/2}\mathbf{B}\!\left(\mathbf{B^{T}B}\right)^{-1}\!\mathbf{B^{T}\bm{\Pi}}^{1/2}\mathbf{Y}^{\ast}$. Since $\mathbf{B}\!\left(\mathbf{B^{T}B}\right)^{-1}\!\mathbf{B^{T}}$
is the orthogonal projector onto the column space of $\mathbf{B}$, and this column space is equivalent to the column space of $\mathbf{M}$, $\mathbf{B}\!\left(\mathbf{B^{T}B}\right)^{-1}\!\mathbf{B^{T}} = \mathbf{I}-\mathbf{qq^{T}}$. Hence, $\bm{\aleph}\!\left(\mathbf{G}^{\ast}\right) = \bm{\Pi}^{-1/2}\!\left(\mathbf{I}-\mathbf{qq^{T}}\right)\!\bm{\Pi}^{1/2}\mathbf{Y}^{\ast} = \mathbf{Y}^{\ast} \in \mathcal{Y}^{\prime}$, a contradiction. Hence, $\mathcal{Y}^{\prime}=\mathcal{Y}$, and therefore, $\bm{\aleph}$ is a surjection from $\mathcal{G}$ to $\mathcal{Y}$.

Finally, $\bm{\beth}\circ\bm{\aleph}\!\left(\mathbf{G}\right) = \mathbf{B}^{\dagger}\bm{\Pi}^{1/2}\bm{\Pi}^{-1/2}\mathbf{B}\mathbf{G} = \mathbf{B}^{\dagger}\mathbf{B}\mathbf{G} = \mathbf{G}$, and so $\bm{\aleph}$ is invertible. Hence, $\bm{\aleph}$ is a bijection from $\mathcal{G}$ to $\mathcal{Y}$. 

%%%%%%%%%%%%%%%%%%%%%%%%%%%%%%%%%%%%%%%%%%%%%%%%%%%%%%%%%%%%%
%%%%%%%%%%%%%%%%%%%%%%%%%%%%%%%%%%%%%%%%%%%%%%%%%%%%%%%%%%%%%
\section{Computing $\mathbf{B^{T}}\bm{\Pi}^{-1/2}\mathbf{L}\bm{\Pi}^{-1/2}\mathbf{Bx}$} \label{app:proof:BTLB}
If $\mathbf{M^{T}} = \left[\mathbf{\hat{q}}\,|-q_{1}\mathbf{I}_{n-1} \right]$, it can be verified that $\left(\mathbf{M^{T}M}\right)^{-1/2}\!\mathbf{M^{T}} = \left[\mathbf{\hat{q}}\,|-\mathbf{I}_{n-1}+\mathbf{\hat{q}\hat{q}^{T}}\!/\!\left(1+q_{1}\right)\right]$. Hence, $\mathbf{s} = \mathbf{B^{T}}\bm{\Pi}^{-1/2}\mathbf{L}\bm{\Pi}^{-1/2}\mathbf{Bx}$ can be computed without constructing any dense matrix by performing the following steps:
\begin{enumerate}
	\item $\mathbf{p} := \left[\mathbf{\hat{q}^{T}x}\,|-\mathbf{t^{T}}+\left(\frac{\mathbf{\hat{q}^{T}x}}{1+q_{1}}\right)\mathbf{\hat{q}^{T}} \right]^{\mathbf{T}}$,
	\item $\mathbf{r} := \bm{\Pi}^{-1/2}\mathbf{L}\bm{\Pi}^{-1/2}\mathbf{p}$,
	\item $\mathbf{s} := \left(\frac{r_{1}+\mathbf{r^{T}q}}{1+q_{1}}\right)\!\mathbf{\hat{q}} - \mathbf{\hat{r}}$, where $\mathbf{\hat{r}} = \left[r_{2},r_{3},\ldots ,r_{n} \right]^{\mathbf{T}}$.
\end{enumerate}

%------------------------------------------------------------------------

{\small
\bibliographystyle{ieee}
\bibliography{refs}
}

\end{document}